\documentclass[10pt,twocolumn,letterpaper]{article}
\usepackage[accsupp]{axessibility}  

\usepackage[cvpr]{cvpr}      

\usepackage{graphicx}
\usepackage{amsmath}
\usepackage{amssymb}
\usepackage{booktabs}
\usepackage{url}
\usepackage{multirow}
\usepackage{graphics}
\usepackage{graphicx}
\usepackage{amsfonts}
\usepackage{textcomp}
\usepackage{amssymb}
\usepackage{mathrsfs}
\usepackage{amsmath}
\usepackage{bbm}
\usepackage{color}
\usepackage[normalem]{ulem}
\usepackage{adjustbox}
\usepackage{diagbox}
\usepackage{tabu}
\usepackage{algorithm}
\usepackage{algpseudocode}
\usepackage{epigraph}
\usepackage[font=small,labelfont=bf]{caption}
\usepackage{verbatim}
\usepackage{wrapfig}

\usepackage{enumitem}

\definecolor{darkgreen}{rgb}{0.2, 0.76, 0.25}
\definecolor{darkred}{rgb}{0.8, 0.16, 0.25}
\definecolor{darkblue}{rgb}{0.0, 0.16, 0.75}

\definecolor{maroon}{rgb}{0.76, 0.13, 0.28}

\definecolor{codegreen}{rgb}{0,0.6,0}
\definecolor{codegray}{rgb}{.95,.95, .95}
\definecolor{codepurple}{rgb}{0.58,0,0.82}
\definecolor{backcolour}{rgb}{0.95,0.95,0.92}

\usepackage[pagebackref=true,breaklinks=true,colorlinks,bookmarks=false,pagebackref,breaklinks,colorlinks]{hyperref}
\hypersetup{
colorlinks=true,
linkcolor=darkred,
filecolor=magenta,      
urlcolor=darkblue,
citecolor=darkgreen,
}

\usepackage[capitalize]{cleveref}
\crefname{section}{Sec.}{Secs.}
\Crefname{section}{Section}{Sections}
\Crefname{table}{Table}{Tables}
\crefname{table}{Tab.}{Tabs.}

\newcommand{\sheng}[1]{\textcolor{black}{#1}}

\newcommand{\shengc}[1]{\textcolor{black}{#1}}

\def\cvprPaperID{3} 
\def\confName{CVPR}
\def\confYear{2022}

\begin{document}

\title{SSR-GNNs: Stroke-based Sketch Representation with Graph Neural Networks}

\author{
Sheng Cheng\\
{\tt\small scheng53@asu.edu}
\and
Yi Ren\\
{\tt\small yiren@asu.edu}
\and
Yezhou Yang\\
{\tt\small yz.yang@asu.edu }
}
\maketitle

\begin{abstract}
This paper follows cognitive studies to investigate a graph representation for sketches, where the information of strokes, i.e., parts of a sketch, are encoded on vertices and information of inter-stroke on edges.
The resultant graph representation facilitates the training of a Graph Neural Networks for classification tasks, and achieves accuracy and robustness comparable to the state-of-the-art against translation and rotation attacks, as well as stronger attacks on graph vertices and topologies, i.e., modifications and addition of strokes, all without resorting to adversarial training.
Prior studies on sketches, e.g., graph transformers, encode control points of stroke on vertices, which are not invariant to spatial transformations. In contrary, we encode vertices and edges using pairwise distances among control points to achieve invariance. Compared with existing generative sketch model for one-shot classification~\cite{lake2015human}, our method does not rely on run-time statistical inference.
Lastly, the proposed representation enables generation of novel sketches that are structurally similar to while separable from the existing dataset. 
\end{abstract}


\section{Introduction}
Unlike the human vision system, it is well acknowledged that end-to-end deep learning methods lack intermediate representations that enable innate invariance to spatial translation and rotation~\cite{hinton2021represent, hinton1979some, bi2017,popsci2017,levesque2011winograd,clark2015elementary,newyorker2013,nytimes2017gm}. 

While such transformation invarianace can potentially be achieved through expensive robust (adversarial) training, it is believed that invariance (1) should be an innate property rather than an external model constraint, and (2) should not trade off recognition accuracy significantly. This motivates us to revisit the canonical computer vision perspective (such as object representation by components \cite{biederman1987recognition}, and local visual representation design \cite{lowe2004distinctive}) towards an explicit representation design for possessing innate properties. 

A commonly sought-after solution is to identify a part-whole structure~\cite{hinton2021represent}, following the insights of how the human vision system~\cite{hinton1979some, biederman1987recognition} parses scenes into atomic parts for recognition and generation, while both parts and the topologies of parts are invariant to spatial transformations.
The part-whole structure is also supported by the Gestalt principles~\cite{desolneux2007gestalt} and cognitive science~\cite{lake2017building, lake2012concept, lake2015human}.

Building on top of existing work and within the context of computer vision for sketch recognition, we present a part-whole representation where strokes, as parts, are connected as a graph to form a sketch. 
The focus on sketches draws inspiration from studies in biology and cognitive science~\cite{kubilius2016deep, landau1988importance, ha2017neural, ritter2017cognitive}. 
For example, \cite{landau1988importance, ritter2017cognitive} show that human vision relies more on shapes than on textures or colors. 
Studies also show that successful CNNs learn shape representations from natural images~\cite{geirhos2018imagenettrained, kriegeskorte2015deep, kubilius2016deep, hosseini2018assessing}. 


\begin{figure*}[thp]
    \centering
    \includegraphics[width=0.9\linewidth]{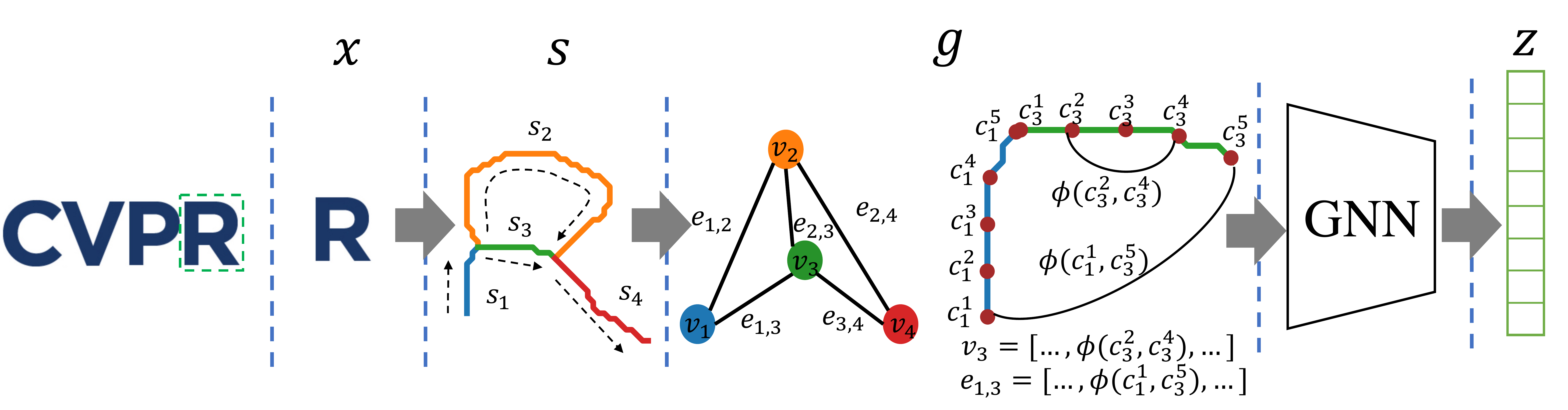}
    \caption{\textbf{An overview of SSR-GNNs.} We take ``R" for example. The image ``R" is composed of 4 strokes $\mathcal{S}(x) = [s_1, s_2, s_3, s_4]$. Each stroke is composed of 5 control points. $s_3 = [c_3^1, c_3^2, c_3^3, c_3^4, c_3^5]$. The 4 strokes are associates with 4 vertices $\mathcal{V} = [v_1, v_2, v_3, v_4]$ in graph $g(x)$. The value of vertex is the pairwise distance between control points. $v_3 = [\phi(c_3^p, x_3^q)]_{p=1, q=1}^{p=5, q=5}$. The value of edge is the pairwise distance between two connecting control point. $e_{1,3} = [\phi(c_1^p, c_3^q)]_{p=1, q=1}^{p=5, q=5}$. After passing to a learnable graph neural network(GNN), the $z$ is the high level representation of the image $x$. Dash arrow indicates the order of control point in a stroke from start point to end point.
    }
    \label{fig:overview}
\end{figure*}

Fig.~\ref{fig:overview} shows an example of the proposed representation: From an input image of ``R'', we adopt unsupervised image processing~\cite{lake2015human} to first identify fork points that separate strokes, and estimate control points of these strokes to form an undirected graph representation where each vertex contains stroke information, and edges specify interactions between strokes.
Specifically, to innately equip our representation with spatial invariance, each vertex encodes pairwise spatial distances between each pair of control points for the corresponding stroke (see $v_3$ in Fig.~\ref{fig:overview}), yielding an $n \times n$ matrix. If two strokes (each with $n$ control points) are connected, we then form an edge in the resulting graph. The edge encodes the pair-wise distance between each control point from one stroke to each of the other stroke (see $e_{1,3}$ in Fig.~\ref{fig:overview}). The distances between control points are invariant to spatial transformations, therefore our graph design is innately spatially-invariant.

To validate this representation, we take sketch-based classification/robust feature learning, and novel pattern generation as the testing tasks. To leverage the graph representation for classification, we train a Graph Neural Network (GNN)~\cite{scarselli2008graph}, which allows variable input graph topologies and preserves spatial invariance. 

We claim the following contributions:

\begin{itemize}
    \item Through extensive experiments on MNIST and two subsets of the Quickdraw dataset~\cite{ha2017neural}, we show that the proposed models are innately robust to rotations and translations, while maintaining acceptable classification accuracy.
    \item In addition, we show that the proposed models are robust to parametric and topological attacks without robust training, which suggests that stroke-based graphs are robust features for perception.
    \item Lastly, we show that the proposed models can be used to generate novel sketchs distinguishable from the training set. E.g., by learning to classify decimal digits, the model can then be used to generate hypothetical ``A''s to ``F''s for a hexadecimal system. This shows that models have strongly structured expression capability. 
\end{itemize}

\section{Related work}

\textbf{Part-whole representations:} 
Studies in cognitive science suggest the human vision system parse visual inputs into part-whole, which are invariant to spatial transformations and change of viewpoints~\cite{hinton2021represent, sabour2018matrix, singh2001part}.
For example, structural description models~\cite{biederman1987recognition, hummel1992dynamic, van2015part, kodratoff1984learning} combine the description of the part components. 
Along the same vein, sketches have been considered as a composition of strokes which are parts and sub-parts representation~\cite{lake2015human}. 
Such part-whole representations have been shown to play a critical role in enabling learning with small data. \cite{lake2015human} shows that one-shot classification/generation can be achieved on labeled graphs through iterative inference based on statistics of sampled graphs. 

\textbf{Graph neural network:} In graph neural networks (GNNs) ~\cite{duvenaud2015convolutional,bruna2013spectral}, vertex and edge information evolve through neighborhood aggregation. By sharing pairwise aggregation models, such as in message passing neural networks (MPNNs) ~\cite{gilmer2017neural,li2015gated}, GNNs are often shown to be generalizable to input graphs with arbitrary sizes. 
\sheng{They are widely used in the area of sketches~\cite{yang2020sketchgcn}, handwriting~\cite{riba2015handwritten} and math formula recognition~\cite{mahdavi2020visual}.}
In this work, we adopt an MPNNs architecture to handle sketches with variable number of strokes and topologies. 
\sheng{Graph-based representation~\cite{cho2013learning, wu2014learning} or graph neural network~\cite{gilmer2017neural,li2015gated} are also invariant to permutation. In this work, the high level representations are inputs transformation invariant because both stroke representation and graph neural network are invariant to transformation.}

\textbf{Sketch-based learning:} 
Sketch is an abstract visual input without the information of texture and color~\cite{xu2020deep}. Cognitive science research shows that human beings are able to grasp the major semantic meaning from an image directly from its sketch form~\cite{kriegeskorte2015deep}. 
Current studies typically primarily investigate sketch classification using existing deep neural networks.  
\sheng{CNNs based methods~\cite{yu2017sketch, song2017deep} are applied on raster sketch image. 
\shengc{RNNs based methods~\cite{xu2018sketchmate, jia2017sequential, he2017sketch, ha2017neural, xu2020learning}, as well as textual convolution network (TCN)~\cite{xu2020deep}, transformer (Sketchformer)~\cite{ribeiro2020sketchformer}, model sketch as sequence of control points or strokes.}
Graph based methods~\cite{yang2021sketchgnn, xu2021multigraph, yang2021sketchaa, yang2020sketchgcn} explore the topological information for sketch. Graph transformer~\cite{xu2021multigraph} encodes the control point as vertex and 1-hop, 2-hop, global hop connection as edge. SketchAA~\cite{yang2021sketchaa} learns the abstraction and hierarchy of the grid blocks of sketch image by encoding them as graph. SketchGNN~\cite{yang2021sketchgnn} learns the semantic segmentation of sketch. The vertex of SketchGNN is the single point on sketch and the edge is the single stroke that connecting two adjacent points.}
\shengc{Furthermore, current studies of sketch are extended to multimodality, such as sketch with video~\cite{xu2020fine, collomosse2009storyboard}, sketch with word, text, cartoon and natural image~\cite{xie2019deep}.} In this work, unlike the previous works which only focus on improving the classification accuracy only, we take sketch based classification as a testing task to validate that the newly proposed representation design is spatially robust. Especially in adversarial cases, our sketch classification maintains a high level of performance. 

\section{Methods}
Our method contains the following elements: (1) A pre-process step where an input image $x$ is converted to a set of strokes $\mathcal{S}(x)$. (2) The strokes and their connections are then encoded as a graph $g(x)$. Node and edge features are designed to achieve rotation and translation invariance. (3) A GNN is learned based on a labeled dataset $\{(g(x), y)\}$.
Fig.~\ref{fig:overview} summarizes the learning pipeline. Details are explained as follows.

\subsection{Acquiring Strokes from an Image} \label{sec:preprocess}
We decompose an input image $x \in \mathbb{R}^{d_x}$ into a set of strokes, denoted by $\mathcal{S}(x)$, where $|\mathcal{S}(x)|$ varies by image.
We follow the preprocess procedures of \cite{lake2015human}, which include thinning the image~\cite{lam1992thinning}, detecting fork points ~\cite{liu1999identification}, and finally merging the noisy and redundant fork points by the maximum circle criterion ~\cite{liao1990stroke}. The ordering of the strokes, i.e., the definition of the start and end points of strokes, is then derived from a walk throughout the fork points which follows the rules moving from left to right and from top to bottom. Fig.~\ref{fig:overview} demonstrates the procedure for ``R''.

Building upon \cite{lake2015human}, each stroke is first approximated as a uniform cubic b-spline parameterized by the control points, the offset, and the scaling factor. We then sample $n$ control points on stroke by re-fitting the uniform cubic b-spline. The number of control points are set to be the same for all strokes and tuned for each experiment.  Unlike \cite{lake2015human} which takes a set of uniform cubic b-spline control points, we set aside the offset for enforcing boundary constraints, and the scale factor for accurately computing the pairwise distances. We represent a stroke $s_i$ using its $n$ control points: $s_i = [c_i^{p}]_{p=1}^{n}$. 

\subsection{Stroke-based Graph Representation}

We further convert the set of strokes into a graph $g(x) = (\mathcal{V}, \mathcal{E})$, where each vertex $v_i \in \mathcal{V}$ corresponds to a stroke $s_{i} \in \mathcal{S}(x)$, and an edge $e_{i,j} \in \mathcal{E}$ exists when the start (end) point of $s_i$ is the end (start) point of $s_j$. 
To achieve rotation and translation invariance, we assign each vertex $v_i$ the set of pairwise Euclidean distances between all sample points from $s_i$. 
In particular, $v_i = [\phi(c_i^{p}, c_i^{q})]_{p=1, q=1}^{p=n, q=n}$ where $c_i^{p}$ is the $p$th control point of the $i$th stroke and $\phi(\cdot,\cdot)$ is the Euclidean distance. Similarly, we assign each edge $e_{i,j}$ the set of pairwise Euclidean distances from all control points from $s_i$ to those from $s_j$: $e_{i,j} = [\phi(c_i^{p}, c_j^{q})]_{p=1, q=1}^{p=n, q=n}$. We note that the ordering of elements in $v_i$ and $e_{i,j}$ is defined based on the start and end points of $s_i$ and $s_j$, i.e., switching the start and end points of $s_i$ will change $v_i$ and $e_{i,j}$. 

\subsection{GNN architecture}\label{sec:mpnn}

We adopt MPNNs to handle the variable graph sizes encountered in MNIST and QuickDraw datasets. 
The MPNN contains three components: (M) message passing, (U) update, and (R) readout ~\cite{battaglia2018relational}, which are defined as: 
\begin{equation*}
\begin{aligned}
    &(M) \quad & m_{v_i}^{(t+1)} &= \sum_{s_i, s_j \text{are connected}} M_t(v_i^{(t)}, v_j^{(t)}, e_{i,j}^{(t)}), \\
    &(U) \quad & v_i^{(t+1)} &= U_t(v_i^{(t)}, m_{v_i}^{(t+1)}), \\
    &(R) \quad & z &= R(v_i^{(T)}| v_i^{(T)} \in v).
\end{aligned}
\end{equation*}
The message passing and update phase execute $T$ times. The message at phase $t+1$, $m_{v_i}^{(t+1)}$, is encoded by vertex $v_i^{(t)}$, adjacent vertices $v_j^{(t)}$s and edges $e_{i,j}^{(t)}$ at step $t$. The new vertex $v_i^{(t+1)}$ is updated by current vertex $v_i^{(t)}$ and message $m_{v_i}^{(t+1)}$. After $T$ steps, the feature $z$ is computed from a readout function. 
$M_t(\cdot,\cdot,\cdot)$, $U_t(\cdot,\cdot)$, $R(\cdot)$ are learnable components. $z$ is input to a linear classifier with softmax outputs. To simplify the notation, we define the MPNN as $f(\cdot)$ and linear classifier as $f_{\theta}(\cdot)$.


\subsection{Learning Objectives}

Here we introduce the formulations of the learning problems for the two experiments to be discussed in Sec.~\ref{sec:exp}. 

\paragraph{Classification} 
\label{sec:obj1}
We use a standard cross-entropy loss for learning a classifier $f_{\theta} \circ f$. Given a dataset $\mathcal{D}:=\{(x,y)\}$, the loss is:
\begin{equation}
        \min_{f, f_{\theta}} \quad  \mathbb{E}_{(x,y)\sim \mathcal{D}} 
        \left[
        \psi(y,f_\theta \circ f(g(x)))
        \right], 
    \label{eq:cls}
\end{equation}
where $\psi(\cdot, \cdot)$ is the cross-entropy.

\begin{figure*}
    \centering
    \begin{adjustbox}{width=0.58\linewidth}
        \begin{tabular}{ccc}
            \begin{adjustbox}{width=0.3\linewidth}
                \begin{tabular}{ccc}
                    \includegraphics[width=0.1\linewidth]{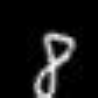} &
                    \includegraphics[width=0.1\linewidth]{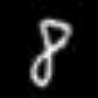} &
                    \includegraphics[width=0.1\linewidth]{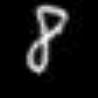} \\
                    \includegraphics[width=0.1\linewidth]{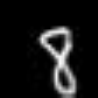} &
                    \includegraphics[width=0.1\linewidth]{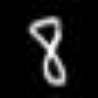} &
                    \includegraphics[width=0.1\linewidth]{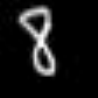} \\
                    \includegraphics[width=0.1\linewidth]{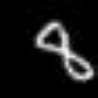} &
                    \includegraphics[width=0.1\linewidth]{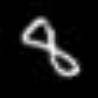} &
                    \includegraphics[width=0.1\linewidth]{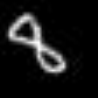} 
                \end{tabular} 
             \end{adjustbox}
             &
             \begin{adjustbox}{width=0.3\linewidth}
                \begin{tabular}{ccc}
                    \includegraphics[width=0.1\linewidth]{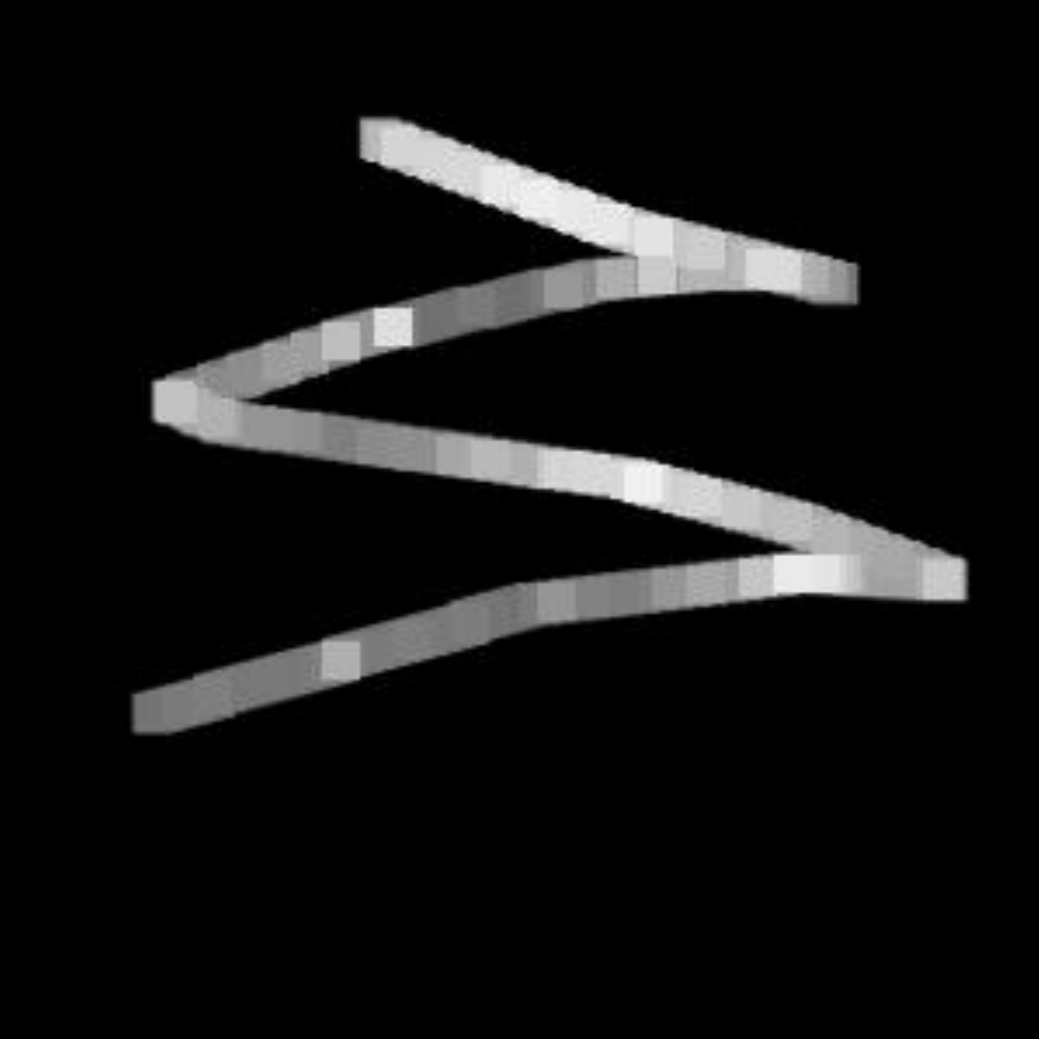} &
                    \includegraphics[width=0.1\linewidth]{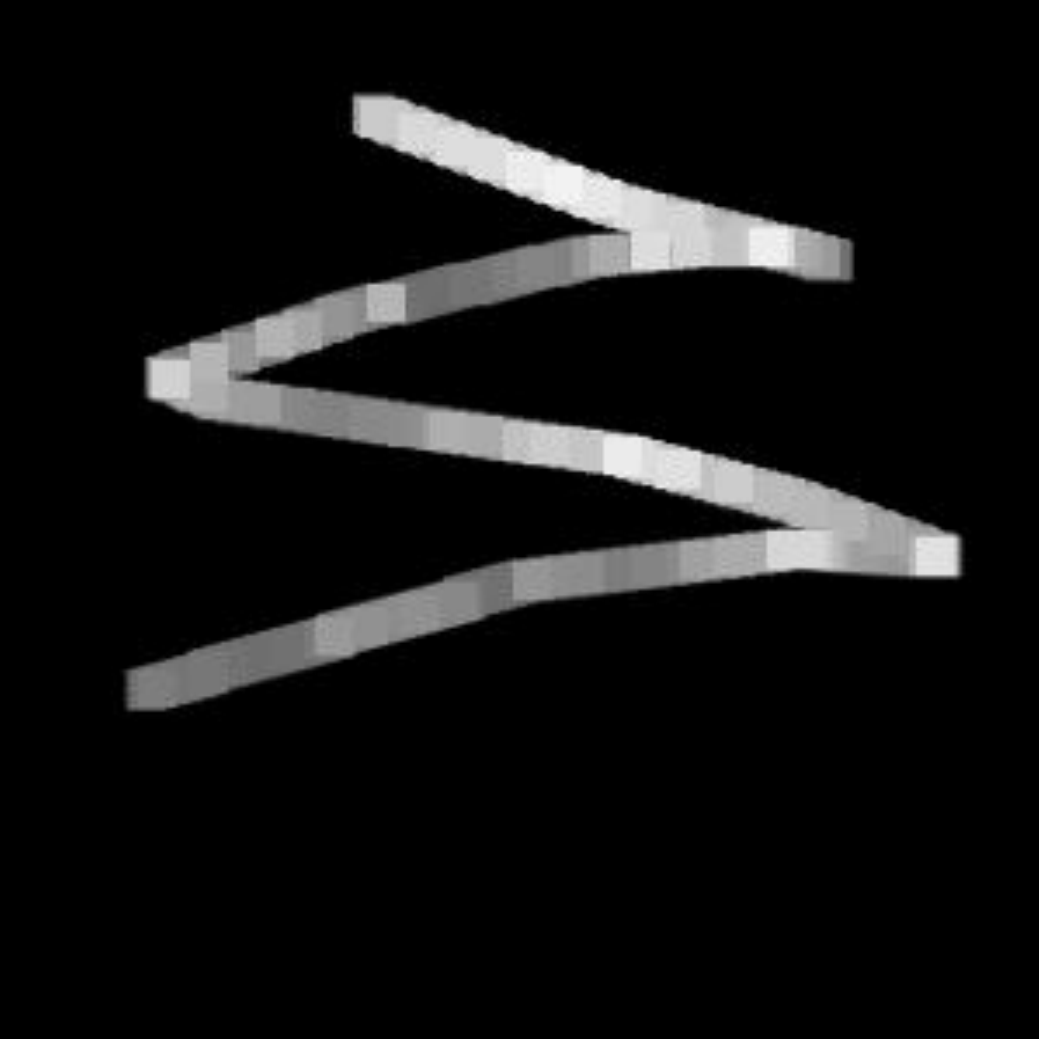} &
                    \includegraphics[width=0.1\linewidth]{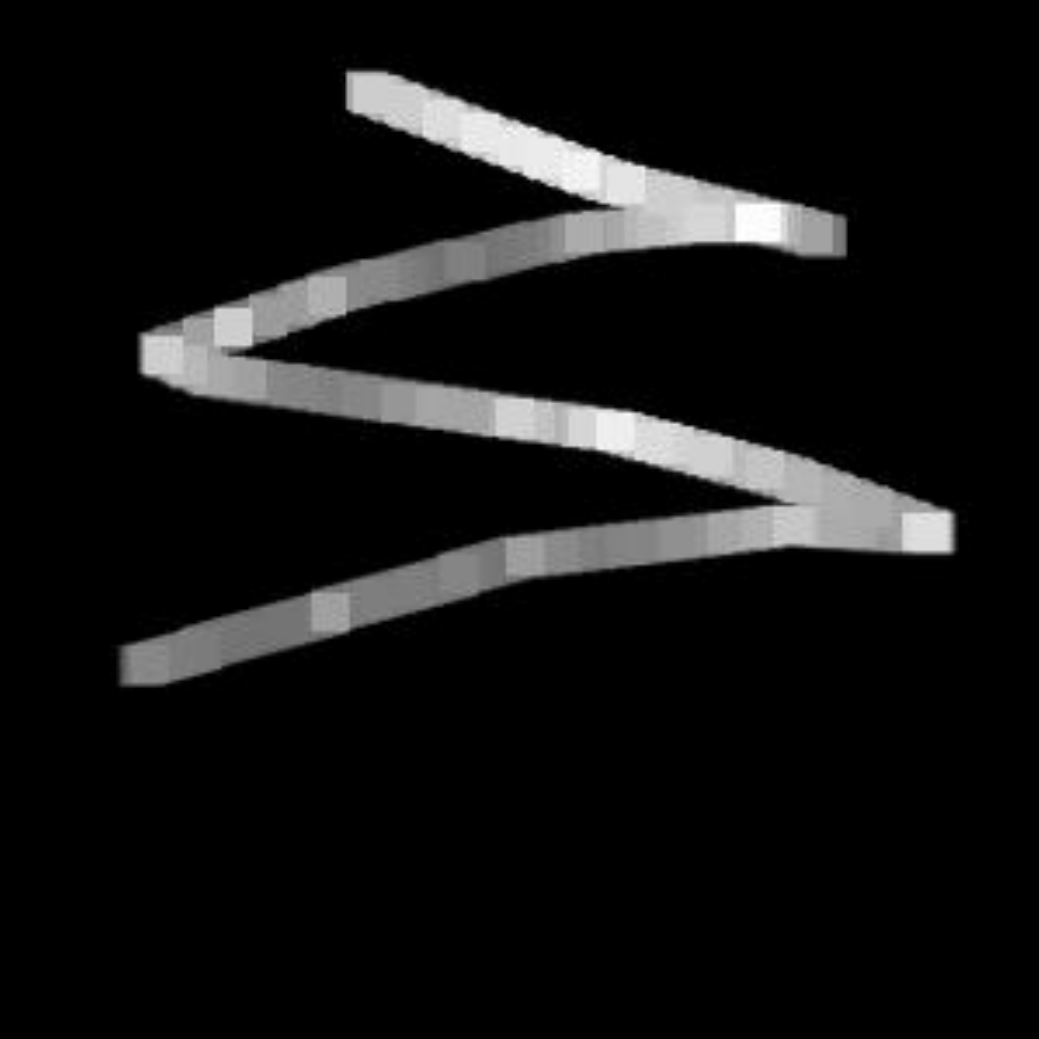} \\
                    \includegraphics[width=0.1\linewidth]{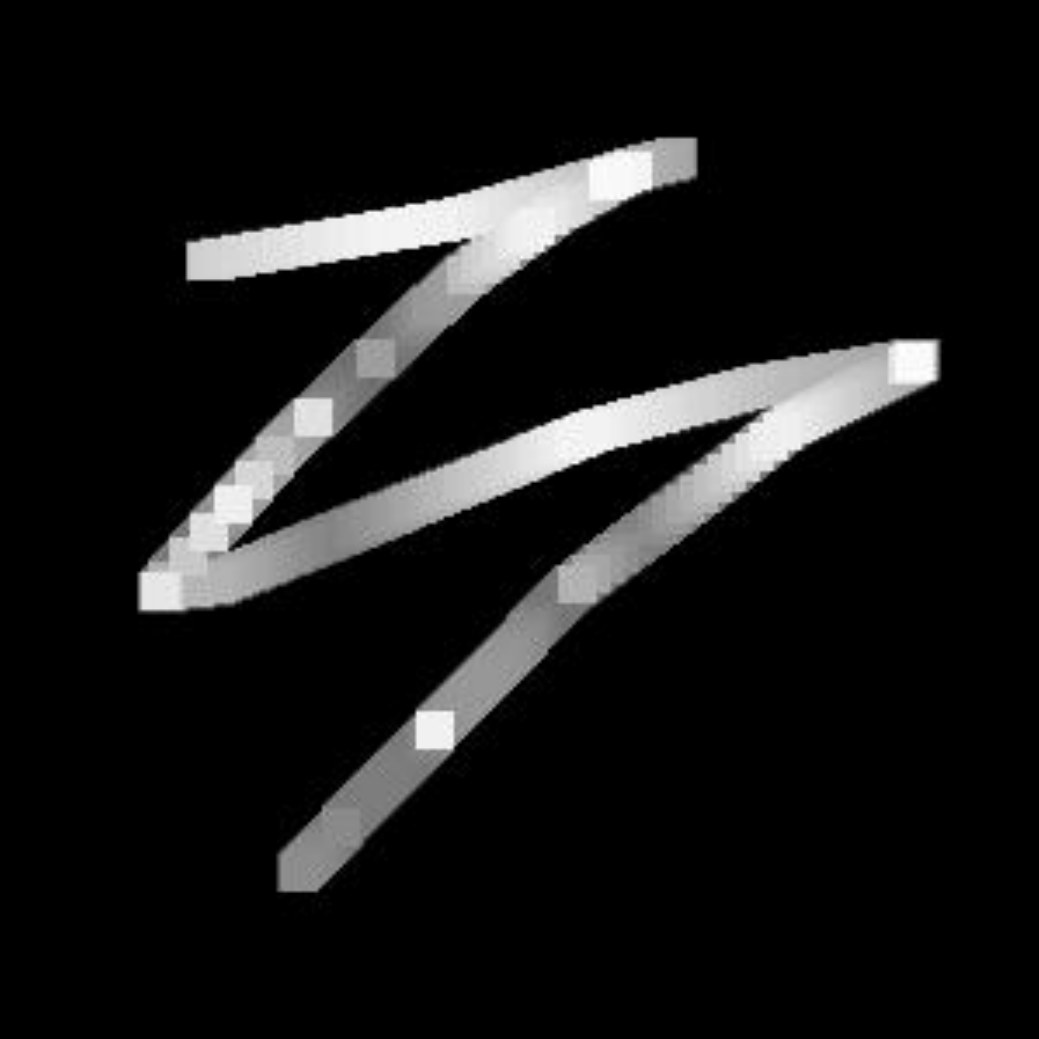} &
                    \includegraphics[width=0.1\linewidth]{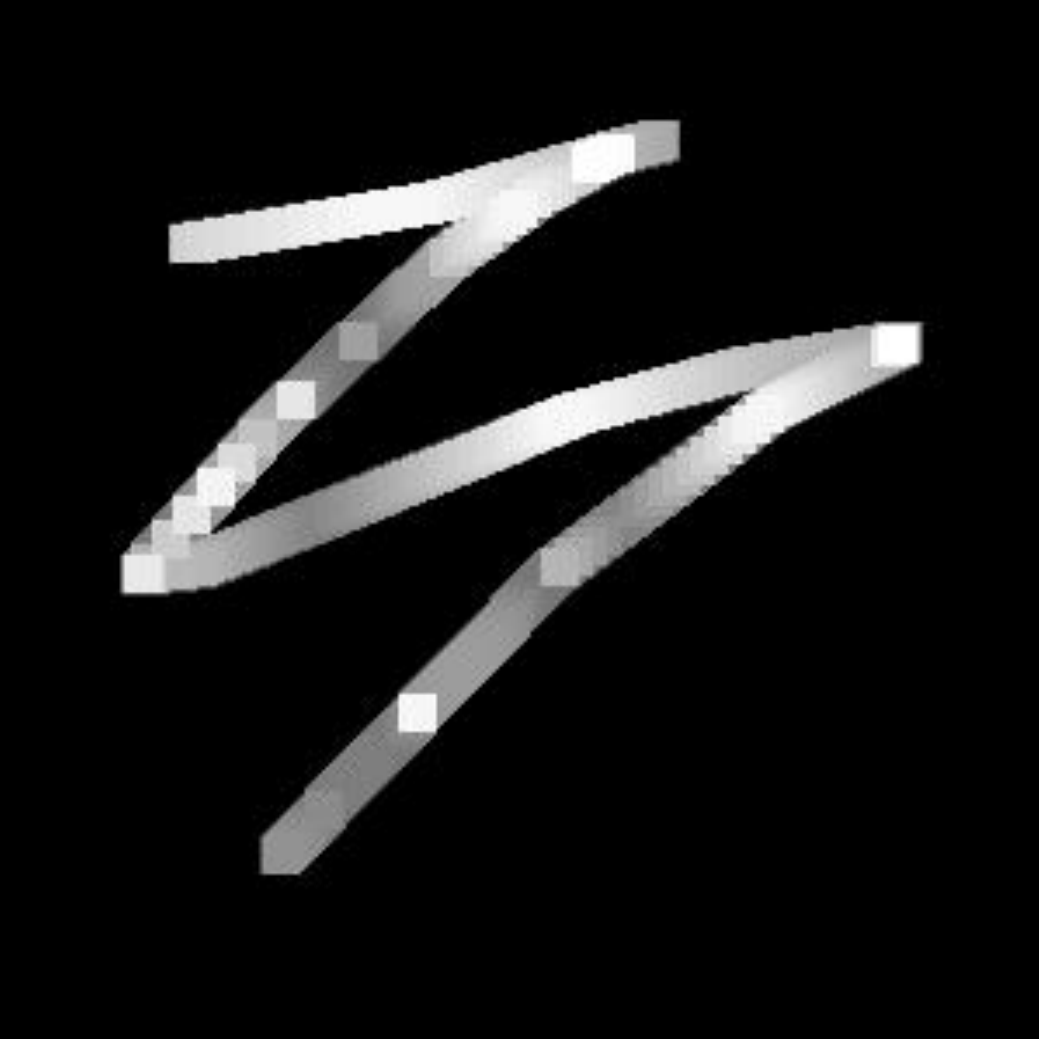} &
                    \includegraphics[width=0.1\linewidth]{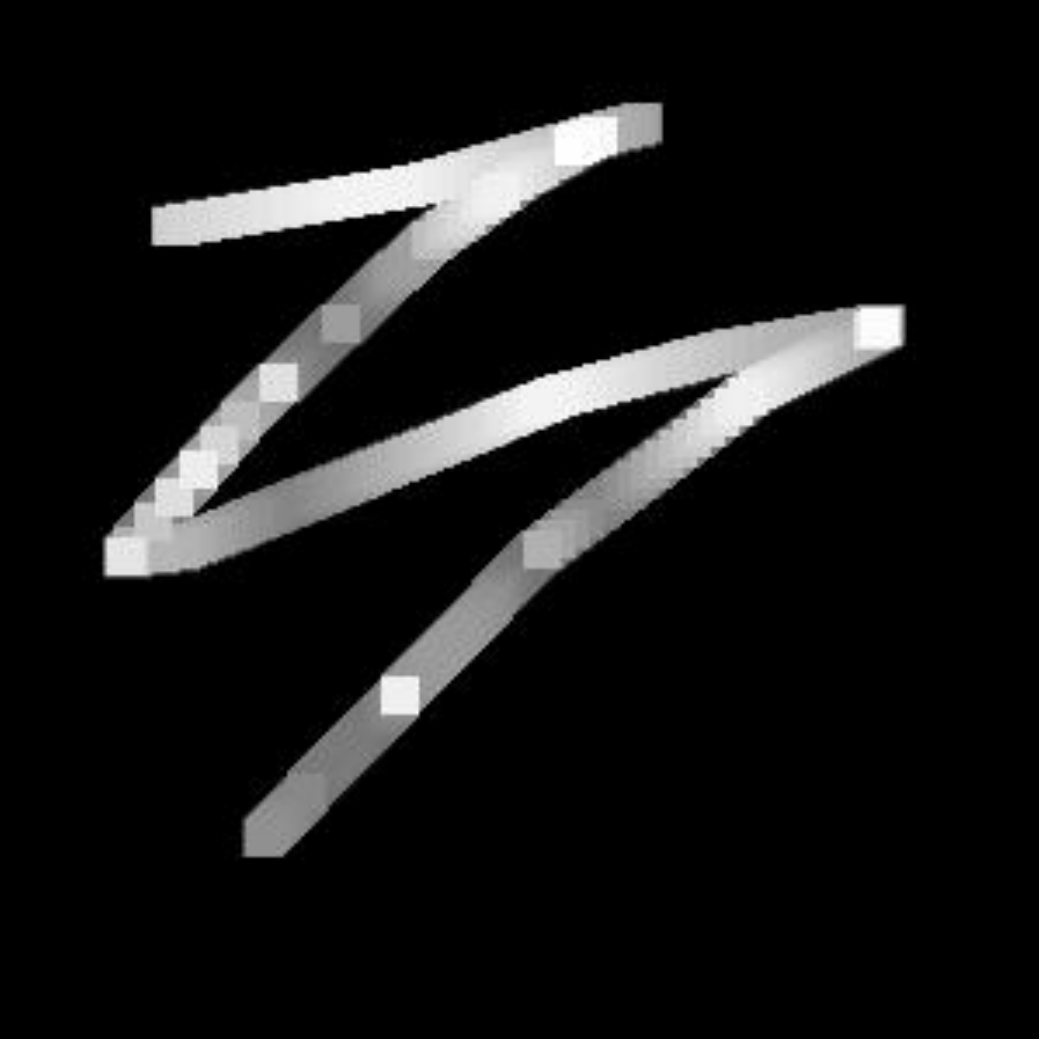} \\
                    \includegraphics[width=0.1\linewidth]{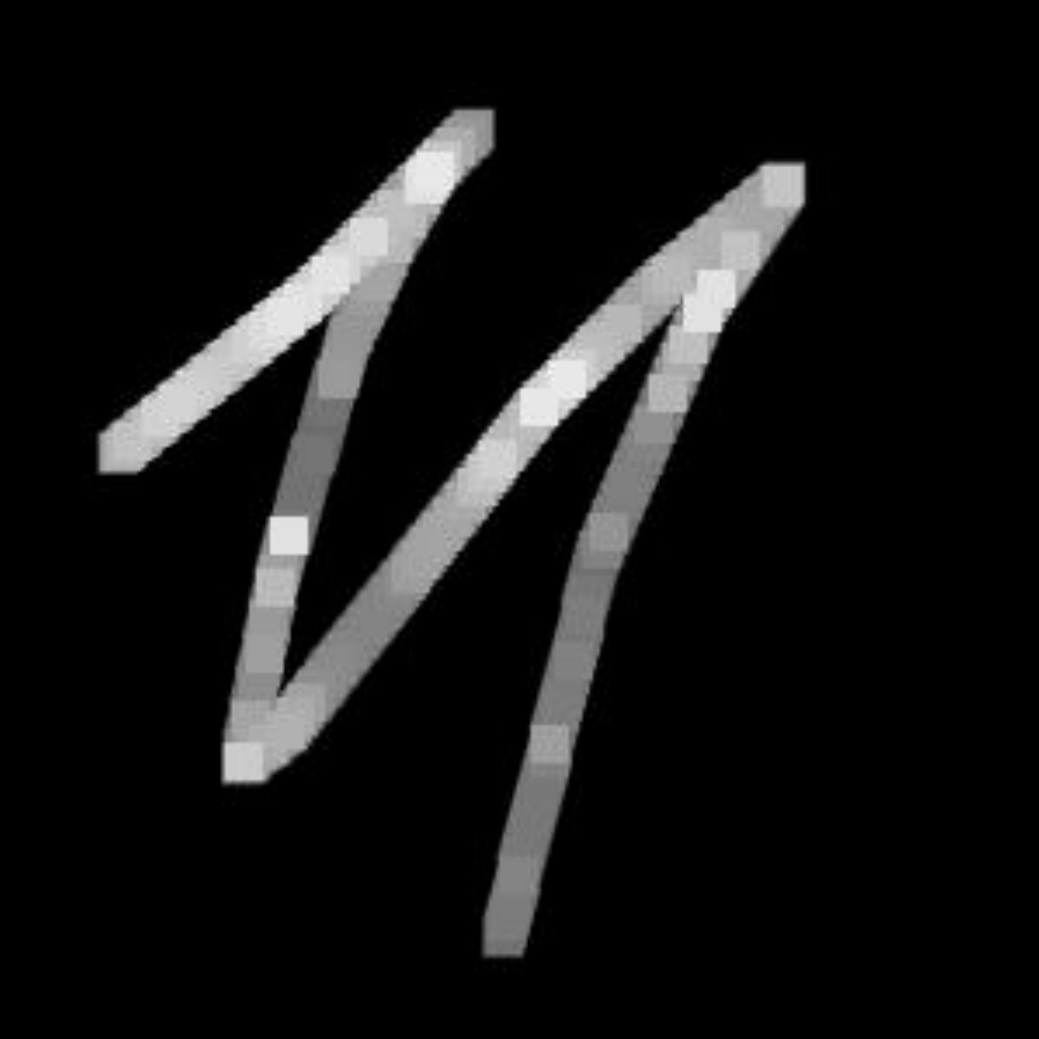} &
                    \includegraphics[width=0.1\linewidth]{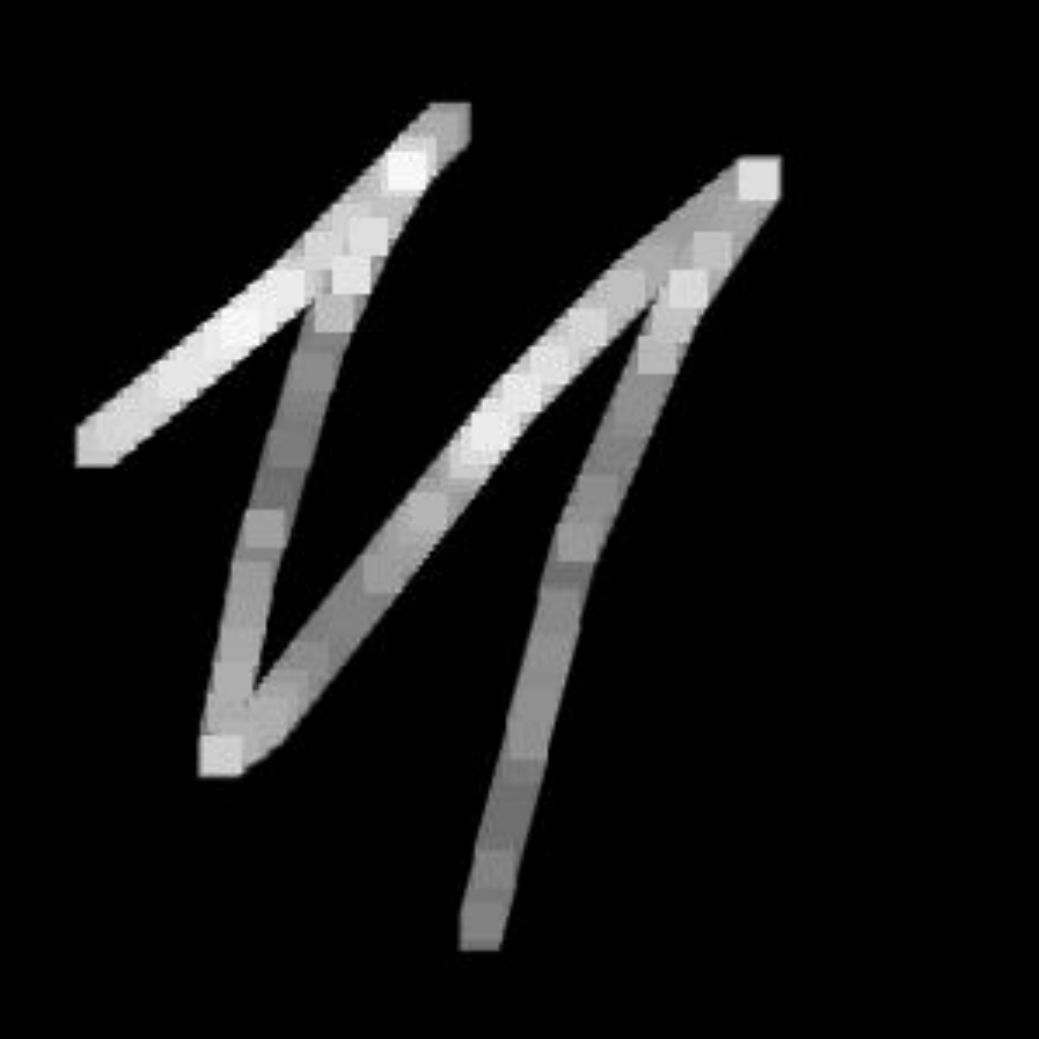} &
                    \includegraphics[width=0.1\linewidth]{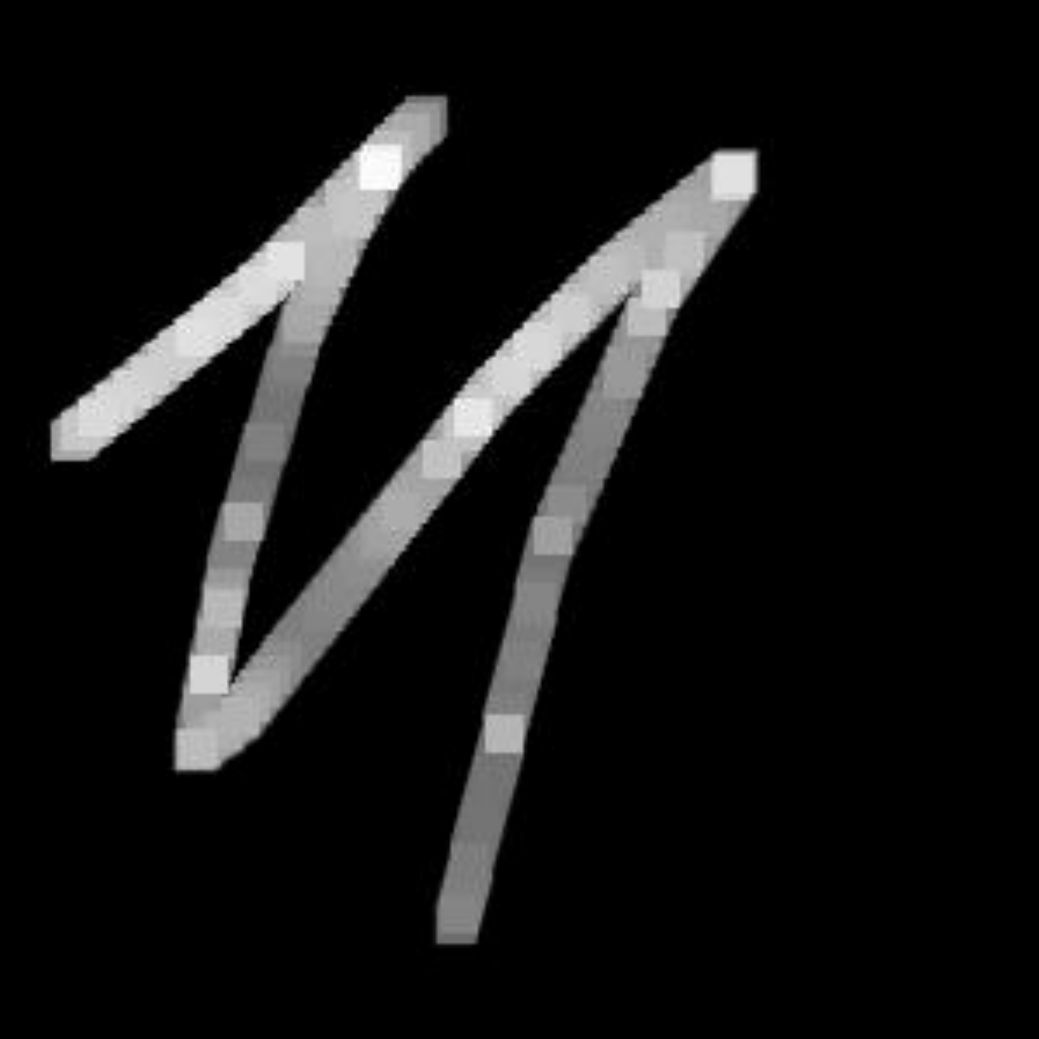}
                \end{tabular} 
             \end{adjustbox}
             &
             \begin{adjustbox}{width=0.3\linewidth}
                \begin{tabular}{ccc}
                    \includegraphics[width=0.1\linewidth]{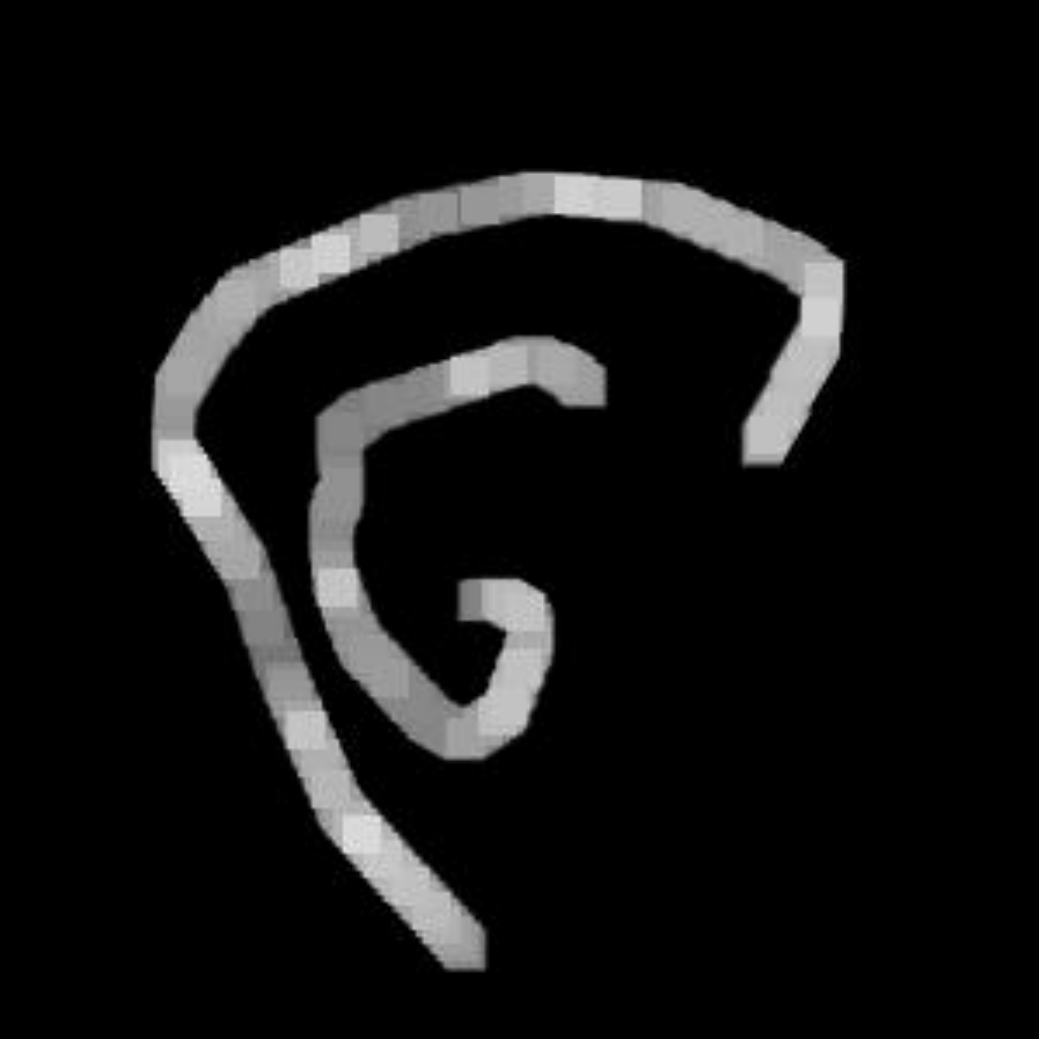} &
                    \includegraphics[width=0.1\linewidth]{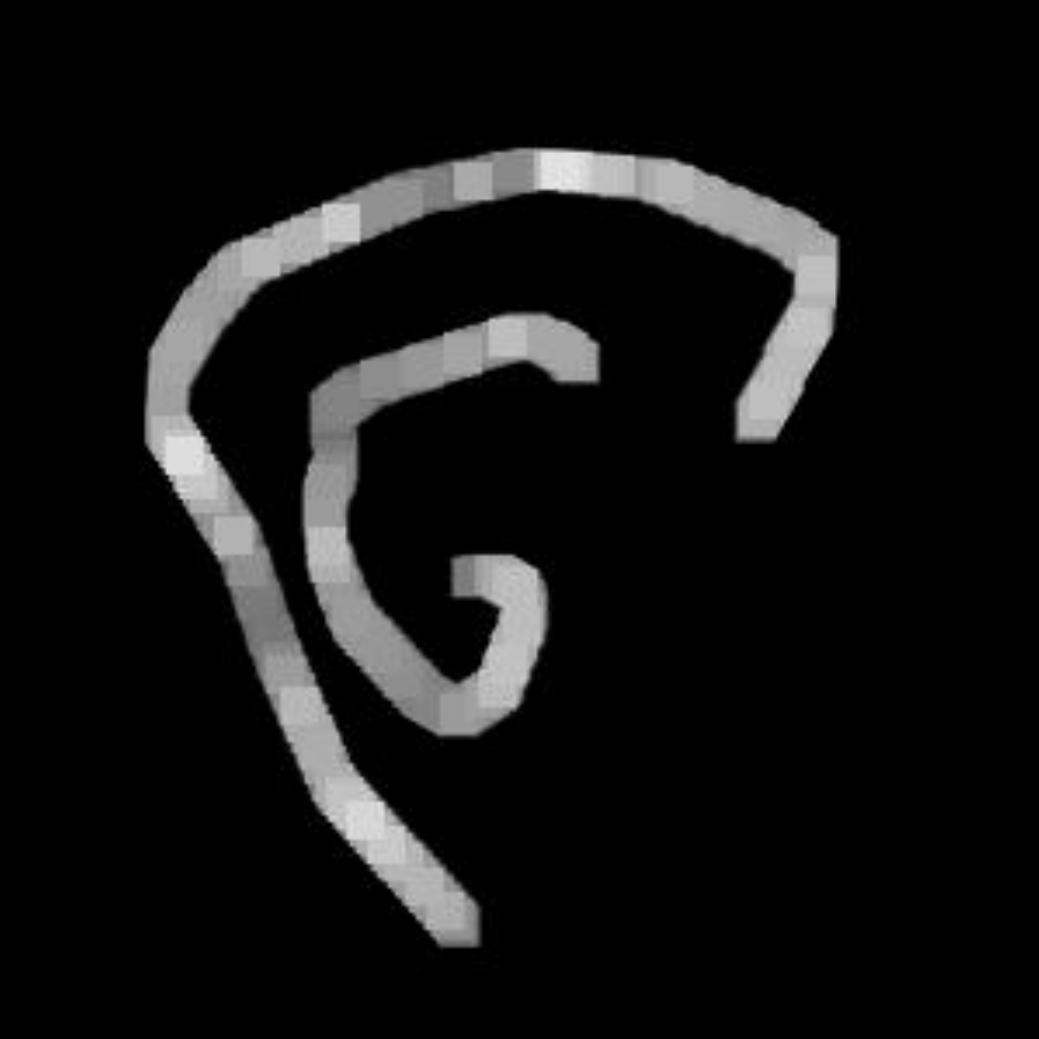} &
                    \includegraphics[width=0.1\linewidth]{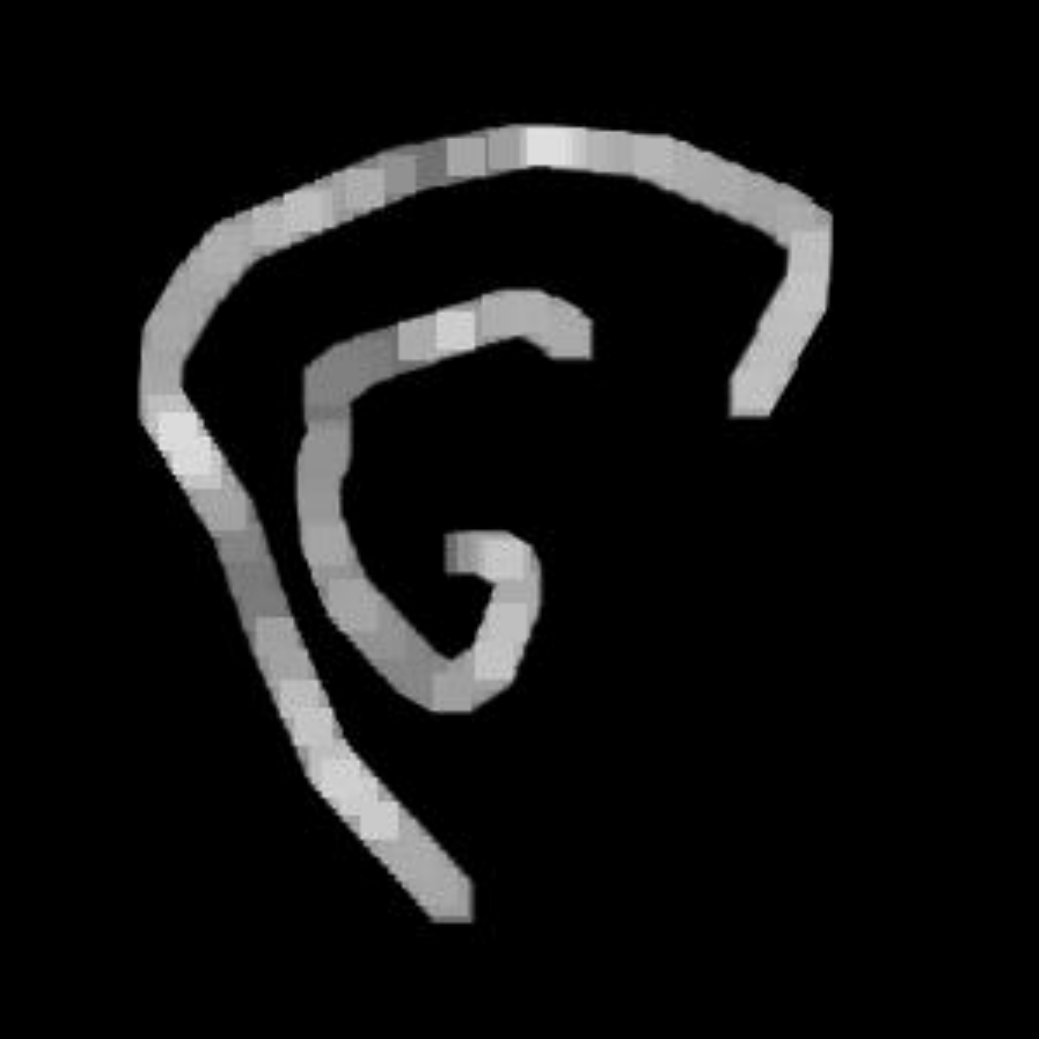} \\
                    \includegraphics[width=0.1\linewidth]{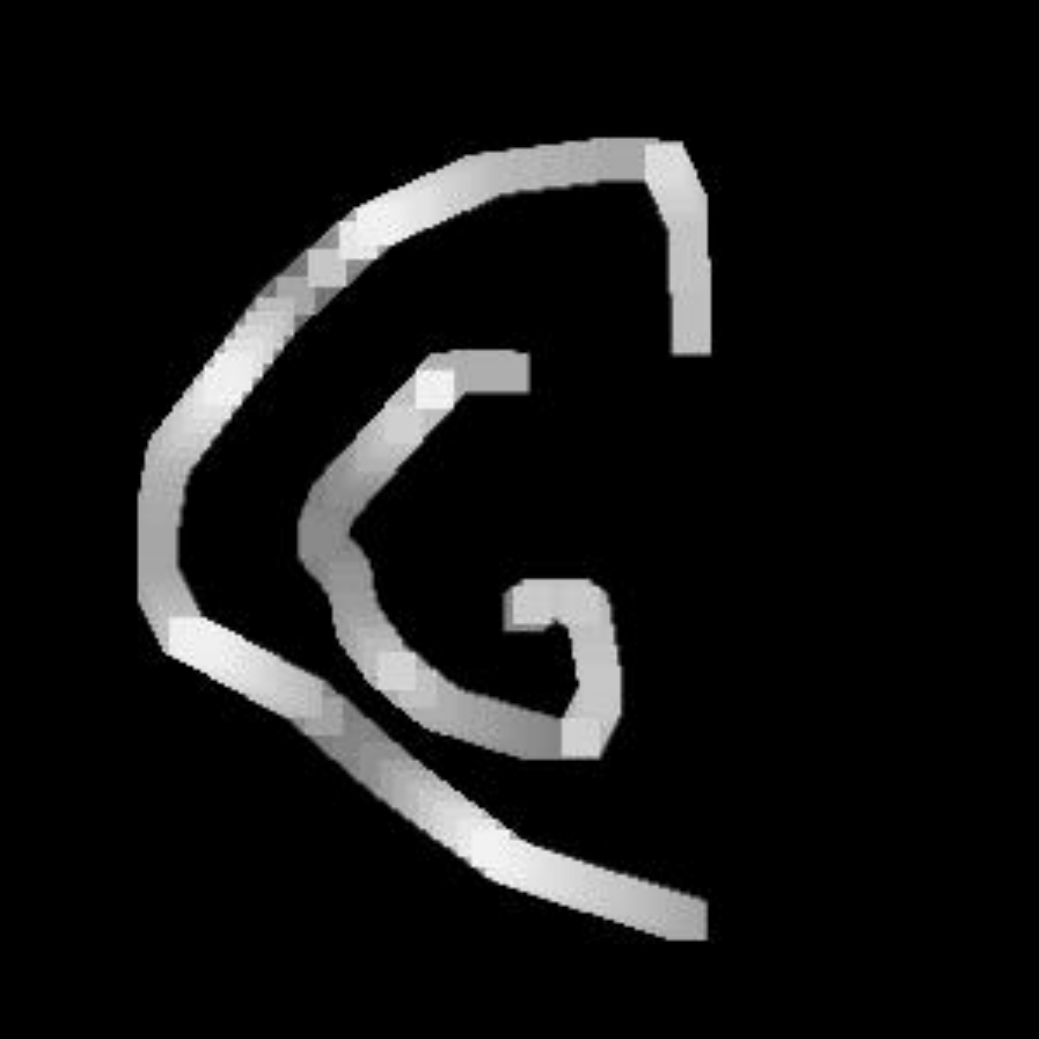} &
                    \includegraphics[width=0.1\linewidth]{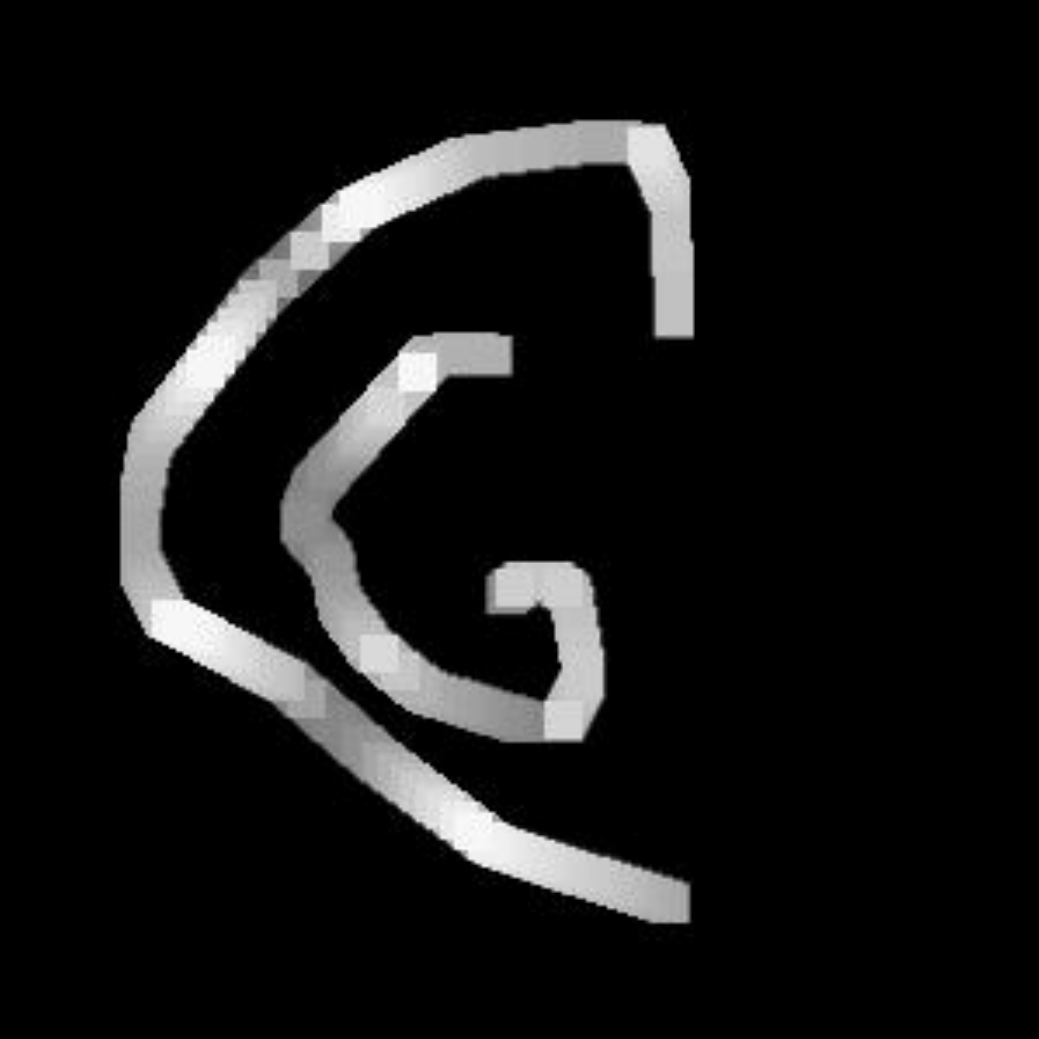} &
                    \includegraphics[width=0.1\linewidth]{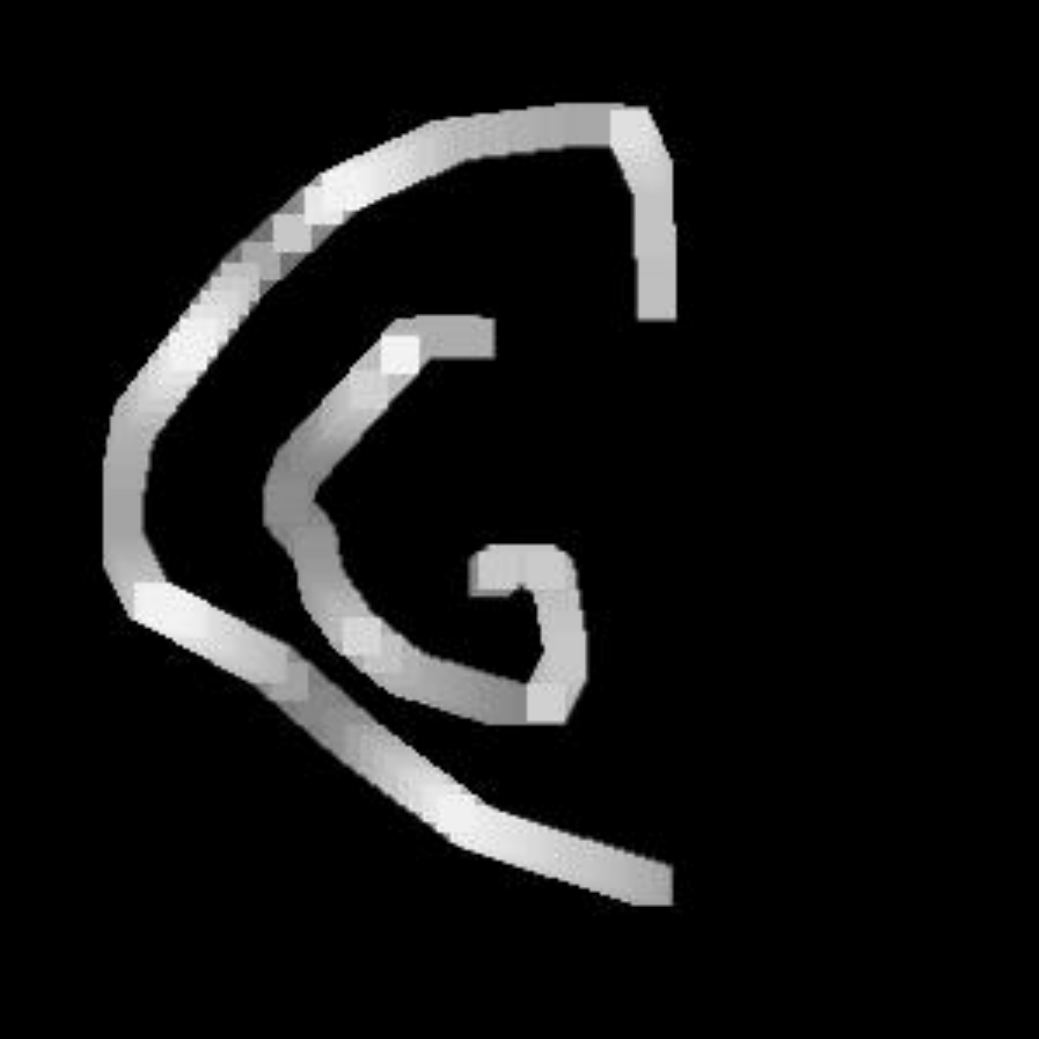} \\
                    \includegraphics[width=0.1\linewidth]{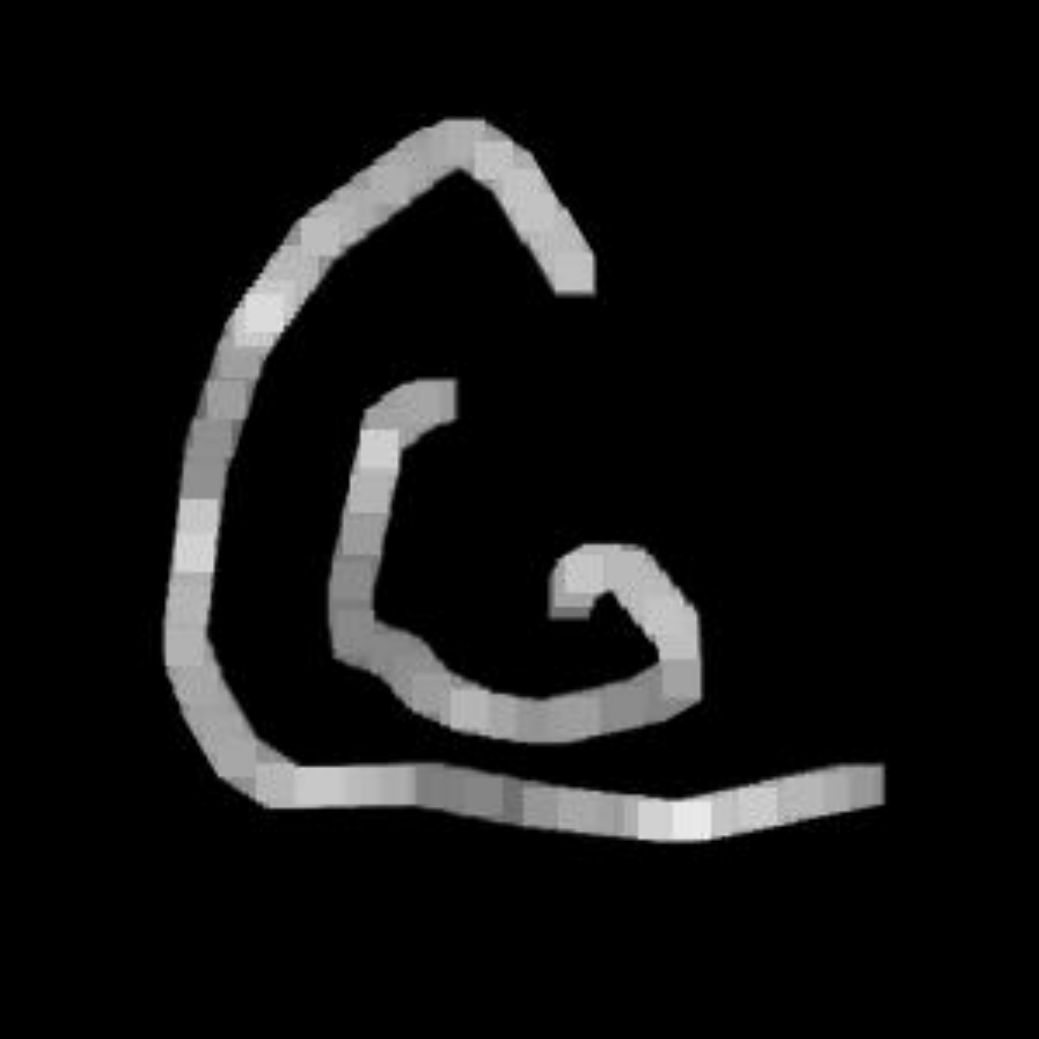} &
                    \includegraphics[width=0.1\linewidth]{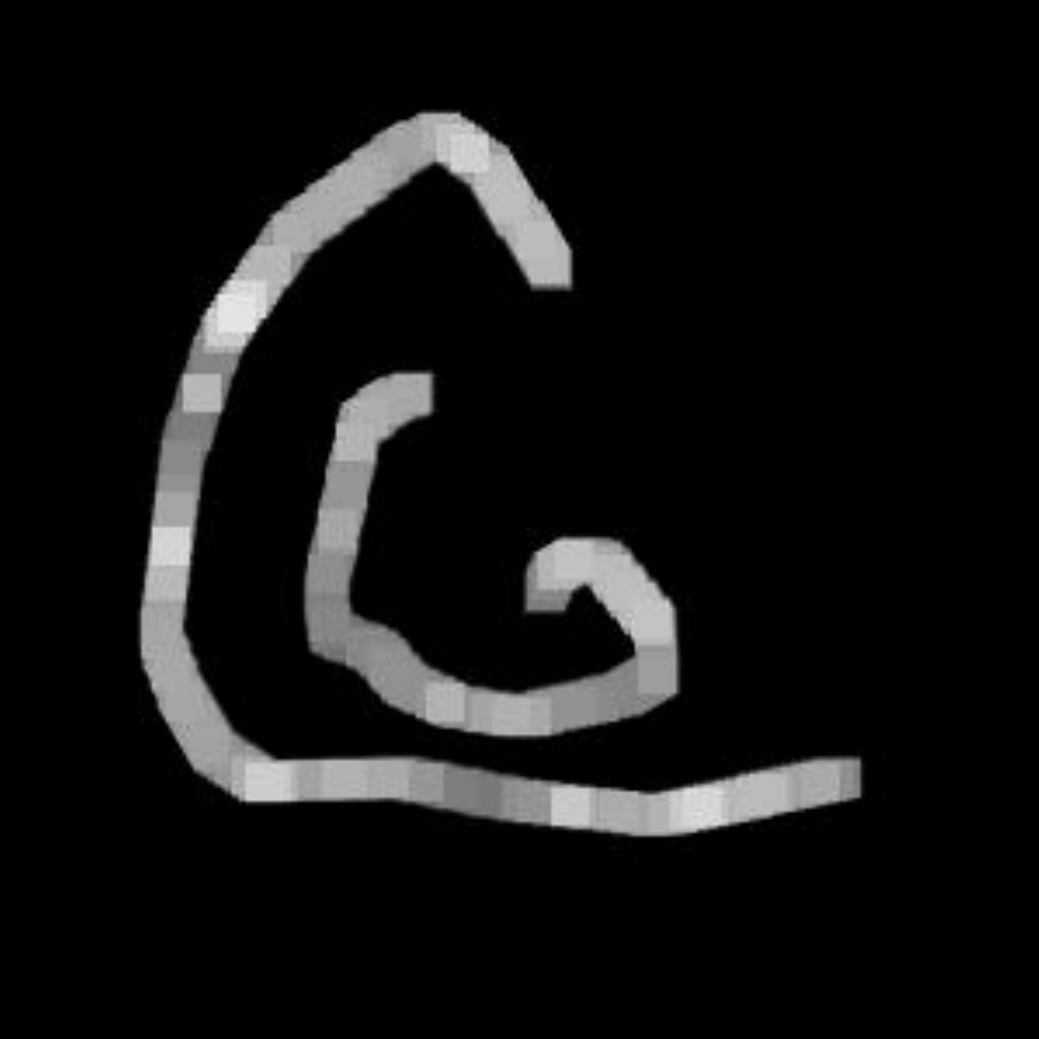} &
                    \includegraphics[width=0.1\linewidth]{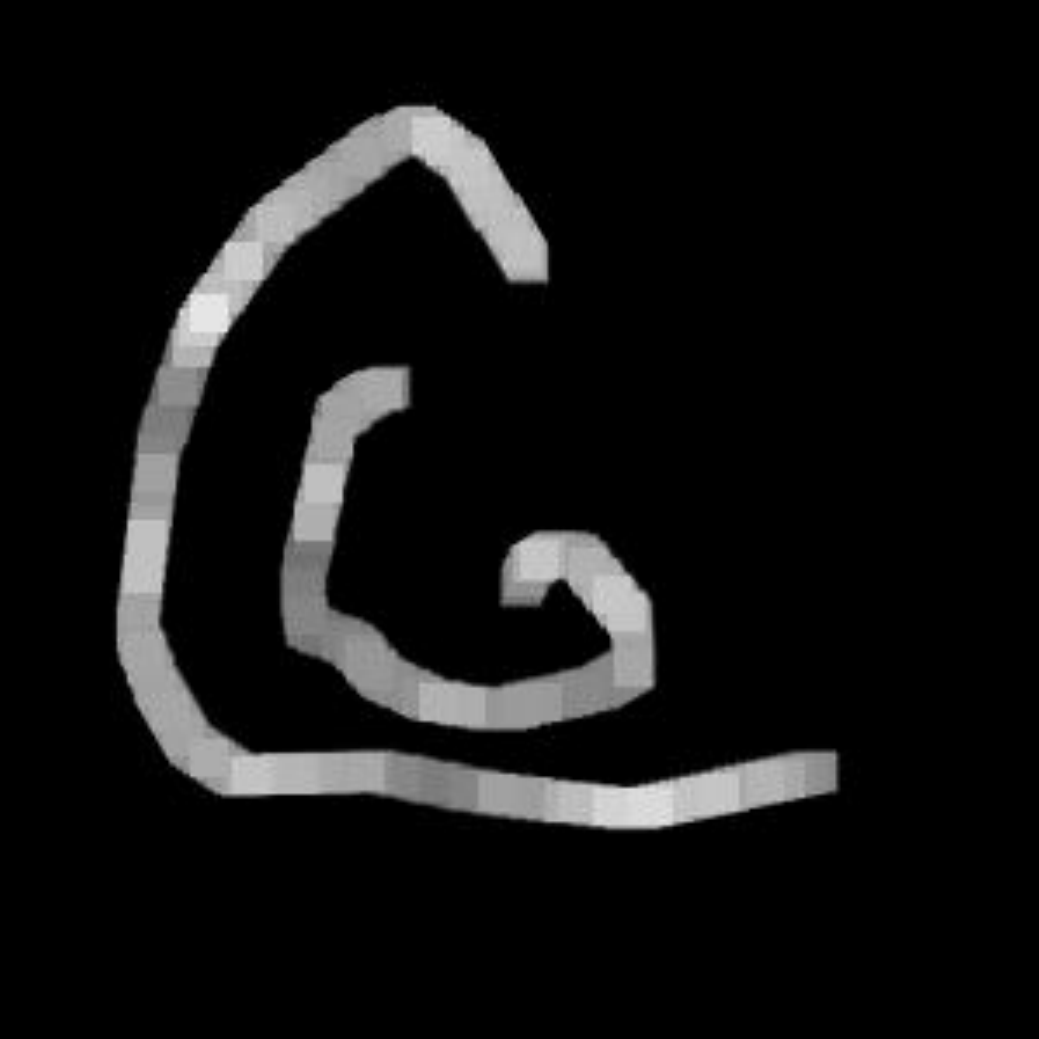} 
                \end{tabular} 
             \end{adjustbox}
        \\
        \LARGE(a) MNIST & \LARGE (b) Quickdraw(Shape) & \LARGE (c) Quickdraw(Body)
        \end{tabular}
    \end{adjustbox}
    \caption{\textbf{Spatial transformation on MNIST, Quickdraw(shape), Quickdraw(body).} The grid of each row is the rotation of $-\theta, 0, \theta$. The grid of each column is the translation of $(-\delta_x, -\delta_y), (0, 0), (\delta_x, \delta_y)$. For MNIST, $\theta = 30^\circ, \delta_x = \delta_y = 3\text{px}$. For Quickdraw, $\theta = 30^\circ, \delta_x = \delta_y = 10\text{px}$. }
    \label{fig:sample_attack}
\end{figure*}   

\paragraph{Sketch modification and generation} \label{sec:obj2}
In experiments, we will demonstrate the robustness of the classifier $f \circ f_{\theta}$ against targeted attacks via vertex-wise and topology-wise perturbations. For topology-wise attacks, we specifically consider adding a stroke $s$ to an existing sketch. In both cases, we let the target label be a one-hot vector $y$, the set of control points of the stroke of interest be $\mathcal{C}$ and the resultant graph be $g(\mathcal{C})$. The attacks solve:
\begin{equation}
        \min_{\mathcal{C}} \quad \psi(y,f_\theta \circ f(g(\mathcal{C}))) + L(g(\mathcal{C})),
    \label{eq:attack}
\end{equation}
where $L(\cdot)$, as explained below, constrains the control points to be structurally similar to the training data. In addition, we also incorporate boundary constraints on $\mathcal{C}$ to limit the generated sketches within the boundaries of the image. 

\textbf{Penalty on sketch structure:} We consider two types of penalties on a generated graph $g=(\mathcal{V},\mathcal{E})$ to regulate its structure. First, the generated sketches should have pairwise distances between control points similar to those from the training data.
Second, the angles between neighbouring pairs of control points, denoted by $r\in \mathcal{R}$, should also be similar to those from the data. We incorporate these requirements through the following penalty:
\begin{equation}
\begin{aligned}
    & L(g) =  \mathbb{E}_{(v,e,r) \in (\mathcal{V},\mathcal{E},\mathcal{R})} \\
    & \left[-\lambda_1\log p(v;\mathcal{D})-\lambda_2\log p(e;\mathcal{D})-\lambda_3\log p(r;\mathcal{D})\right],
    \label{eq:discon}
\end{aligned}
\end{equation}
where $p(v;\mathcal{D})$ and $p(e;\mathcal{D})$ are the empirical distributions of the pairwise distances among control points within a stroke and between strokes from the dataset, respectively, and, $p(r;\mathcal{D})$ is that of the angles between neighbouring pairs of control points. $\lambda_1 = \lambda_2 = 10^{-5}$ and $\lambda_3 = 0.1$ are tuned to allow the attack loss to dominate.

We will also demonstrate the utility of the proposed model at generating new sketches that are structurally similar to the training data, yet semantically different. To do so, we first train a one-class classifier $f_1(\cdot)$ so that all training data belongs to the same group, i.e., $f_1(f(g(x)))>0$ for all $x \in \mathcal{D}$. Given a sketch $g(\mathcal{C})$ parameterized by the set of control points to be tuned, we solve the following problem
\begin{equation}
        \min_{\mathcal{C}} \quad \{ \max\{0, 1+f_1(f(g(\mathcal{C})))\} +  L(g(\mathcal{C})) \}, 
    \label{eq:gen}
\end{equation}
The hinge loss used here aims to push the generated sketch out of the training set.


\begin{table*}[!h]
\centering
\small
    \begin{tabular}{c|c|c|c|c}
        Method & \multicolumn{1}{c|}{Evaluation} & \multicolumn{1}{c|}{MNIST} & \multicolumn{1}{c|}{Quickdraw (Shape)} & Quickdraw (Body) \\ \hline
        
        \multirow{3}{*}{CNNs~\cite{engstrom2019exploring, xu2021multigraph}} & Accuracy &  99.31\% & 87.14\% & 80.10\% \\
         & Spatial Robustness & 26.02\% & 21.90\% & 31.10\% \\ 
         & Parameter Size & 600,810 & 25,315,474 & 25,315,474 \\ 
         \hline

        \multirow{3}{*}{RNNs~\cite{xu2021multigraph}} & Accuracy & - & 75.43\% & 68.30\% \\
         & Spatial Robustness & - & 0.00\% & 0.00\% \\ 
         & Parameter Size & - & 5,724,249 & 5,724,249 \\ 
         \hline

        \multirow{3}{*}{Graph Transformer~\cite{xu2021multigraph}} & Accuracy & - & 80.71\% & 75.4\% \\
         & Spatial Robustness & - & 0.10\% & 6.57\% \\ 
         & Parameter Size & - & 39,984,729 & 39,984,729 \\ 
         \hline
         
         \multirow{3}{*}{Ours} & Accuracy & 93.01\% & 73.00\% & 64.20\% \\
         & Spatial Robustness & 93.01\% & 73.00\% & 64.20\% \\ 
         & Parameter Size & 546,634 & 8,707,868 & 8,707,868 \\ 
    \end{tabular}
    \caption{\textbf{The accuracy and spatial robustness on three dataset (MNIST, Quickdraw (Shape), Quickdraw (Body)).} We compare our method with CNNs (Inception-V3 for Quickdraw), RNNs, and graph transformer. Our evaluation metrics are accuracy, spatial robustness and the parameter size.}
    \label{table:robustness}
\end{table*}

\section{Experiments}
\label{sec:exp}

We conduct two sets of experiments. First, we empirically show that the proposed model is robust to rotation and translation on classification tasks for MNIST and a subset for QuickDraw, while maintaining accuracy comparable to the SOTA, all without adversarial training. We also evaluate model robustness against vertex-wise and topology-wise attacks specific to graph inputs. 
Second, we show that our model is capable of generating novel sketches that are semantically different from the training set. Specifically, we demonstrate the generation of hypothetical digits that are separable from MNIST digits in the feature space.

\subsection{Classification and Robustness} \label{sec:robustness}
\paragraph{Dataset and pre-processing}
We use two standard datasets: MNIST and Google Quickdraw~\cite{ha2017neural}. MNIST is a hand-written digit dataset containing numerical digits from $0$-$9$. 
Google Quickdraw is a human hand-drawn sketch dataset with $345$ different categories, ranging from The Great Wall, airplane, to hands, squares, and dogs. 
For both datasets, we consider each sketch as an image, shifted to the top-left corner and normalized to $224\times 224$ pixels. It should be noted that the Quickdraw dataset also stores key points of simplified strokes in temporal order~\cite{xu2021multigraph, xu2018sketchmate}, computed by the Ramer-Douglas-Peucker algorithm ~\cite{ramer1972iterative}. This format has been used by existing graph-based classifiers~\cite{xu2021multigraph} and recurrent neural networks~\cite{xu2018sketchmate}. However, this format represents sketches as graphs with a large variance of sizes, e.g., some sketches have multiple strokes with negligible lengths. To this end, we preprocess the data by extracting strokes from the pixelated sketches using the method introduced in Sec.~\ref{sec:preprocess}, while preserving the information about the start and end points of the extracted strokes. During this process, we delete strokes with negligible lengths less than 5 pixels.

We also note that due to the abstract nature of sketches in some categories of Quickdraw, we only adopt two subsets of the dataset for our experiments. The first subset contains all shape categories including \textit{circle, hexagon, line, octagon, square, triangle, zigzag} (see 
Fig.~\ref{fig:sample_attack}(b)), and the second contains all body categories including \textit{arm, ear, elbow, face, finger, foot, hand, nose, toe, tooth} (see Fig.~\ref{fig:sample_attack}(c)). For both subsets, we select $1000$, $100$, $100$ samples per category for training, validation and testing, respectively. 
To avoid strokes being moved out of the image through transformations during robustness evaluation, we zero-pad the image with 40 pixels on each side.
Since the strokes in the sketch image are mostly short ones, we set the number of control points on each stroke to $n = 10$ for all experiments.
We also notice the existence of complicated fork points, i.e., a small cluster of connected strokes in place of a single fork point, due to the use of the maximum circle criterion. To this end, we dilate the sketch with 4 pixels which addresses this issue. 

\paragraph{Network architecture and training details}
We adopt MPNN from Sec.~\ref{sec:mpnn} following the same architecture as the gated graph neural networks (GG-NNs)~\cite{li2015gated}. 
The message passing function is 
\begin{equation}
M_t(v_i^{(t)}, v_j^{(t)}, e_{i,j}^{(t)}) = \Phi_1(e_{i,j}^{(t)}) \cdot v_j^{(t)}.
\end{equation}
The update function is a Gated recurrent unit~\cite{cho2014properties}, where $U_t = \text{GRU}(v_i^{(t)}, m_{v_i}^{(t+1)})$. The readout function is 
\begin{equation}
R = \sum_{v_i^{(T)} \in v} \sigma(\Phi_2(v_i^{(T)}, v_i^{(0)})) \odot (\Phi_3(v_i^{(T)})).
\end{equation}
For MNIST, $\Phi_1, \Phi_2, \Phi_3$ each is a linear four-layer fully-connected network, with the intermediate feature sizes as $128$, $256$, and $128$. The message passing iterations $T$ is set to 1 and the final feature vector size is set to 10. We use the batch size $128$, with an initial learning rate $= 1e^{-4}$.
To handle the complex images in Quickdraw, we increase the depth of our architecture to $8$ linear layers and the dimensions of the intermediate features are $128$, $256$, $512$, $2048$, $521$, $256$, $128$. The message passing iterations is set to $3$ and the dimension of the final feature vector is $1024$. We set the the batch size as eight with an initial learning rate $2e^{-4}$ using the SGD optimizer. The objective function follows Eq.~\ref{eq:cls}.

\paragraph{Baselines}
\textbf{Convolutional Neural Networks (CNNs)}: On MNIST, we train a CNN with two convolutional layers and two linear layers maintaining comparable learnable parameters to our model for fair comparisons. On Quickdraw, we choose the Inception network~\cite{szegedy2015going} as the baseline.

\textbf{Recurrent neural networks (RNNs)}~\cite{ha2017neural} encodes a sketch as a sequence of key points and flag bits, indicating the start or end of the strokes. SketchMate~\cite{xu2018sketchmate} fuses the CNN encoding with the RNN encoding. In our experiment, We use a bi-directional GRU as a baseline. 

\textbf{Graph based networks}~\cite{xu2021multigraph}, including graph convolution network (GCN), graph attention network (GAT) and graph transformer, encode a sketch as graph. We choose graph transformer as our baseline since it shows better performance than GCN and GAT. This method is different from ours: Each vertex is a key point of a stroke, represented by the corresponding coordinates. Therefore, this representation is not invariant to spatial transformations.
In addition, Graph Transformer based methods need information about the start or end points of the strokes as inputs. In our method, we extract this information from fork point detection.

\begin{figure*}[!ht]
    \centering
    \begin{adjustbox}{width=1.0\linewidth}
        \begin{tabular}{c|c}
            \begin{tabular}[width=0.5\linewidth]{ccccccc}
                & \huge Step 0 & \huge Step 0  & \huge Step 100  & \huge Step 1000 & \huge Last step & \huge Last step \\
                & \huge Image &\huge 1(0.9928)  & \huge 7(0.9862) & \huge 7(0.9907) & \huge 7(0.9963) & \huge Image
 
                \\ 
                \raisebox{3.5em}{\Huge(a)} &
                \raisebox{-1mm}{\includegraphics[width=0.205\linewidth]{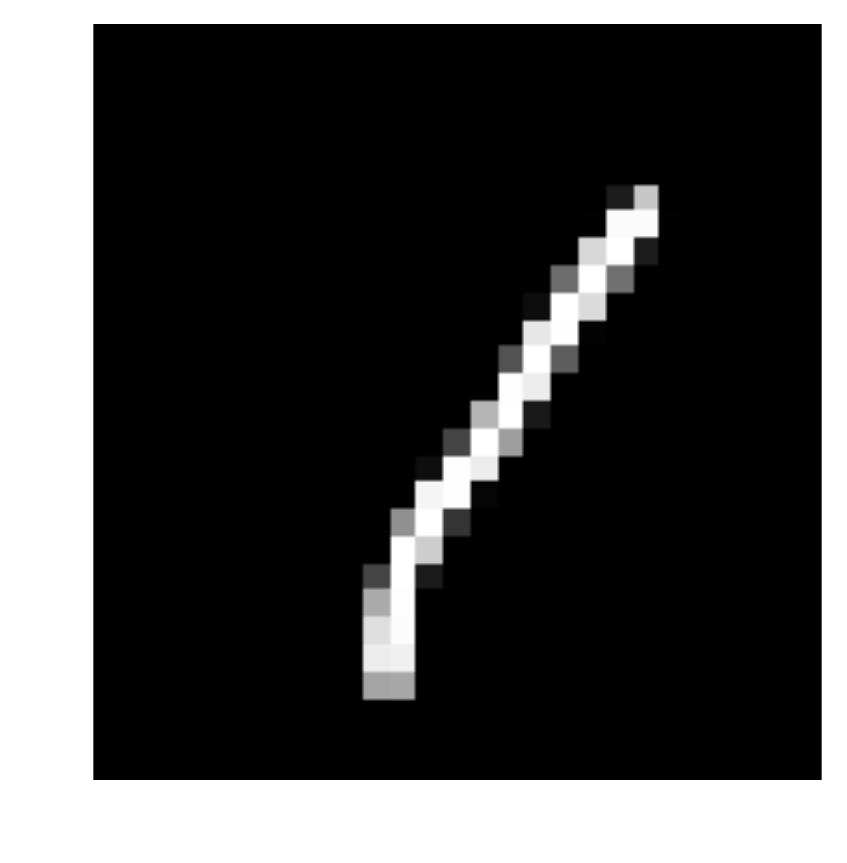}} &
                \includegraphics[width=0.2\linewidth]{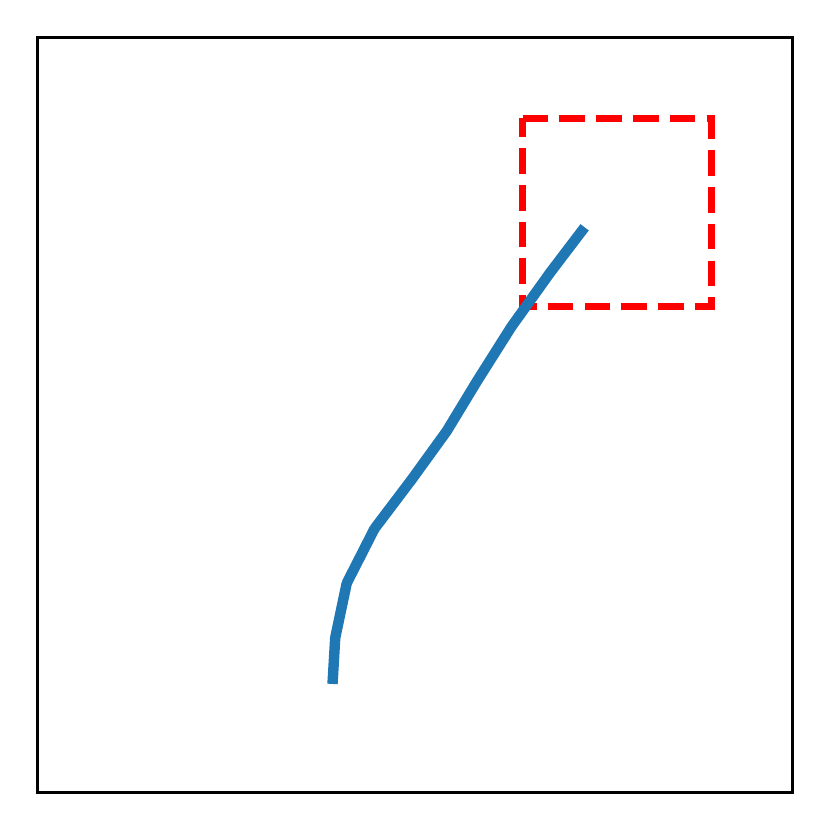} &

                \includegraphics[width=0.2\linewidth]{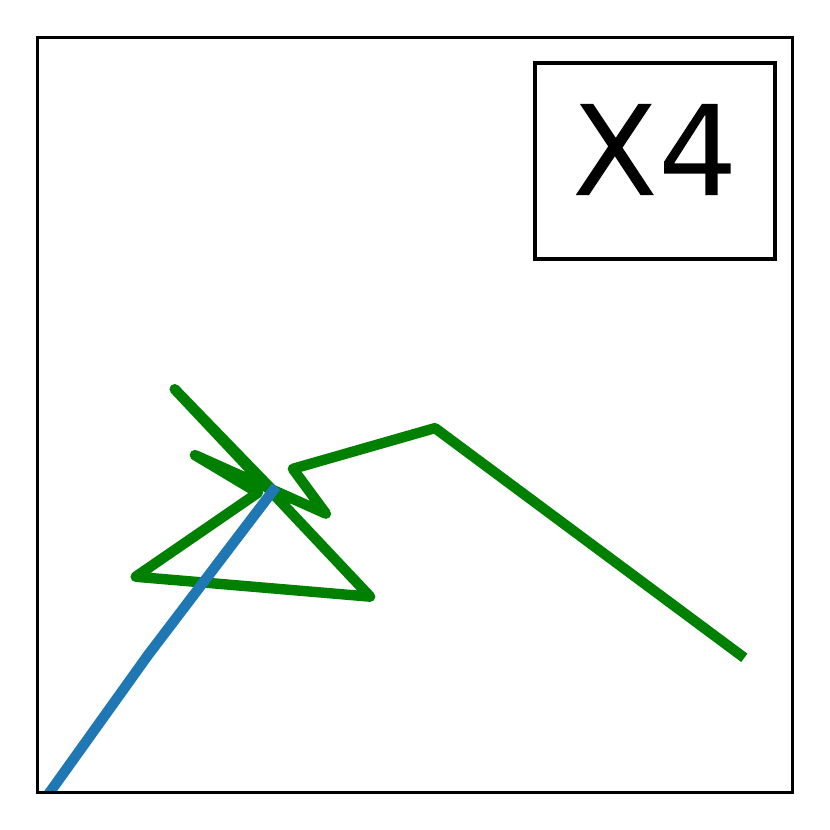} &
                \includegraphics[width=0.2\linewidth]{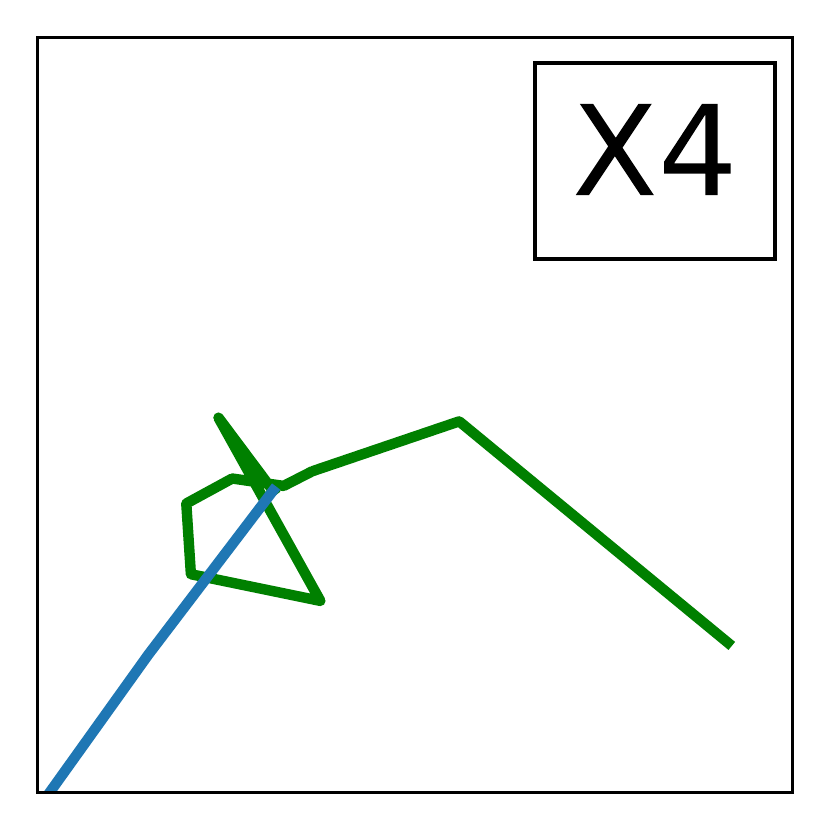} &
                \includegraphics[width=0.2\linewidth]{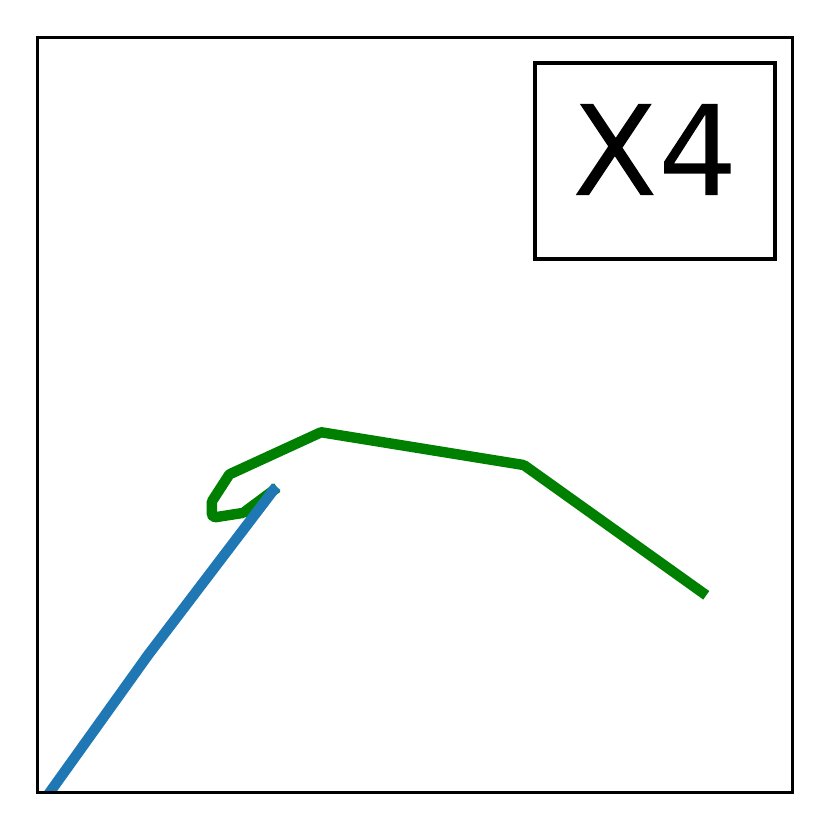} &
                \raisebox{-1mm}{\includegraphics[width=0.205\linewidth]{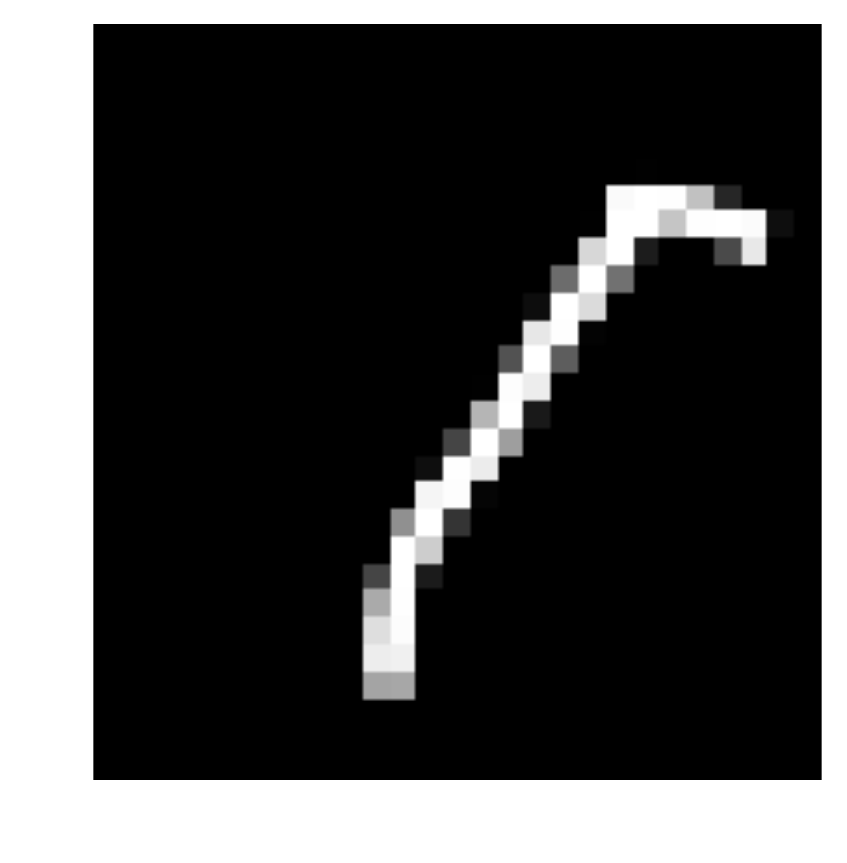}} 
            \end{tabular}
            & 
            \begin{tabular}[width=0.5\linewidth]{ccccccc}
                & \huge Step 0 & \huge Step 0  & \huge Step 100  & \huge Step 1000  & \huge Last step & \huge Last step  \\
                 & \huge Image & \huge 6(0.9789) & \huge 9(0.4573) & \huge 0(0.9640) & \huge 0(0.9952) &  \huge Image\\ 
                \raisebox{3.5em}{\Huge(d)} &
                \raisebox{-1mm}{\includegraphics[width=0.205\linewidth]{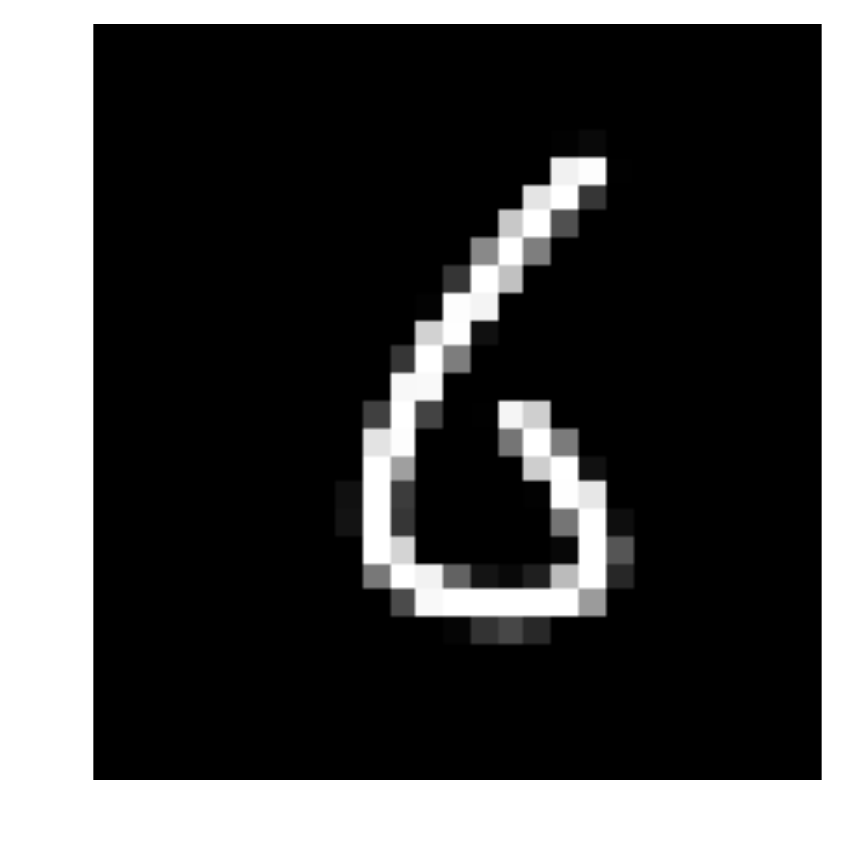}} &
                \includegraphics[width=0.2\linewidth]{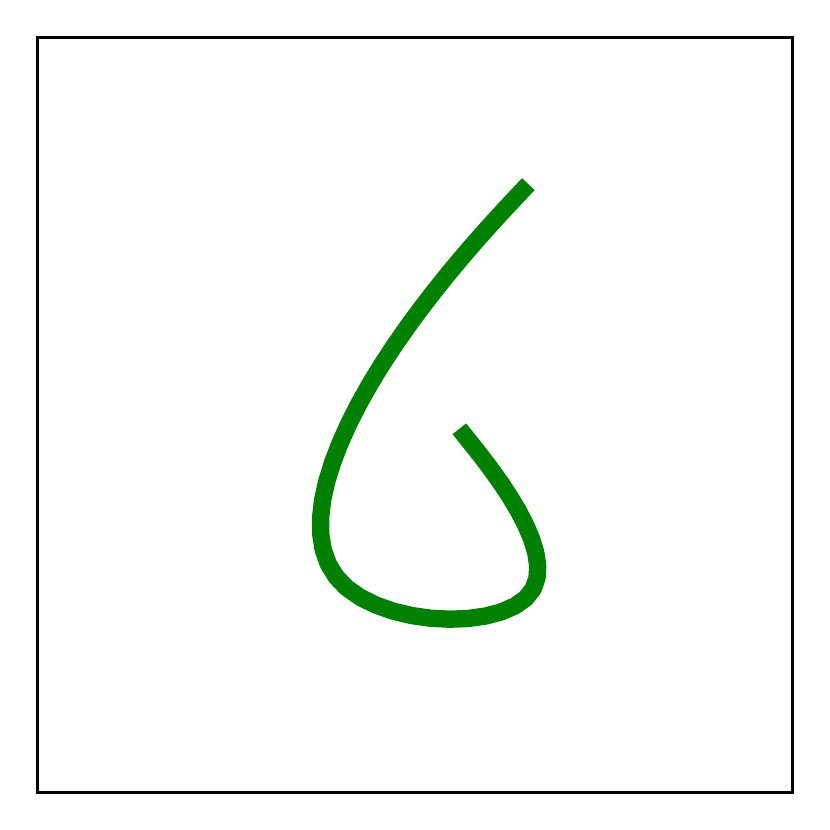} &
                \includegraphics[width=0.2\linewidth]{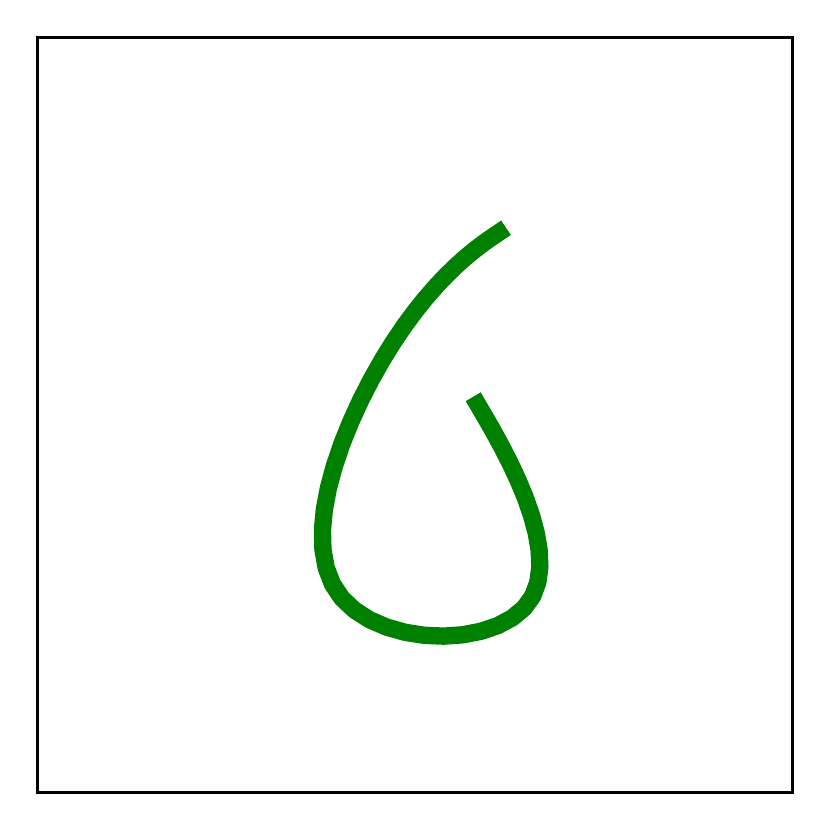} &
                \includegraphics[width=0.2\linewidth]{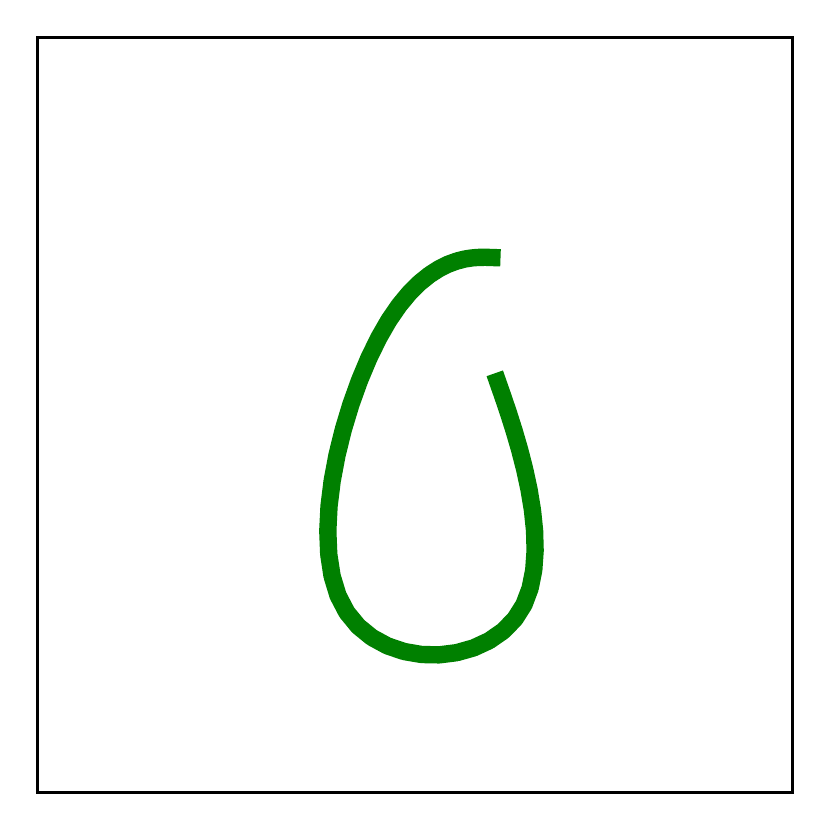} &
                \includegraphics[width=0.2\linewidth]{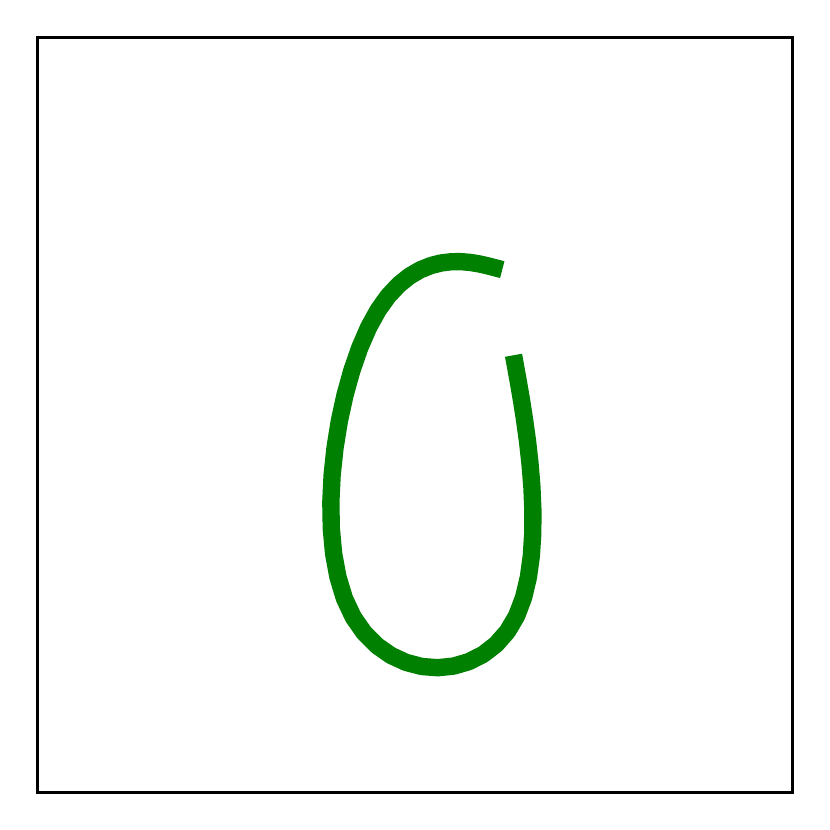} &
                \raisebox{-1mm}{\includegraphics[width=0.205\linewidth]{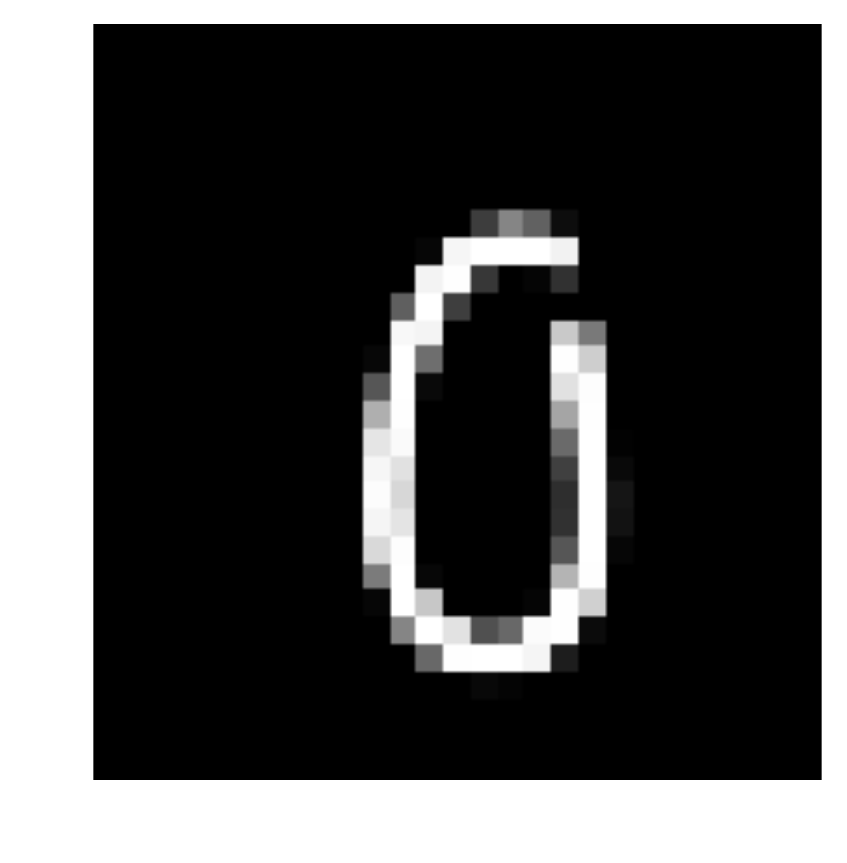}}
           
            \end{tabular}
             \\
            \newline \\ 
            \begin{tabular}[width=0.6\linewidth]{ccccccc}

                &  &\huge 1(0.9964) & \huge 1(0.7139) & \huge 7(0.6389) &  \huge 7(0.9788) & 
                \\ 
                \raisebox{3.5em}{\Huge(b)} &
                \raisebox{-1mm}{\includegraphics[width=0.205\linewidth]{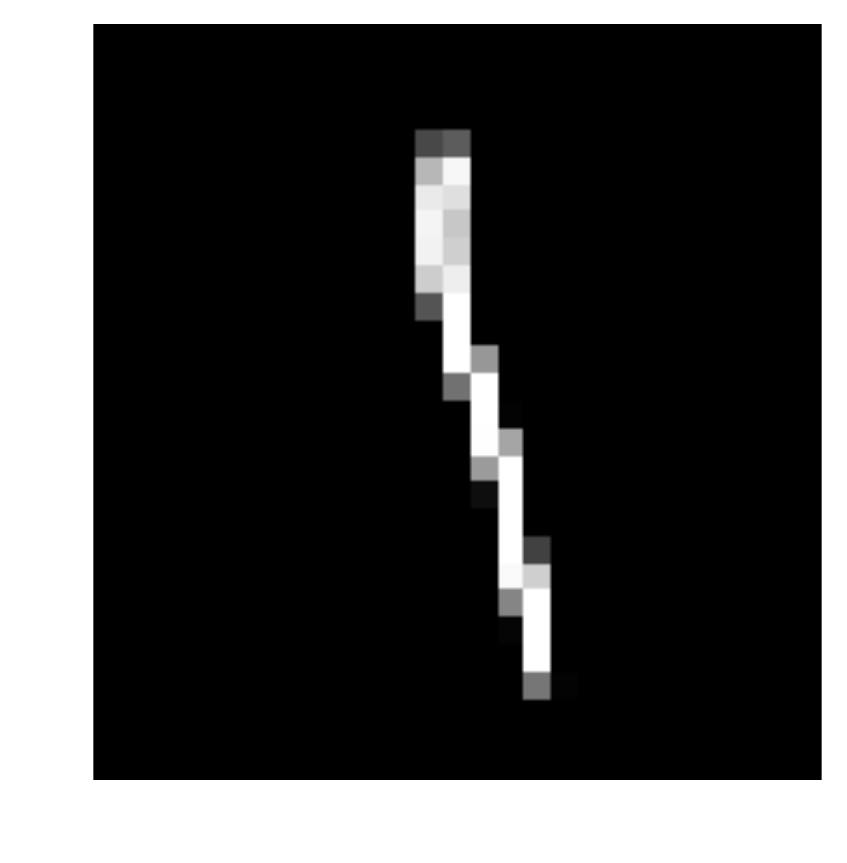}} &
                \includegraphics[width=0.2\linewidth]{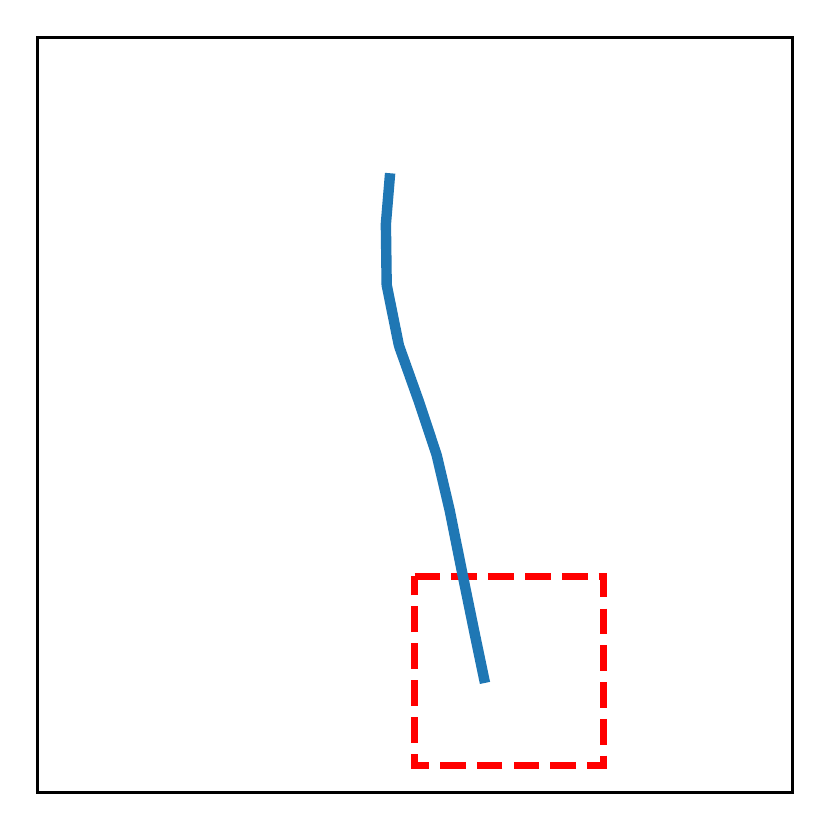} &
                \includegraphics[width=0.2\linewidth]{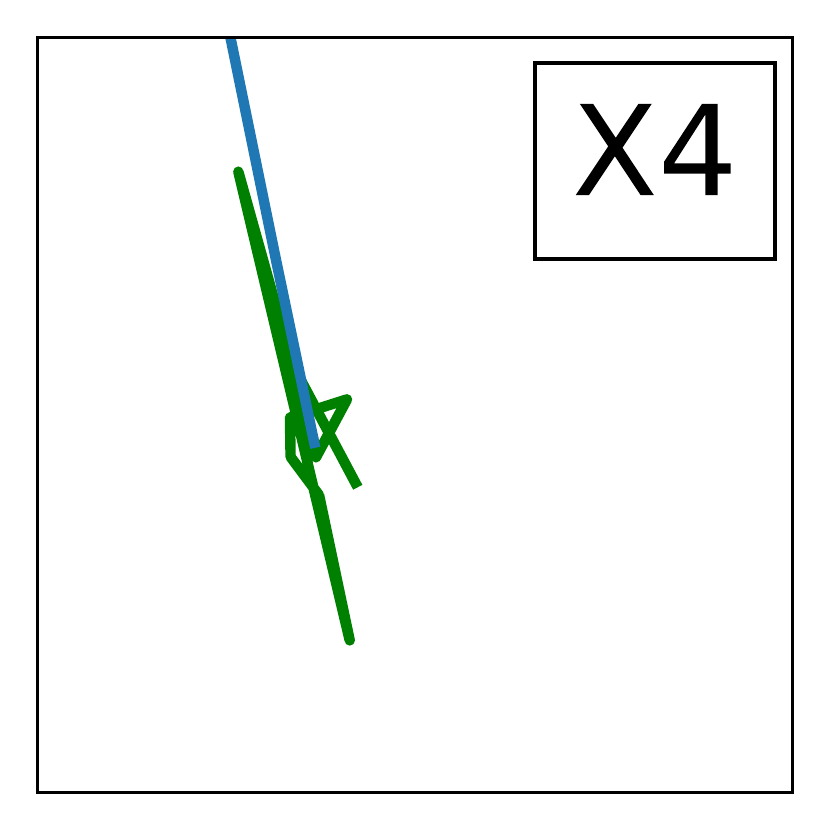} &
                \includegraphics[width=0.2\linewidth]{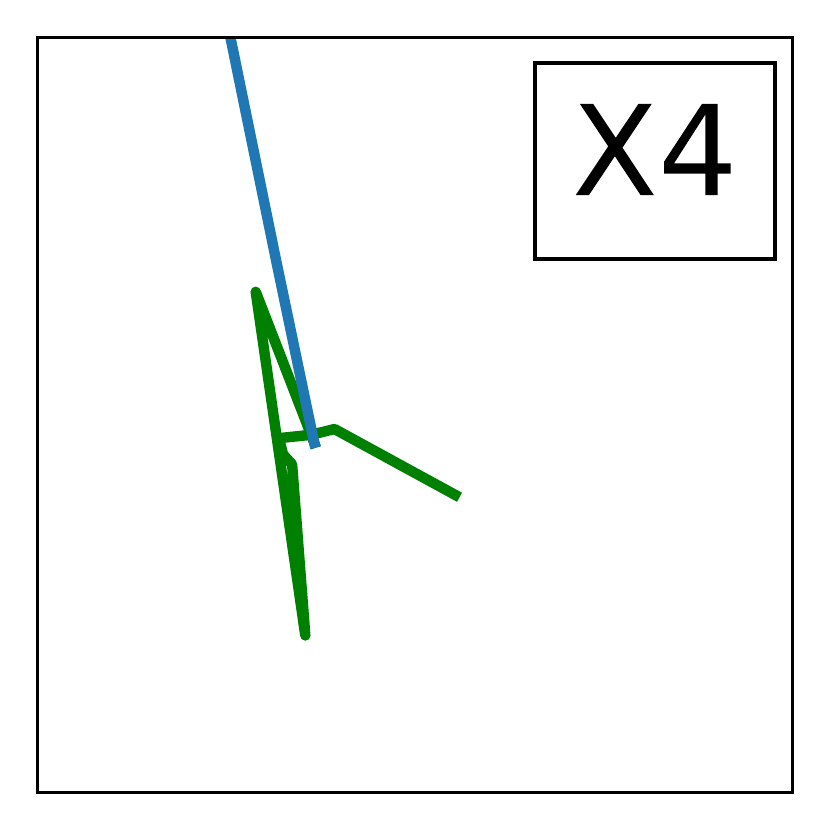} &

                \includegraphics[width=0.2\linewidth]{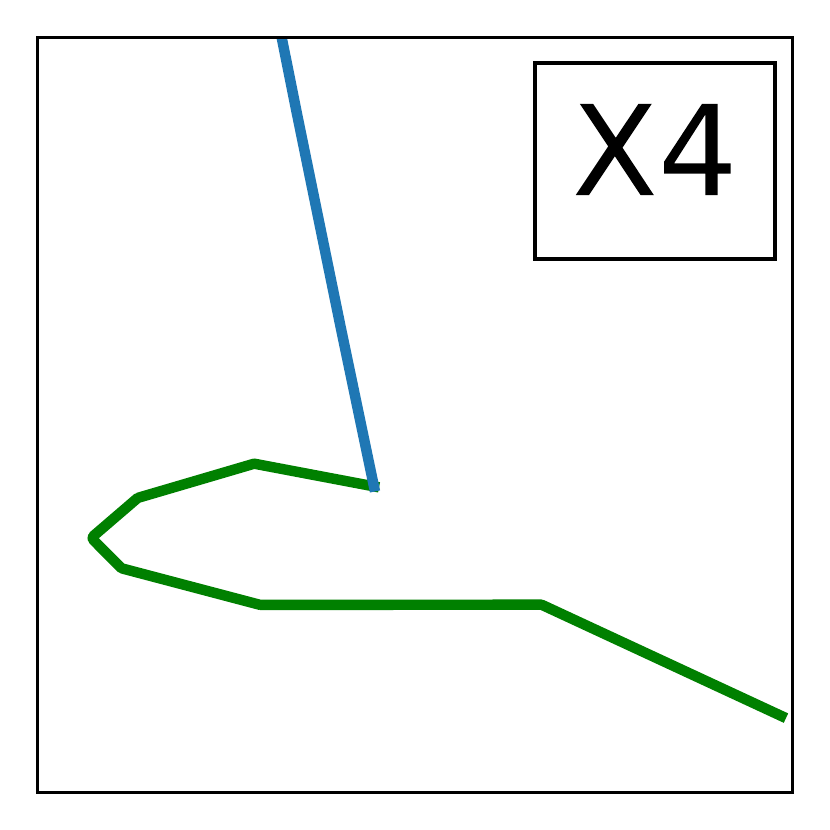} &
                \raisebox{-1mm}{\includegraphics[width=0.205\linewidth]{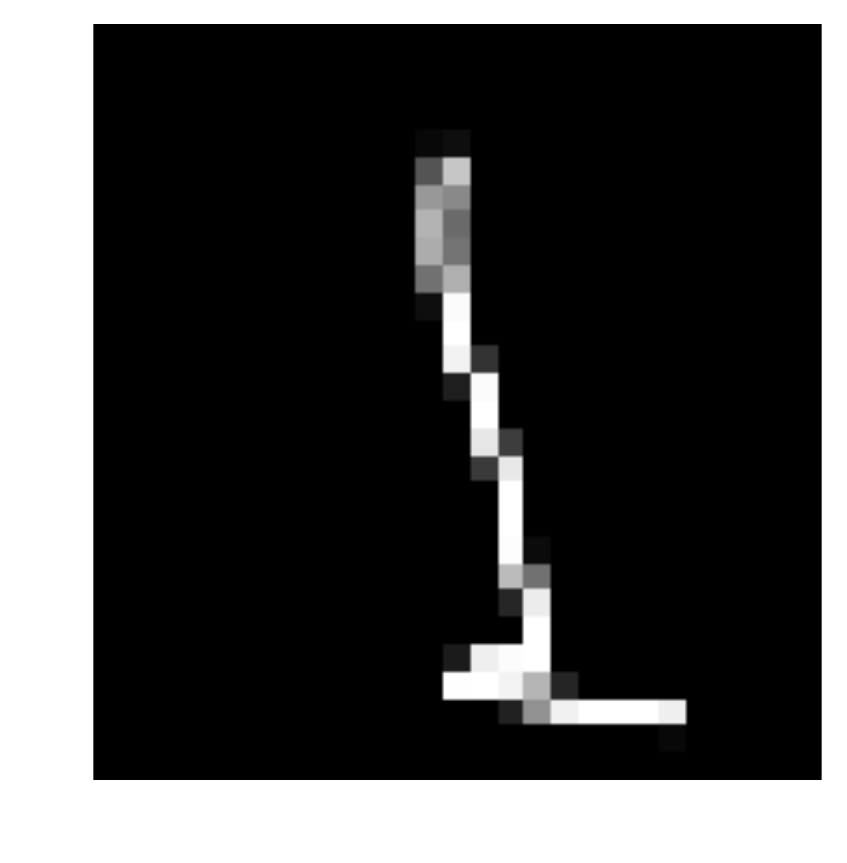}}
            \end{tabular}
            
            & 
            \begin{tabular}[width=0.6\linewidth]{ccccccc}

                &  & \huge 6(0.9773) & \huge 6(0.8451) & \huge 0(0.8163) & \huge 0(0.9946) & \\ 
                \raisebox{3.5em}{\Huge(e)} &
                \raisebox{-1mm}{\includegraphics[width=0.205\linewidth]{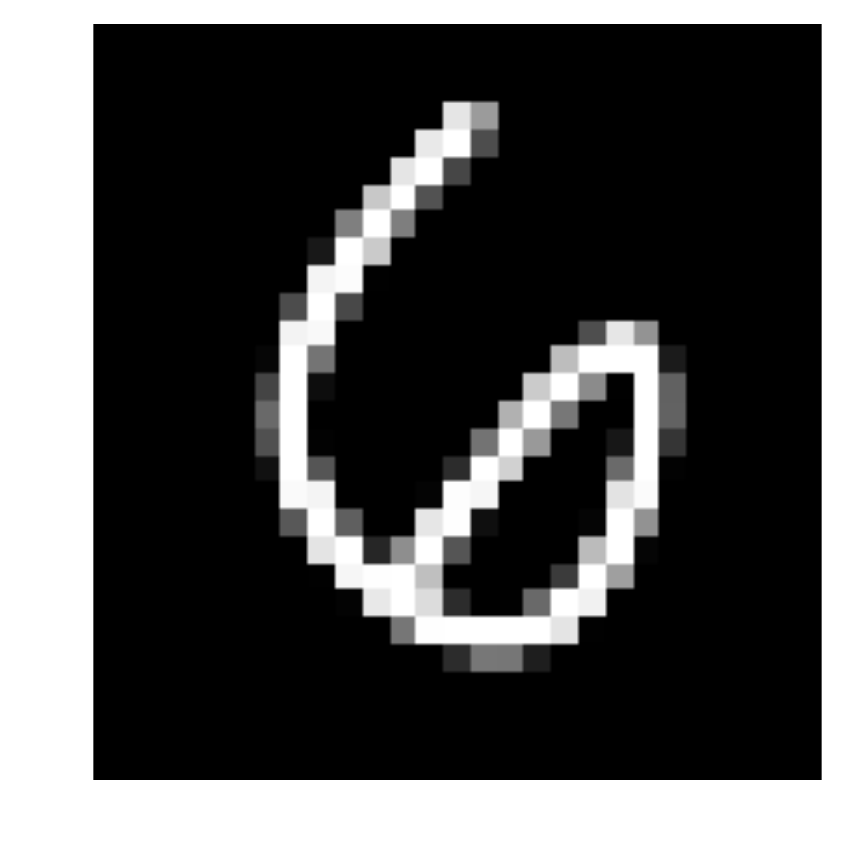}} &
                \includegraphics[width=0.2\linewidth]{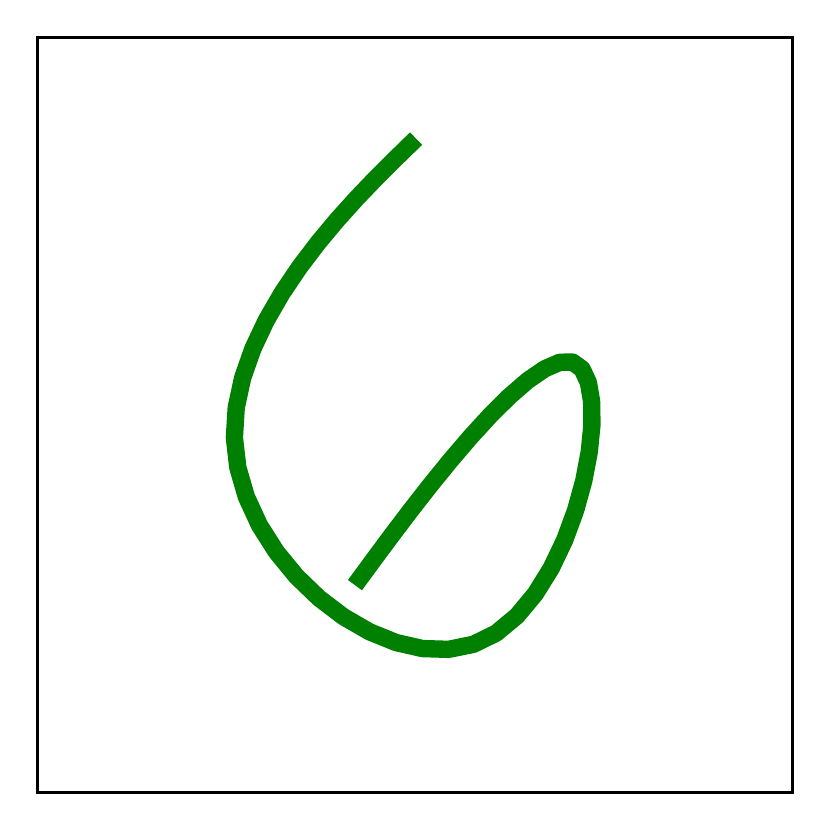} &
                \includegraphics[width=0.2\linewidth]{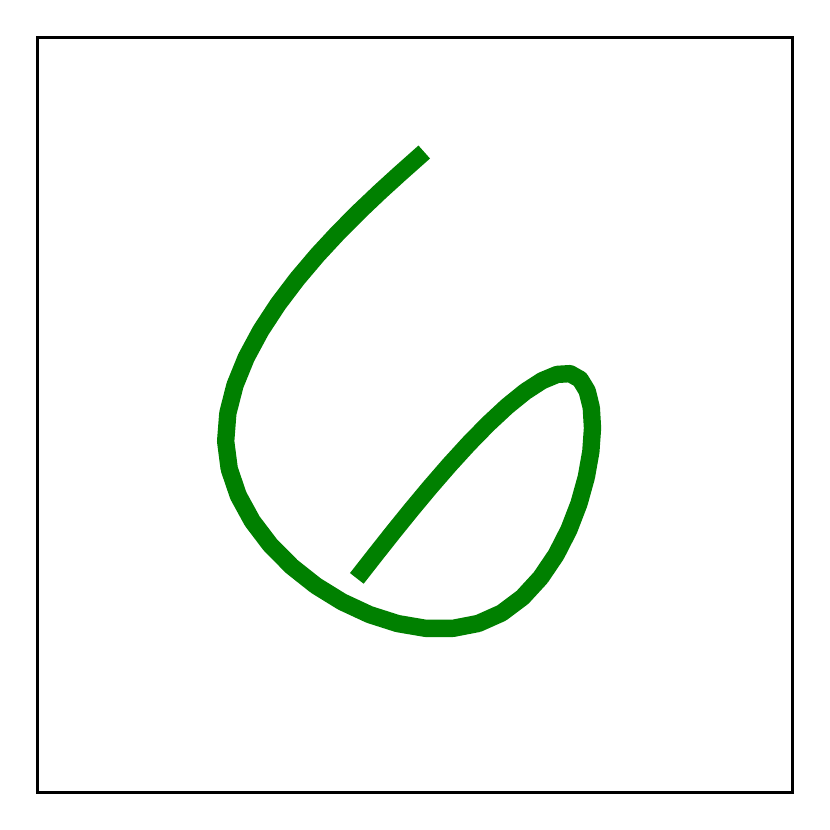} &
                \includegraphics[width=0.2\linewidth]{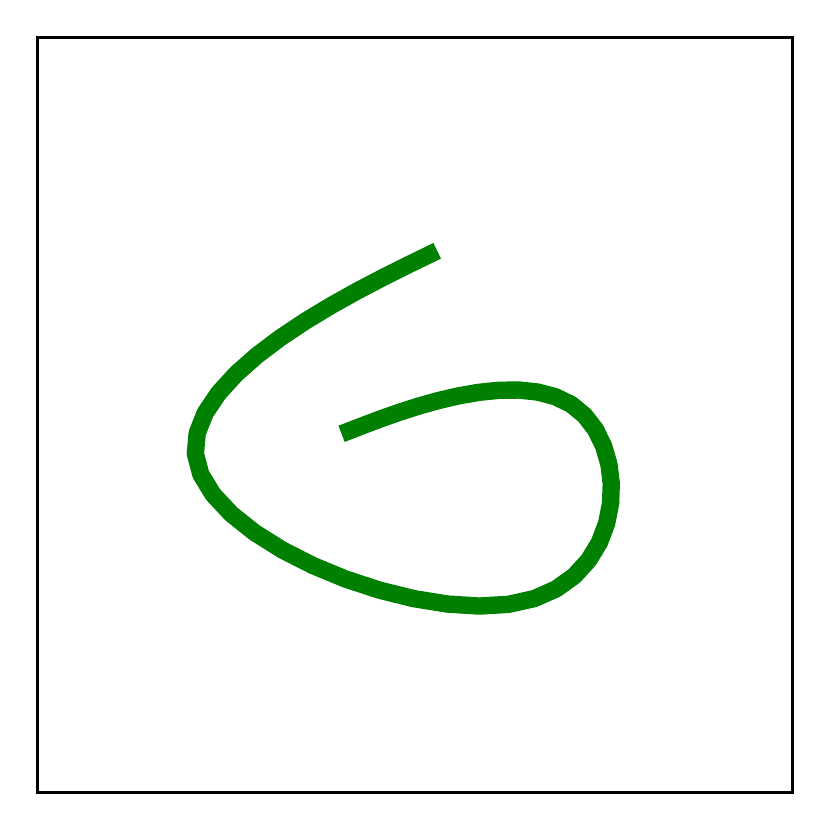} &
                \includegraphics[width=0.2\linewidth]{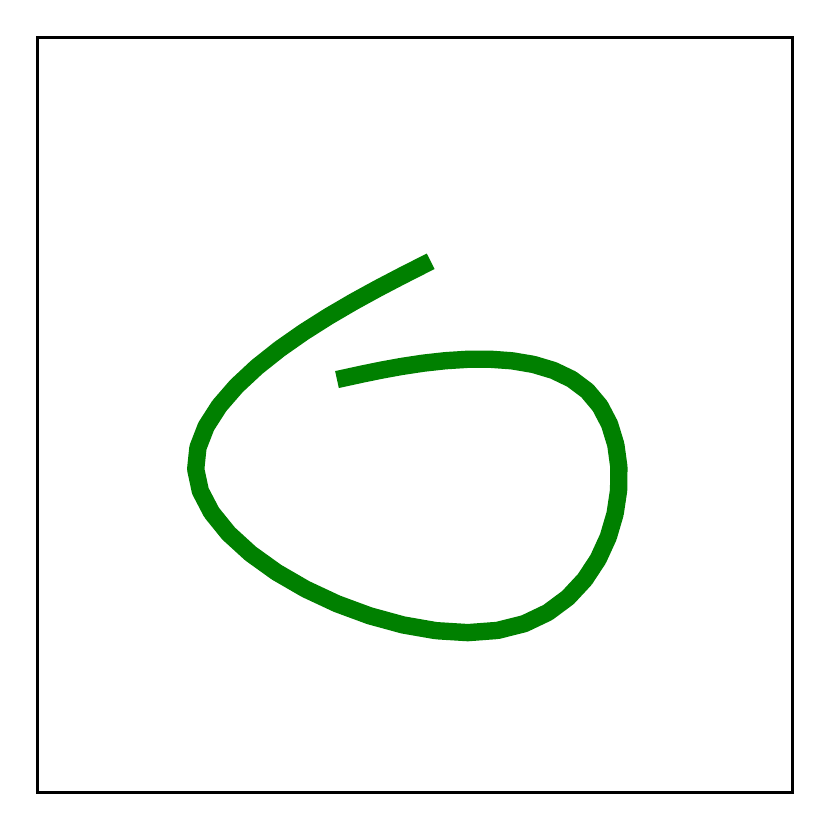} &
                \raisebox{-1mm}{\includegraphics[width=0.205\linewidth]{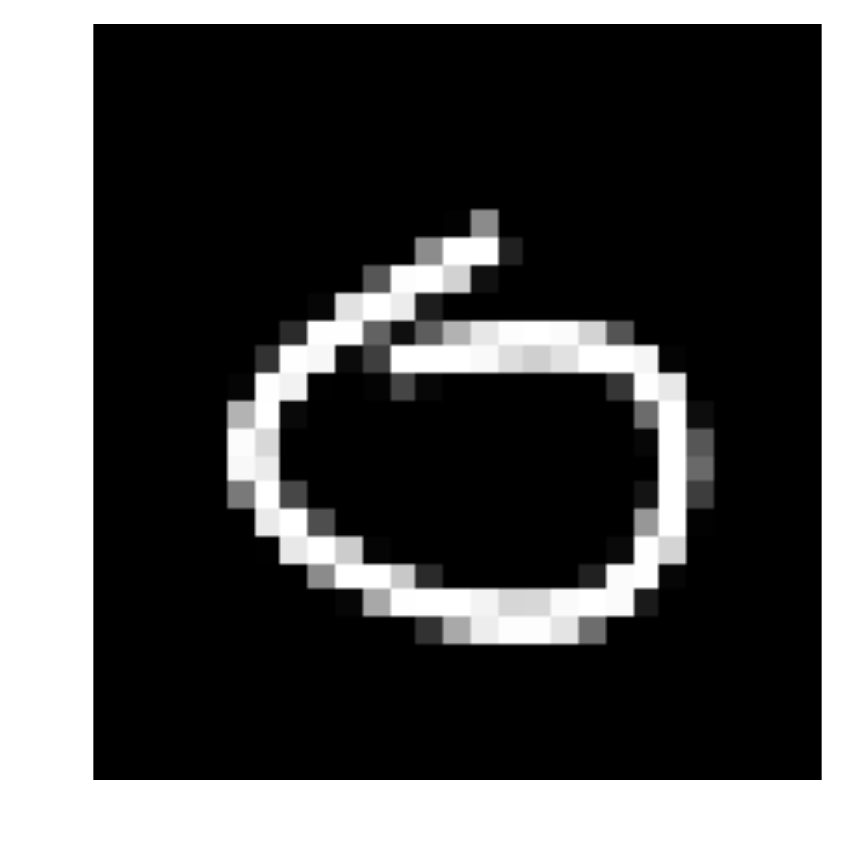}} 
           
            \end{tabular} \\ 
            \newline\\
            \begin{tabular}[width=0.6\linewidth]{ccccccc}

                &  &\huge 7(0.9999)  & \huge 7(0.5978) & \huge 2(0.8600) & \huge 2(0.9655)  \\
                \raisebox{3.5em}{\Huge(c)} &
                \raisebox{-1mm}{\includegraphics[width=0.205\linewidth]{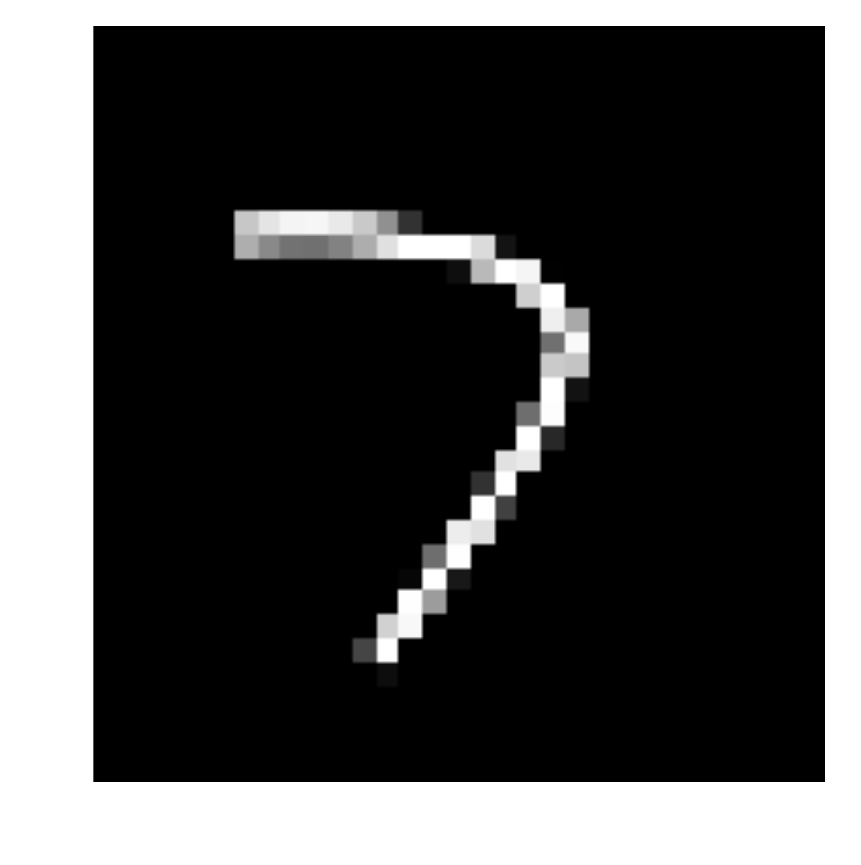}} &
                \includegraphics[width=0.2\linewidth]{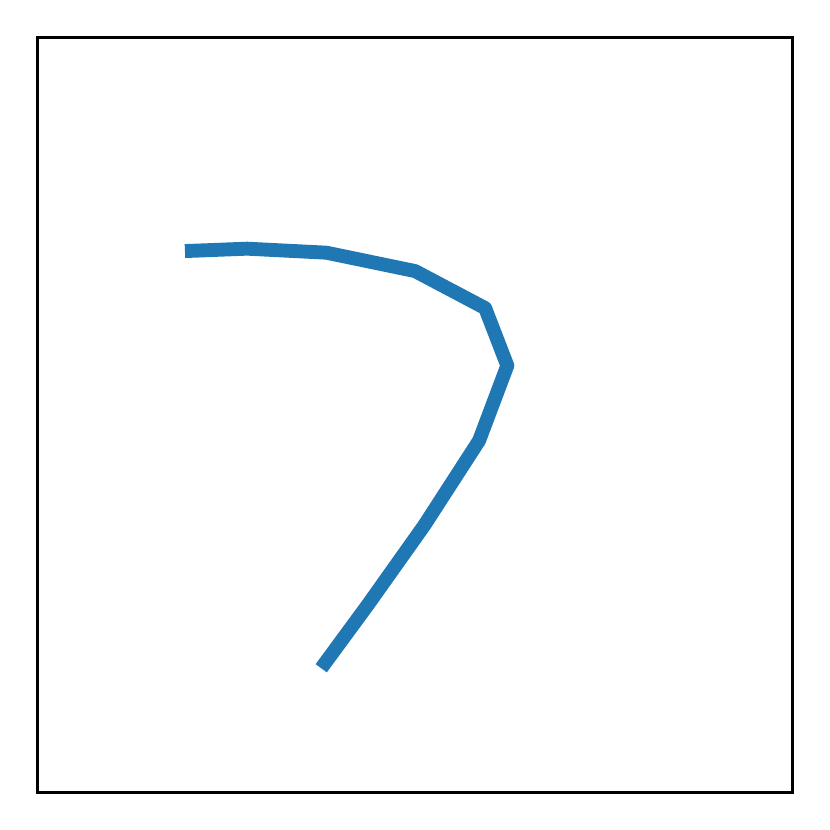} &
                \includegraphics[width=0.2\linewidth]{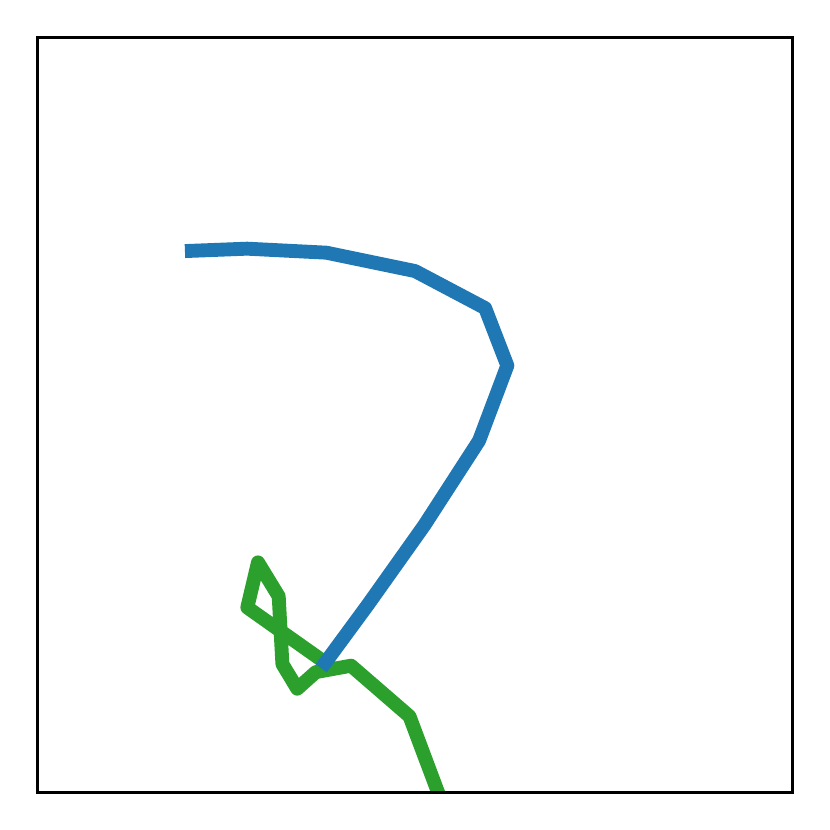} &
                \includegraphics[width=0.2\linewidth]{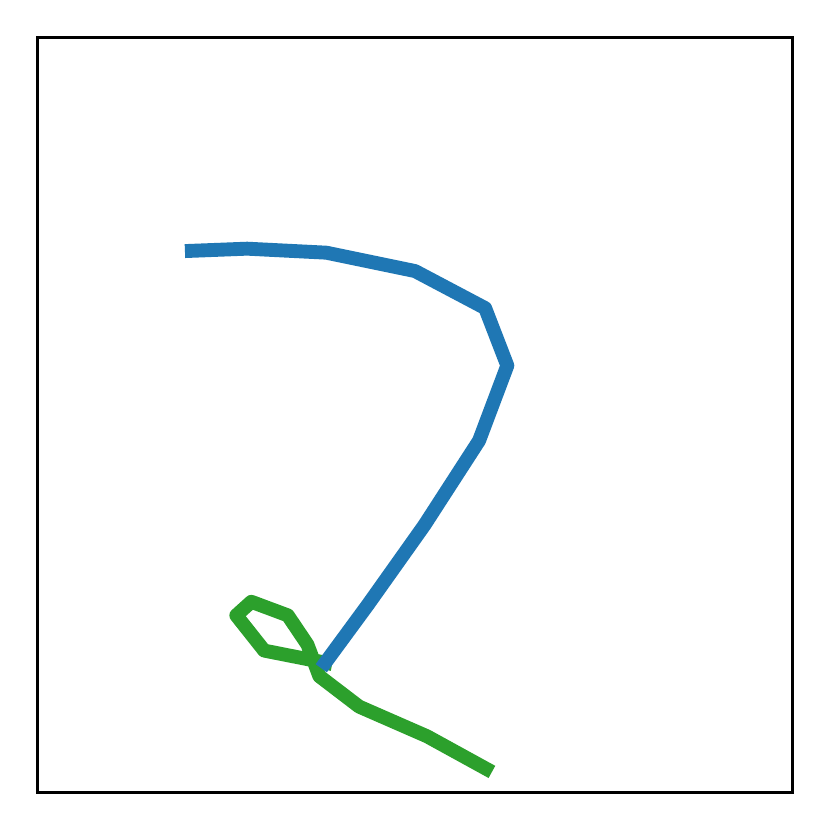} &
                \includegraphics[width=0.2\linewidth]{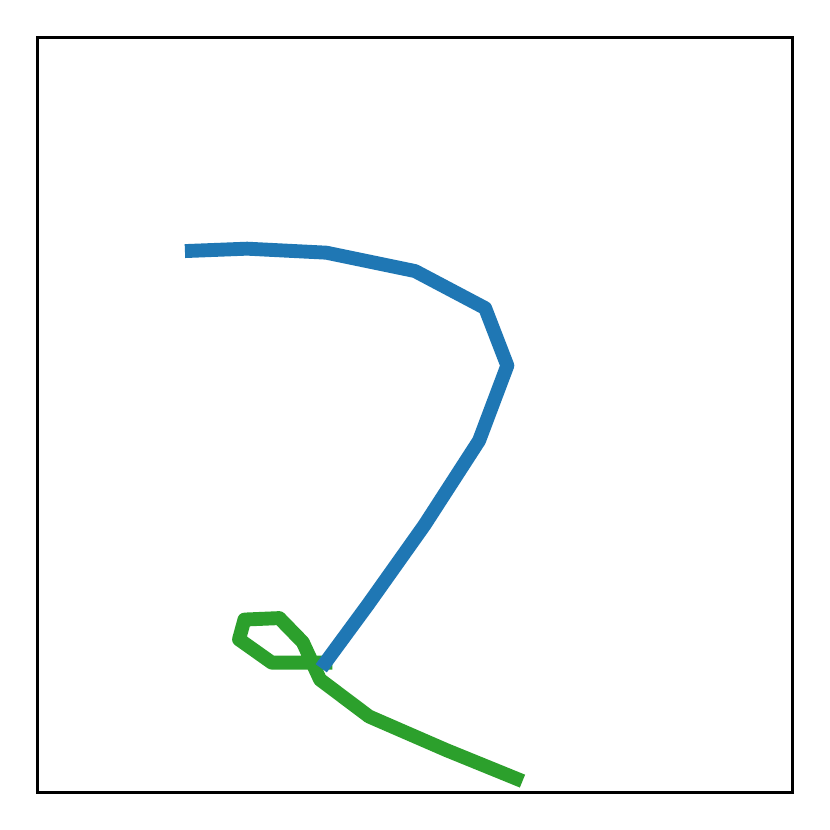} &
                \raisebox{-1mm}{\includegraphics[width=0.205\linewidth]{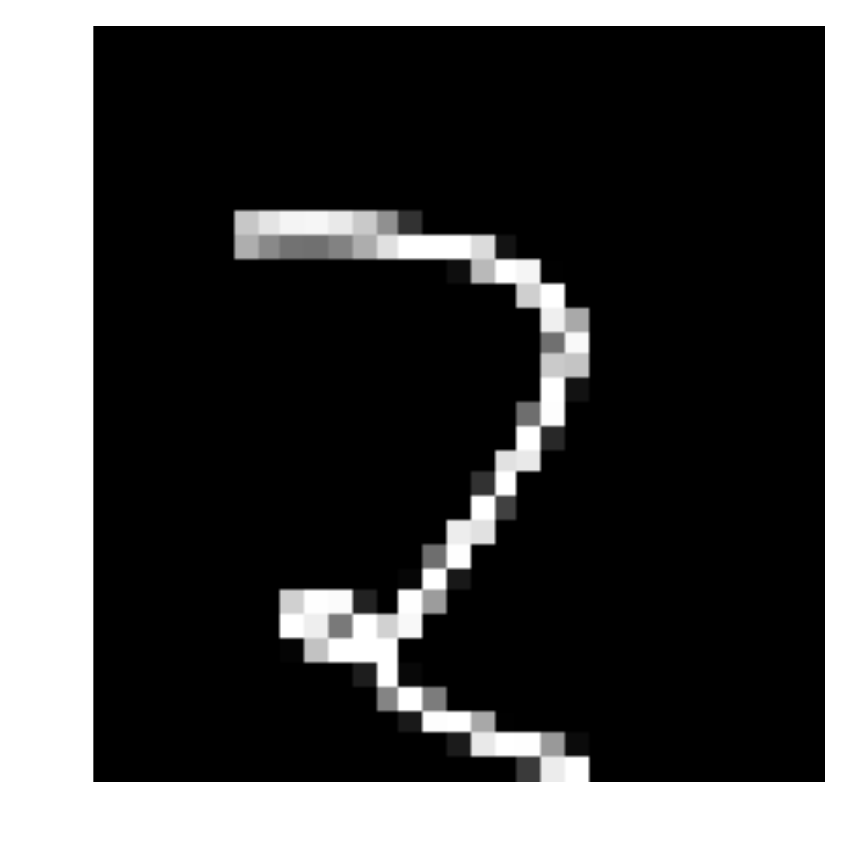}}
            \end{tabular}
            &
            \begin{tabular}[width=0.6\linewidth]{ccccccc}
                &  & \huge 7(0.9995) & \huge 7(0.6192) & \huge 1(0.9312) &  \huge 1(0.9911) &   \\ 
                \raisebox{3.5em}{\Huge(f)} &
                \raisebox{-1mm}{\includegraphics[width=0.205\linewidth]{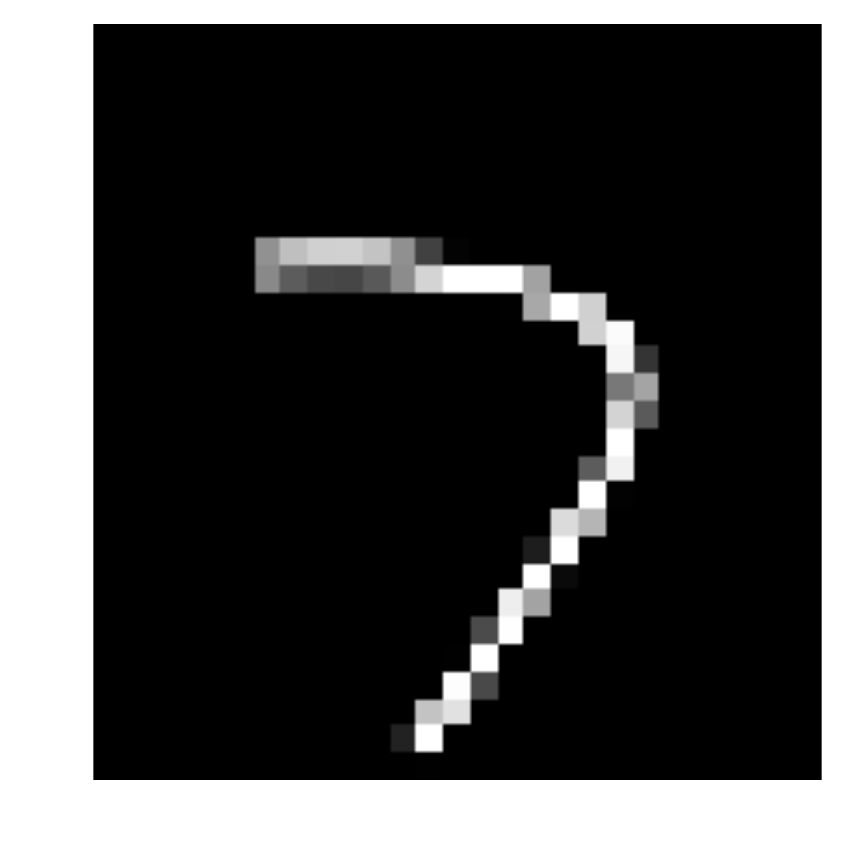}} &
                \includegraphics[width=0.2\linewidth]{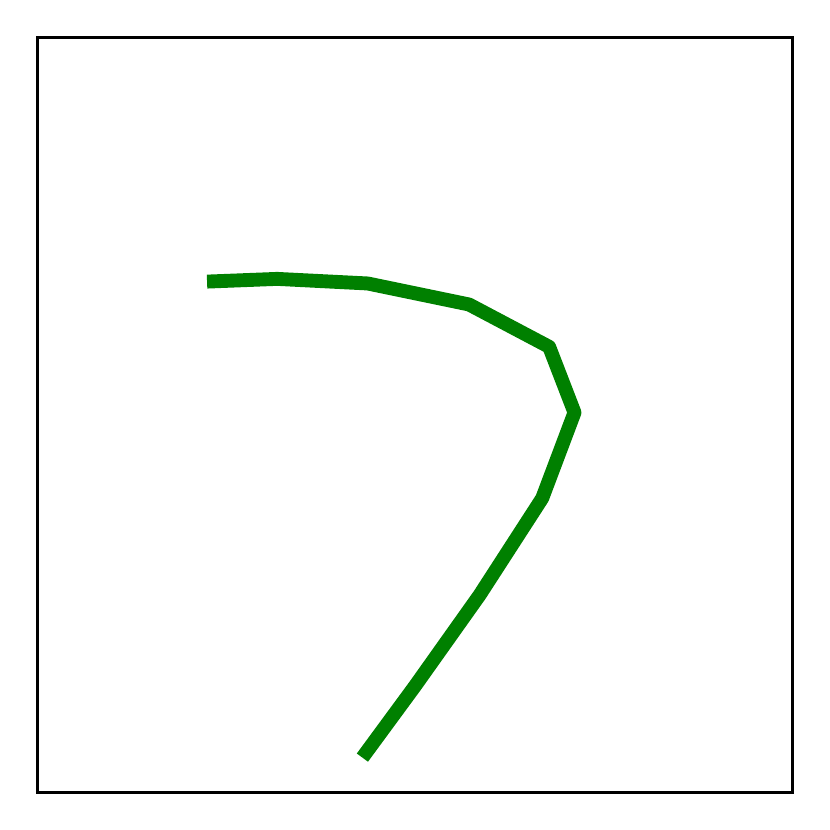} &
                \includegraphics[width=0.2\linewidth]{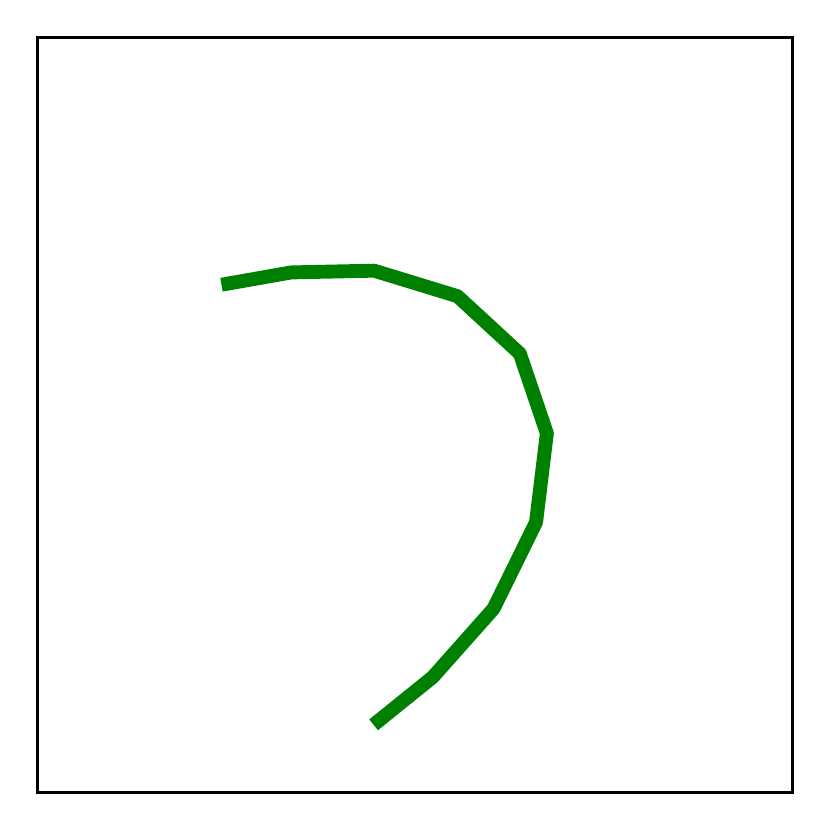} &
                \includegraphics[width=0.2\linewidth]{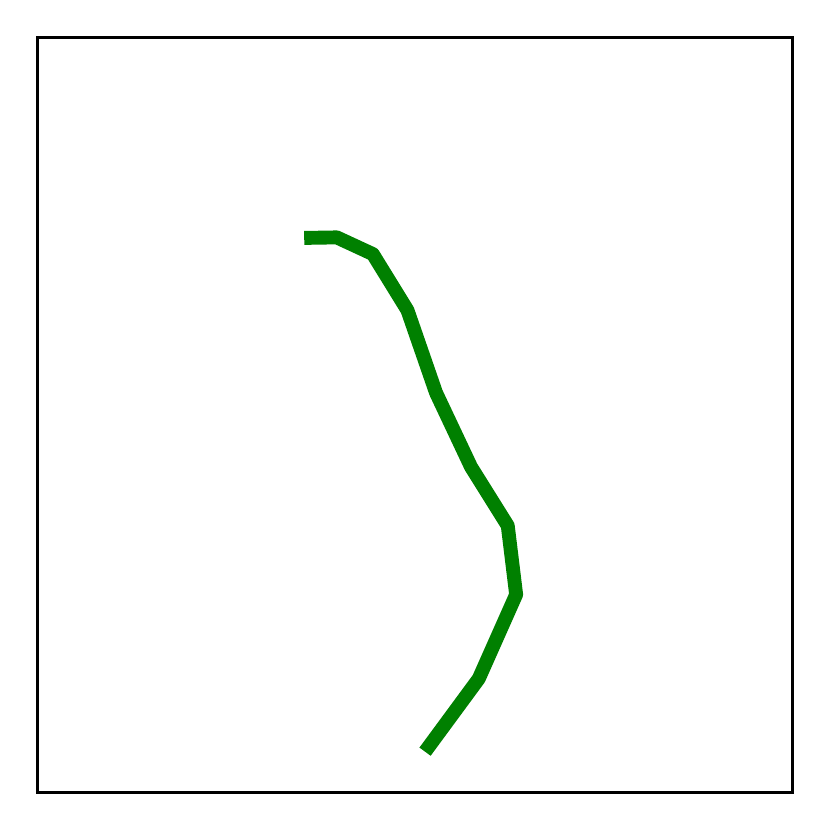} &
                \includegraphics[width=0.2\linewidth]{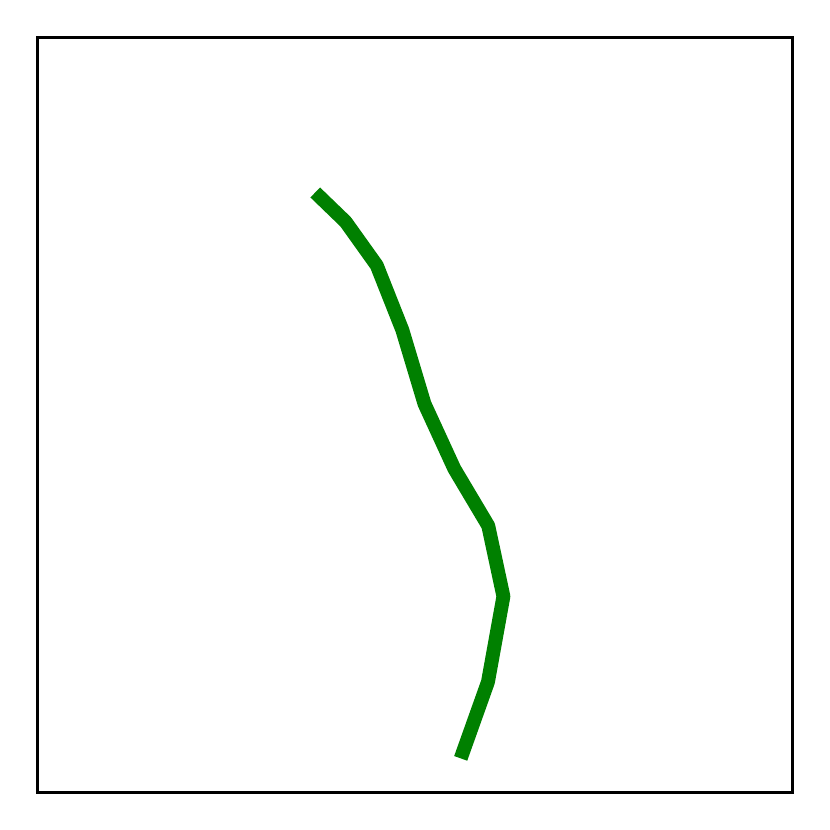} &
                \raisebox{-1mm}{\includegraphics[width=0.205\linewidth]{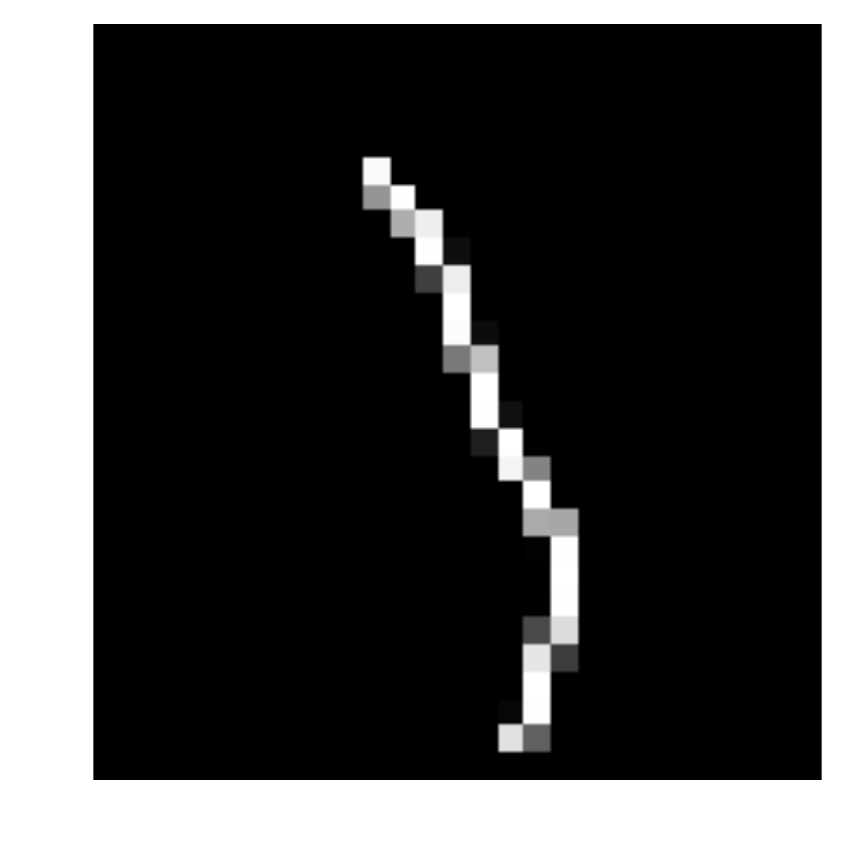}}
            \end{tabular}
 
        \end{tabular}
        
    \end{adjustbox}

    \vspace*{0.5 cm}
    
    \caption{\textbf{Exploring the robust feature by altering the graph topology and modifying the control points.} (a)(b)(c) The robust feature by altering the graph topology. The first image is the original image. The last step image is the image after adding one stroke. The middle ones are intermediate steps. (a)(b) from a single-stroke digit 1 to a two-strokes digit 7. (c) from a single-stroke 7 to a two-strokes 2. In (a)(b), the red block on the step 0 image indicates the ZOOM in windows. (d)(e)(f) show the robust feature by modifying the control points.  (c)(d) from a single-stroke 6 to a single-stroke 0. (f) from a single-stroke 7 to a single-stroke 1.}
    \label{fig:robust_feature1}
 \end{figure*}            

\paragraph{Evaluation}
To evaluate the spatial robustness of the model, we apply rotation $\theta$ and translation $(\delta_x, \delta_y)$ attacks on the input images following \cite{engstrom2019exploring}: 
\begin{equation}
    \begin{bmatrix}
      \cos \theta & -\sin \theta  \\
      \sin \theta & \cos \theta  
    \end{bmatrix}
    \cdot
    \begin{bmatrix}
      x \\
      y 
    \end{bmatrix}
    + 
    \begin{bmatrix}
      \delta_x  \\
      \delta_y  
    \end{bmatrix}
    =
    \begin{bmatrix}
      x' \\
      y'
    \end{bmatrix},
\end{equation}
for pixel coordinates $(x,y)$. For MNIST, we rotate within $\pm 30^{\circ}$ and translate within $\pm 3$ pixels. For Quickdraw, since the image size increases, we increase the maximum translation to $10$ pixels.
To generate the transformed images, we discretize the parameters to grids of rotations and translations (as shown in Fig.~\ref{fig:sample_attack}). We sample $5$ values per translation direction and $31$ values for rotations. Together, the procedure yields $775$ transformed samples per image. 
If one of the transformed images has incorrect predicted label through $f_{\theta} \circ f$, the model is not considered robust against the transformation with respect to that particular image~\cite{engstrom2019exploring}. 

\begin{figure}
    \centering
    \begin{adjustbox}{width=\linewidth}
        \begin{tabular}{cccc}
            \begin{adjustbox}{width=\linewidth}
                \begin{tabular}{cc}                 
                    \large{$-180^\circ$} & \\ \multirow{2}{*}{\includegraphics{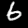} } 
                    & \hspace{-3mm} \normalsize{\color{red} 6:0.908} \\ & \hspace{-3mm} \normalsize{9:0.085}
                \end{tabular}
            \end{adjustbox}
            &
            \begin{adjustbox}{width=\linewidth}
                \begin{tabular}{cc}                 
                    \large{$-150^\circ$} & \\ \multirow{2}{*}{\includegraphics{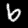} } 
                    & \hspace{-3mm} \normalsize{\color{red} 6:0.918} \\ & \hspace{-3mm} \normalsize{9:0.002}
                \end{tabular}
            \end{adjustbox}
            &
            \begin{adjustbox}{width=\linewidth}
                \begin{tabular}{cc}                 
                    \large{$-120^\circ$} & \\ \multirow{2}{*}{\includegraphics{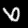} } 
                    & \hspace{-3mm} \normalsize{\color{red} 6:0.988} \\ & \hspace{-3mm} \normalsize{9:0.000}
                \end{tabular}
            \end{adjustbox}
            &
            \begin{adjustbox}{width=\linewidth}
                \begin{tabular}{cc}                 
                    \large{$-90^\circ$} & \\ \multirow{2}{*}{\includegraphics{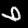} } 
                    & \hspace{-3mm} \normalsize{6:0.000} \\ & \hspace{-3mm} \normalsize{\color{red} 9:1.000}
                \end{tabular}
            \end{adjustbox}
            \\
            \\
            \\
            \\
            \begin{adjustbox}{width=\linewidth}
                \begin{tabular}{cc}                 
                    \large{$-60^\circ$} & \\ \multirow{2}{*}{\includegraphics{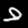} } 
                    & \hspace{-3mm} \normalsize{6:0.000} \\ & \hspace{-3mm} \normalsize{\color{red}9:0.997}
                \end{tabular}
            \end{adjustbox}
            &
            \begin{adjustbox}{width=\linewidth}
                \begin{tabular}{cc}                 
                    \large{$-30^\circ$} & \\ \multirow{2}{*}{\includegraphics{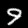} } 
                    & \hspace{-3mm} \normalsize{6:0.001} \\ & \hspace{-3mm} \normalsize{\color{red}9:0.998}
                \end{tabular}
            \end{adjustbox}
            &
            \begin{adjustbox}{width=\linewidth}
                \begin{tabular}{cc}                 
                    \large{$0^\circ$} & \\ \multirow{2}{*}{\includegraphics{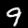} } 
                    & \hspace{-3mm} \normalsize{6:0.000} \\ & \hspace{-3mm} \normalsize{\color{red}9:1.000}
                \end{tabular}
            \end{adjustbox}
            &
            \begin{adjustbox}{width=\linewidth}
                \begin{tabular}{cc}                 
                    \large{$30^\circ$} & \\ \multirow{2}{*}{\includegraphics{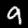} } 
                    & \hspace{-3mm} \normalsize{6:0.000} \\ & \hspace{-3mm} \normalsize{\color{red}9:0.999}
                \end{tabular}
            \end{adjustbox}
            \\
            \\
            \\
            \\
            \begin{adjustbox}{width=\linewidth}
                \begin{tabular}{cc}                 
                    \large{$60^\circ$} & \\ \multirow{2}{*}{\includegraphics{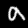} } 
                    & \hspace{-3mm} \normalsize{\color{red}6:0.997} \\ & \hspace{-3mm} \normalsize{9:0.000}
                \end{tabular}
            \end{adjustbox}
            &
            \begin{adjustbox}{width=\linewidth}
                \begin{tabular}{cc}                 
                    \large{$90^\circ$} & \\ \multirow{2}{*}{\includegraphics{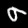} } 
                    & \hspace{-3mm} \normalsize{\color{red}6:0.970} \\ & \hspace{-3mm} \normalsize{9:0.024}
                \end{tabular}
            \end{adjustbox}
            &
            \begin{adjustbox}{width=\linewidth}
                \begin{tabular}{cc}                 
                    \large{$120^\circ$} & \\ \multirow{2}{*}{\includegraphics{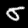} } 
                    & \hspace{-3mm} \normalsize{\color{red}6:0.973} \\ & \hspace{-3mm} \normalsize{9:0.019}
                \end{tabular}
            \end{adjustbox}
            &
            \begin{adjustbox}{width=\linewidth}
                \begin{tabular}{cc}                 
                    \large{$150^\circ$} & \\ \multirow{2}{*}{\includegraphics{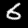} } 
                    & \hspace{-3mm} \normalsize{\color{red}6:1.000} \\ & \hspace{-3mm} \normalsize{9:0.000}
                \end{tabular}
            \end{adjustbox}

        \end{tabular}
    \end{adjustbox}
    \caption{\textbf{The samples of rotation of digit ``9".} On the right of image, we show the confidence score for rotated image predicted as 6 or 9. The prediction with higher confidence is marked in red.}
    \label{fig:69case}
\end{figure}

\paragraph{Classification and spatial robustness results}
Table~\ref{table:robustness} summarizes the experimental results. (1) For accuracy, CNNs achieve the best results. Our method achieves comparable accuracy on MNIST and slightly inferior ones on Quickdraw subsets. 
The accuracy gaps between ours and other baselines, i.e., RNNs and graph transformers, are smaller on Quickdraw subsets. Note that these models do not work directly on MNIST as they require temporal information of the strokes.
(2) Our stroke representation is invariant to rotations and translations, and therefore its robustness stays the same as its accuracy. 
We highlight that our model outperforms baselines on spatial robustness by a large margin. The robustness of CNNs drops to $26.02\%$, $21.90\%$, $31.10\%$ on the three experiments. And the robustness of RNNs and graph transformers are less than $10\%$ on Quickdraw subsets. 
(3) Our models are also efficient parameter-wise, using
only $34$\% and $21$\% of parameters of Inception-V3 and Graph Transformers, respectively. 


\paragraph{Robustness against graph attacks}\label{sec:robust-feat}
Previous studies have explored the connection between model robustness and the learning of robust features, i.e., features that are invariant to attacks~\cite{NEURIPS2019_e2c420d9, tsipras2018robustness}. The above experiments show that our model is robust to attacks in the form of global transformations. In addition, the procedure of stroke extraction is robust to conventional pixel-wise attacks due to its thinning and merging steps. Here, we further investigate the robustness of our model under graph-specific attacks. If successful, our study provides evidence that the proposed graph representation contains robust features that enable model robustness without adversarial training.  



\begin{algorithm*}
    \caption{The generation of new digits}
    \label{generation}
    \begin{algorithmic}
        \State \textbf{Initial:} Function: $f, f_{\theta}$, Existing sets: $\mathcal{D}$, mean (parameterized with $10$ control points): $t$ and variance: $\sigma$, Hyper-parameters: $\alpha, \gamma,  \lambda_1, \lambda_2, \lambda_3$;
        \State \textbf{Output:} $t, \sigma$;
        \While{not converged}
            \State $\mathcal{T} \sim \mathcal{N}(t, \sigma^2) $
            \While{not converged}
                \State $ \mathcal{L} = - \mathbb{E}_{x \sim D}\left[\log(f_\theta \circ f(g(x)))\right]  
         - \mathbb{E}_{x \sim T}\left[\log(1-f_\theta \circ f(g(x)))\right] $ ;
         \State $\theta \gets \theta - \gamma \frac{\partial \mathcal{L}}{\partial \theta}$;
            \EndWhile
            \While{not converged}
                \State $\mathcal{T} \sim \mathcal{N}(t, \sigma^2) $
                \State $\mathcal{L} = -\mathbb{E}_{x \sim T}\left[\log(1-f_\theta \circ f(g(x)))\right] + \mathbb{E}_{(v,e,r) \in (\mathcal{V},\mathcal{E},\mathcal{R})} 
                \left[-\lambda_1\log p(v;\mathcal{D})-\lambda_2\log p(e;\mathcal{D})-\lambda_3\log p(r;\mathcal{D})\right]$;
                \State $t \gets t - \alpha \frac{\partial \mathcal{L}}{\partial t}$; $\sigma \gets \sigma - \alpha \frac{\partial \mathcal{L}}{\partial \sigma}$;

            \EndWhile
            
        \EndWhile
    \end{algorithmic}
\end{algorithm*}

\textbf{Altering the graph topology}: \label{sec:add}
In this experiment, we consider adding/deleting vertices to alter the graph topology.
We conduct the experiment by adding one stroke, denoted by $s_2$, on digit ``1''s with a single stroke $s_1$. We optimize $s_2$ using Eq.~\eqref{eq:attack} by targeting the resultant graph to be classified as digit ``7''. Since our representation is spatially invariant, the start point of $s_2$ is fixed and connected to either side of $s_1$ and all other trainable $n-1$ control points of $s_2$ are initialized with the same values as the start point. 
The procedure to get $s_2$ follows the setting described in Sec.~\ref{sec:obj2}: We apply penalties on angles and pairwise distances, where $\mathcal{D}$ in Eq.~\ref{eq:discon} denotes the set of two-stroke ``7''s.

We visualize in Fig.~\ref{fig:robust_feature1}(a) the evolution of the added stroke during the optimization of Eq.~\eqref{eq:attack}. 
Part of the new stroke evolves towards being flat at step 100, while the converged stroke becomes flat. 
Similar experiments are shown in Fig.~\ref{fig:robust_feature1}(b,c) on modifying ``1'' to ``7'', and ``7'' to ``2''. Considering that our model is rotation (and mirror) invariant, the results suggest that the graph representation is robust in that when the attacks are successful, the contents of the images have to be changed semantically towards the target labels.

It should also be noted that while a rotated ``7'', as in Fig.~\ref{fig:robust_feature1}(b), should not be considered as a ``7'' from human perspective, this only happens because the added stroke is considered to move from right to left (as the first stroke of the connected two). The stroke extraction procedure described in Sec.~\ref{sec:preprocess} considers strokes to move from left to right and top to down, and therefore will avoid classifying the outcome of Fig.~\ref{fig:robust_feature1}(b) as ``7'' (since the added stroke will be considered as the second stroke of the sequence and moves from left to right). This also explains why our model can correctly classify ``6'' and ``9'' even with its invariance property in Fig.~\ref{fig:69case}: We set the strokes of the two to have different start and end control points. 

Lastly, the results in Fig.~\ref{fig:robust_feature1}(a) reveals that our model is in fact invariant to mirroring (here the added stroke will be considered as moving from right to left through the stroke extraction procedure, and thus its graph representation is equivalent to its mirrored version). While this invariance is undesirable, it is in fact commonly observed as a property of human vision system during its early phase (e.g., among children). For example, when start learning to write, children may consider ``b'' and ``d'', ``p'' and ``q'', ``J'' and ``L'' the same, as well as writing mirrored digits~\footnote{This is observed from one of the authors' 5yr old, and confirmed by her teacher.}. Removing the mirror invariance property through representation design or learning, will be deferred to a future study.


\begin{figure*}[ht]
    \centering
    \begin{adjustbox}{width=0.6\linewidth}
        \begin{tabular}{c|c}
        \hspace{-10mm}
            \begin{adjustbox}{height=0.13\linewidth}
                
                \begin{tabular}{ccc}
                \includegraphics{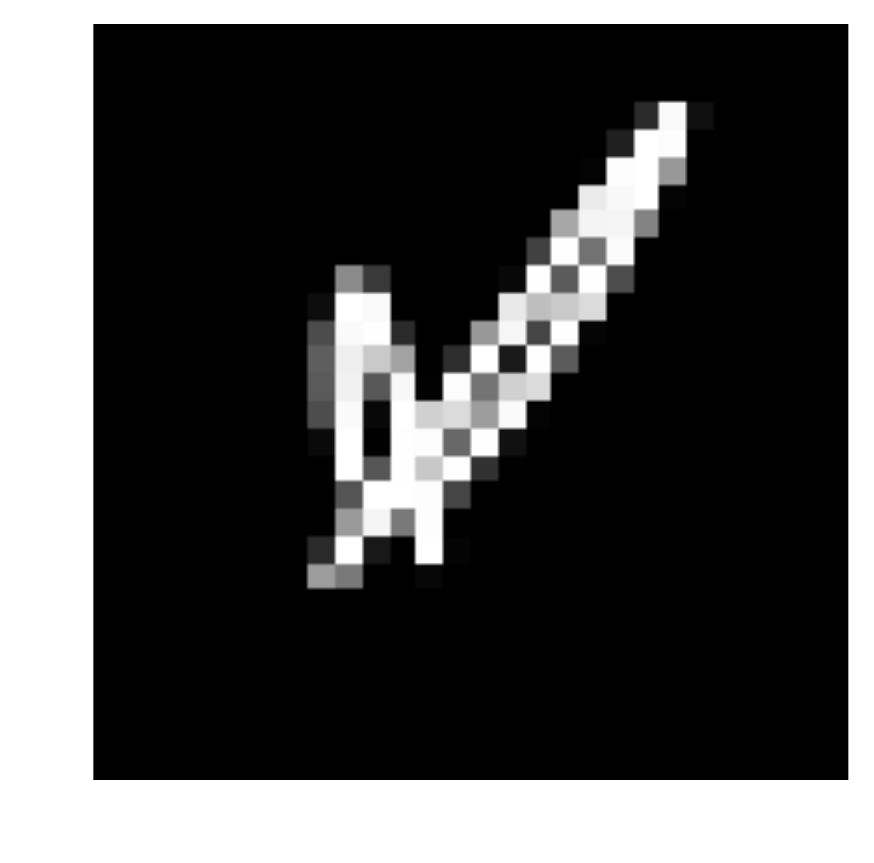} &
                \includegraphics{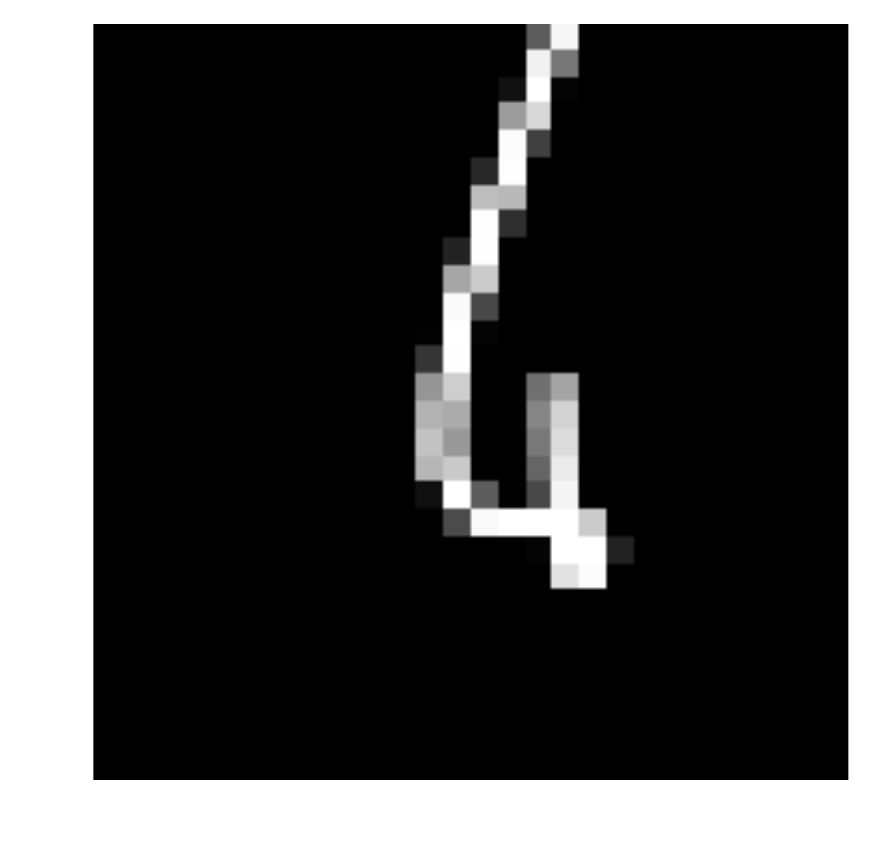} &
                \includegraphics{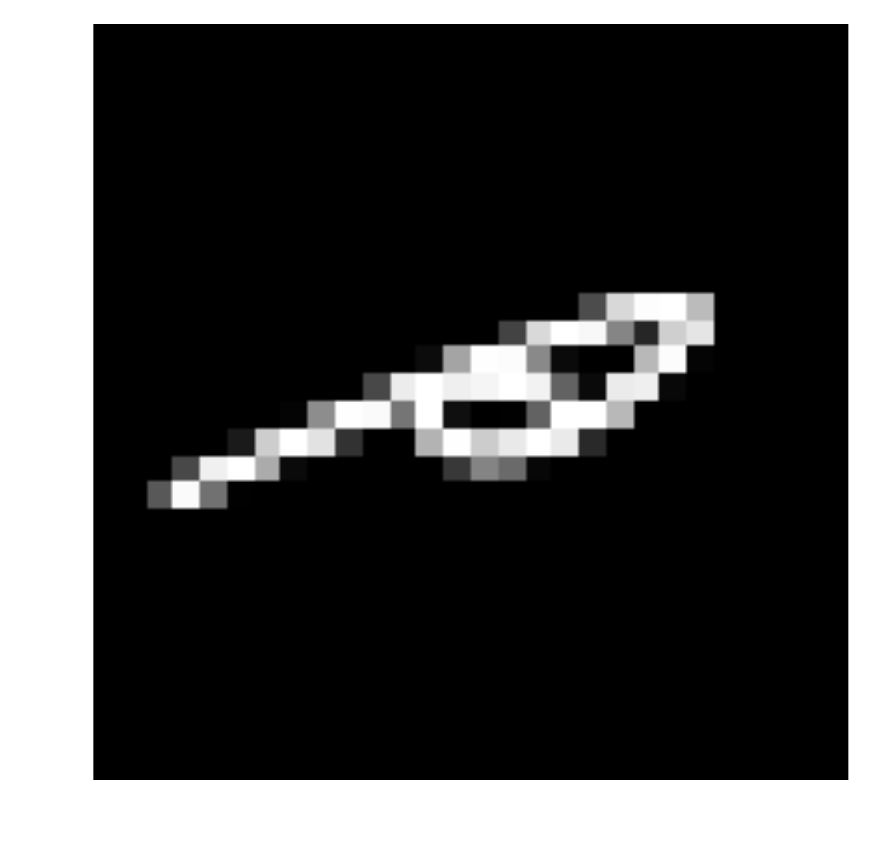} 
                \\
                {\fontsize{50}{60}\selectfont A} & {\fontsize{50}{60}\selectfont B}& {\fontsize{50}{60}\selectfont C}\\
                \includegraphics{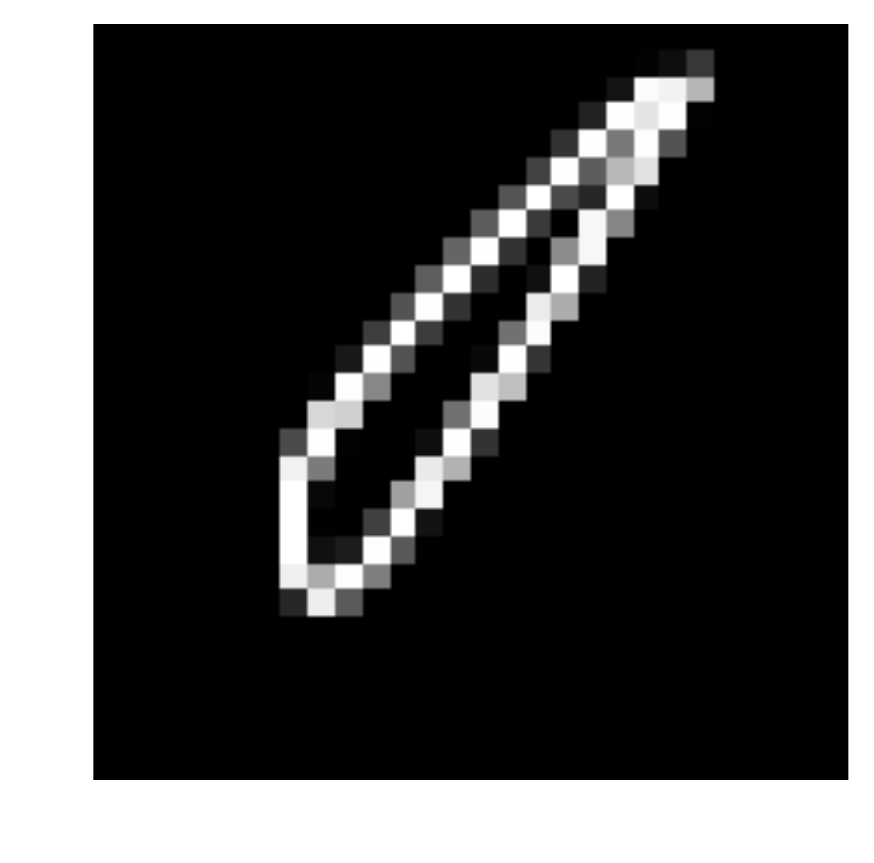} &
                \includegraphics{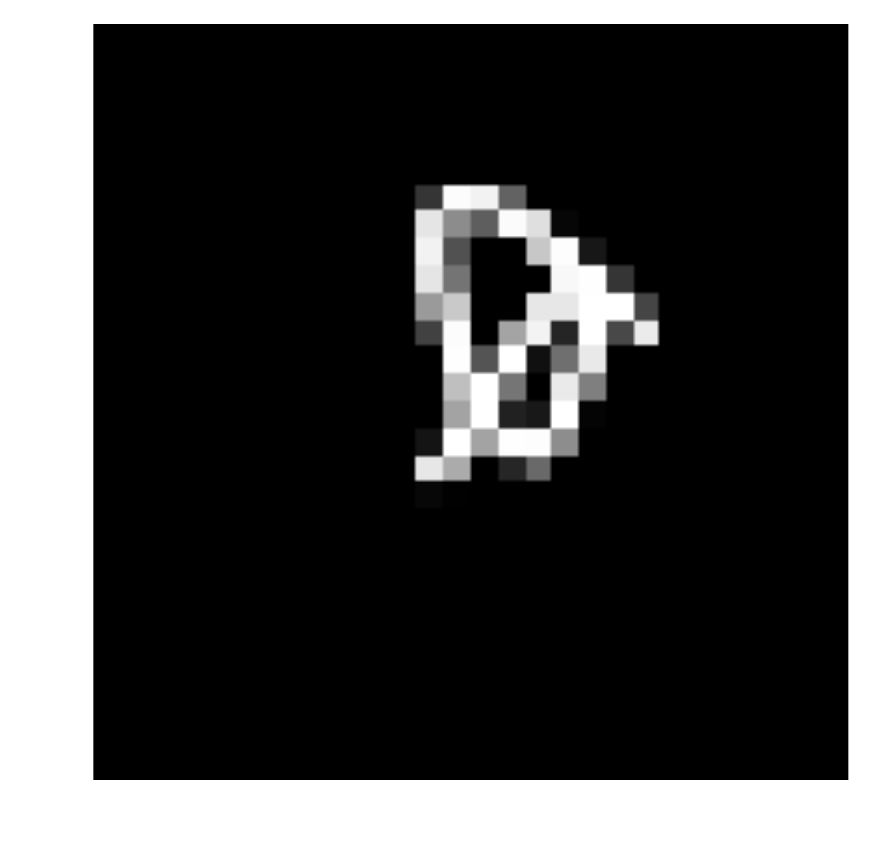} &
                \includegraphics{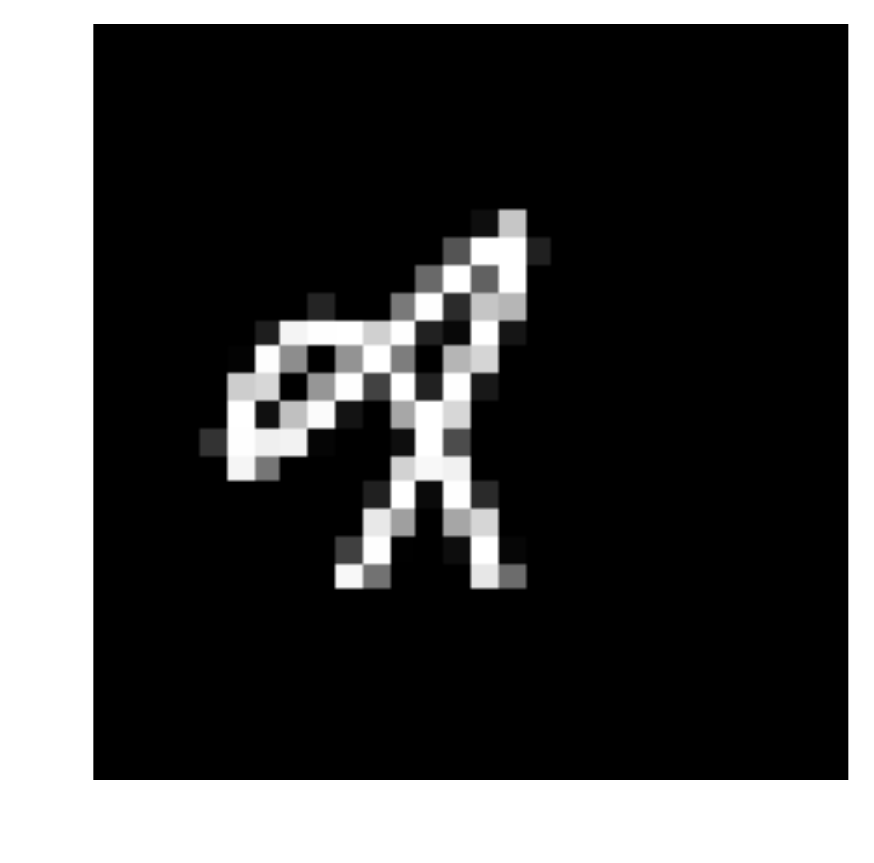} \\
                {\fontsize{50}{60}\selectfont D} & {\fontsize{50}{60}\selectfont E}& {\fontsize{50}{60}\selectfont F}\\
    
            \end{tabular}
        \end{adjustbox}
             &
         \raisebox{-5em}{\includegraphics[height=0.23\linewidth]{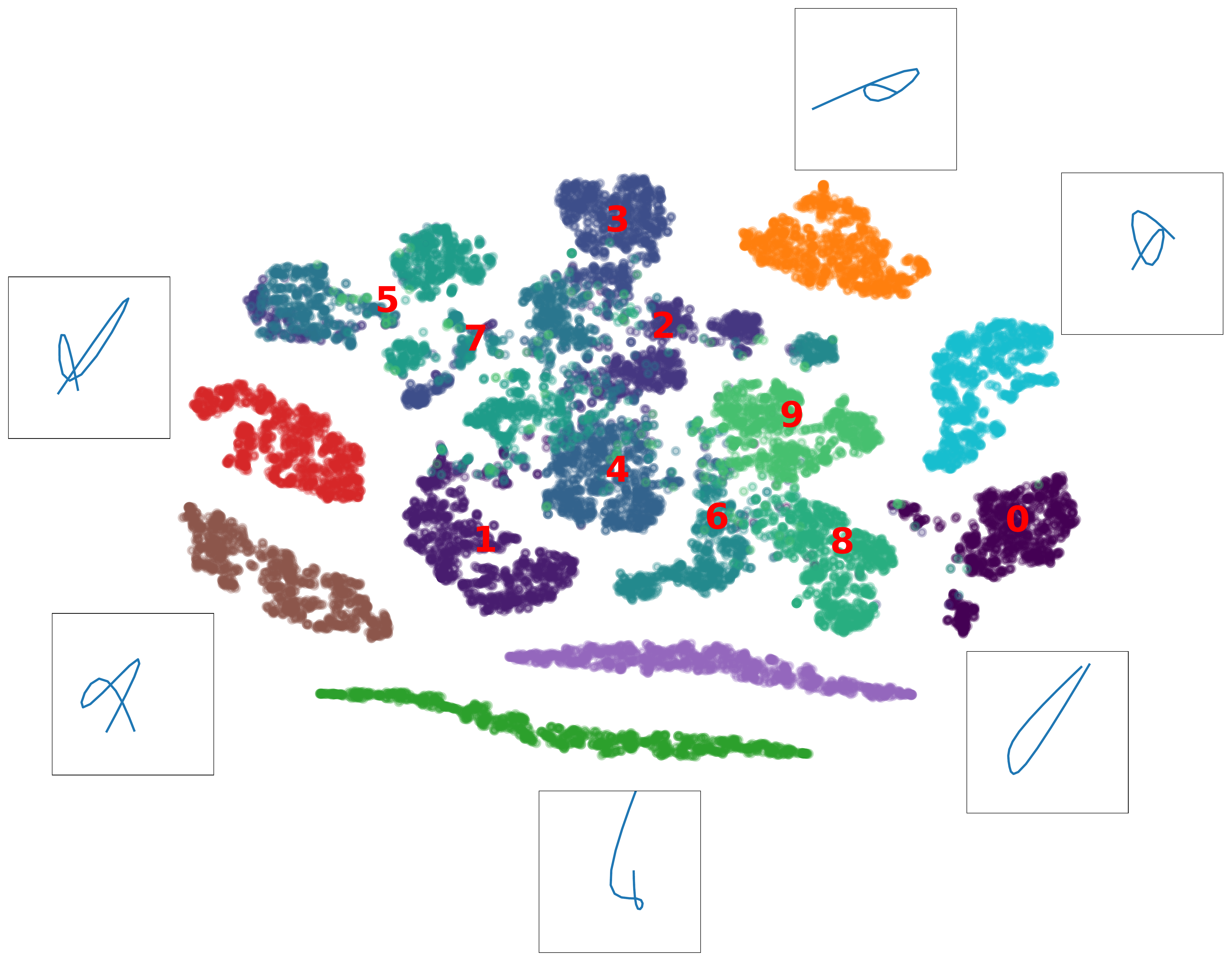}}
        \\
        (a) Generated 6 new digits &
        (b) The generated image's high level distribution
        \end{tabular}
    \end{adjustbox}
    
    \caption{\textbf{The generation of new sketch images.} (a) The new generated digits replacing A-F in hexadecimal system. (b) Projecting each digit's distribution to a 2-dimensional space for visualization by t-SNE~\cite{van2008visualizing}.}
    \label{fig:generation}
\end{figure*}

\textbf{Modifying control points} \label{sec:modifynodes}
In addition to robustness against changes in graph topologies, we show that our model is also robust against changes to control points of strokes (and therefore the graph representation). To this end, we tune the coordinates of the control points of a ``6'' towards label ``0'' using Eq.~\eqref{eq:attack}. 
$\mathcal{D}$ in the attack loss is set to the set of all ``0''s with a single stroke.
Fig.~\ref{fig:robust_feature1}(d) shows evolution of the sketch during the attack. At step 0, the image is recognized as ``6'' with a high confidence. While altering the control points at step 100, it is recognized as ``9'' with a confidence score $0.4573$ (while ``0'' and ``6'' receive scores of only $0.2552$ and $0.2710$. For the final step, the model recognizes the altered image as ``0'', with a high confidence. 
We show experimental results on other input samples in Fig.~\ref{fig:robust_feature1}(e,f). Again, results suggest that the graph representation is robust as successful attacks need to alter the semantic meaning of the sketches.

\subsection{New digits generation} \label{sec:gen}
Here we demonstrate that the graph representation enables generation of novel sketches that are separable in the feature space from the training data, while maintaining structural similarity. 
The underlying rationale is that if our stroke-based graph representation is with strongly structured expression capability, the underlying feature space after supervised training on existing categories could guide a generation process to come up with new categorical patterns. 

The formulation of the generation problem follows Eq.~\ref{eq:gen}.
To initialize a solution, we draw the control points of a new digit from a normal distribution $\mathcal{T} \sim \mathcal{N}(t, \sigma^2)$, where the mean $t$ is randomly sampled in a uniform distribution ranging from 4 to 24, and $\sigma$ is initially set to $4$. We consider cases where the graph topology is fixed. Alg.~\ref{generation} explains the procedure for solving Eq.~\ref{eq:gen}, with two alternating steps. In the first step, we focus on separating the new set $\mathcal{T}$ from the existing dataset $\mathcal{D}$. 
Fixing the feature extractor MPNN $f$, the binary classification objective function is given as:
\begin{equation}
    \begin{aligned}
        \min_{f_{\theta}} \quad  & - \mathbb{E}_{x \sim \mathcal{D}}\left[\log(f_\theta \circ f(g(x)))\right]  \\
        & - \mathbb{E}_{x \sim T}\left[\log(1-f_\theta \circ f(g(x)))\right] .
    \end{aligned}
\end{equation}
In the second step, we update the distribution of $\mathcal{T}$ through $t$ and $\sigma$ following Eq.~\ref{eq:gen}. The dataset $\mathcal{D}$ here in the loss function is the entire MNIST training set.

In our experiment, we generate a sequence of novel digits with a single stroke that are separable from the MNIST digits in the feature space, as illustrated in Fig.~\ref{fig:generation}(a). It is worth noting that the newly generated digits share a similar visual style to MNIST hand-written digits (although a quantitative analysis will require a Turing test~\cite{lake2015human}), and at the same time visually distinguishable from them. Fig.~\ref{fig:generation}(b) further confirms our claim, as we can see that on the space formed by the final MPNN network, all novel digits are separable from each other, and are distinguishable from the original set. The new digits generation experiment validates that our model has a strong structured expression capability. 

\begin{figure}
    \centering
    \begin{adjustbox}{width=0.77\linewidth}
        \begin{tabular}{c|c}
            \Huge{blueberry} & \Huge{broom} \\
            \begin{tabular}{cc}
                \includegraphics{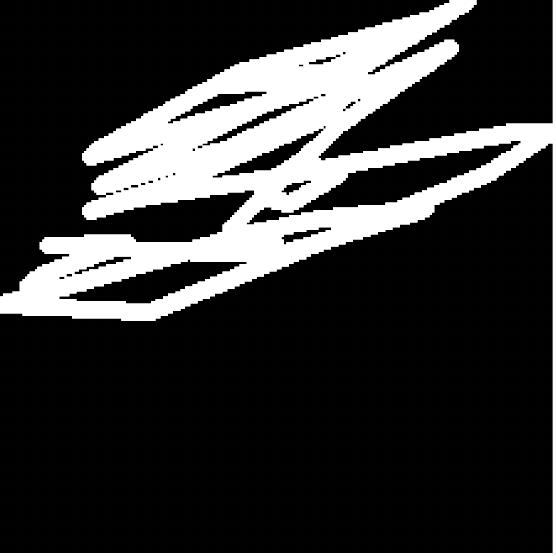} &
                \includegraphics{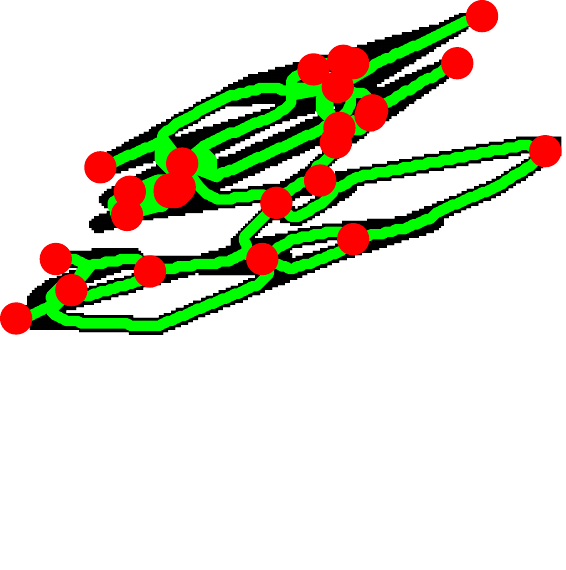}
            \end{tabular}&  
            \begin{tabular}{cc}
                \includegraphics{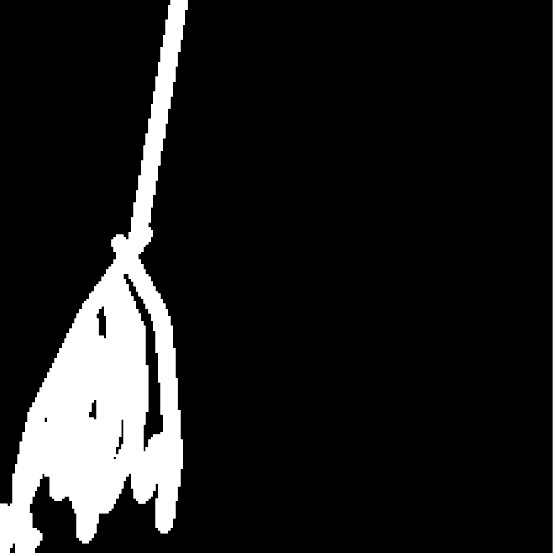} &
                \includegraphics{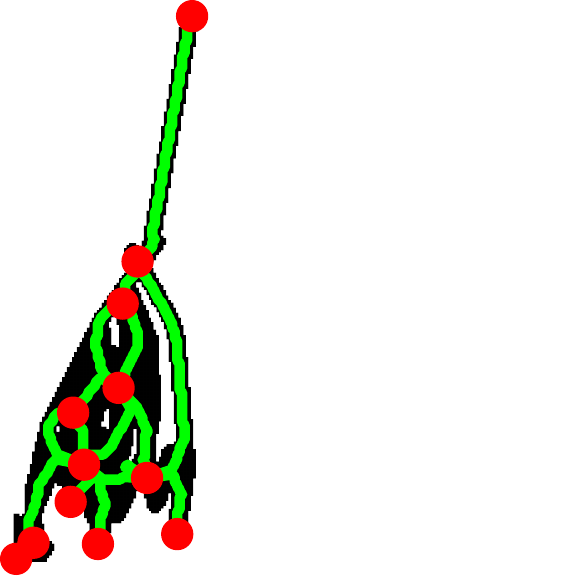} 
            \end{tabular}
            \\
            \hline
            \\
            \Huge{paintbrush} & \Huge{toaster} \\
            \begin{tabular}{cc}
                \includegraphics{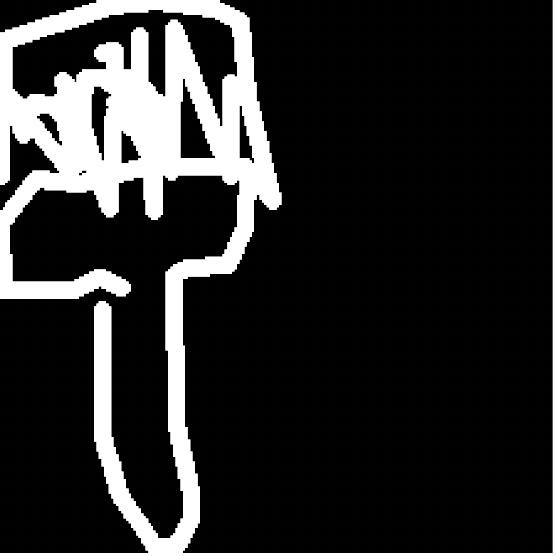} &
                \includegraphics{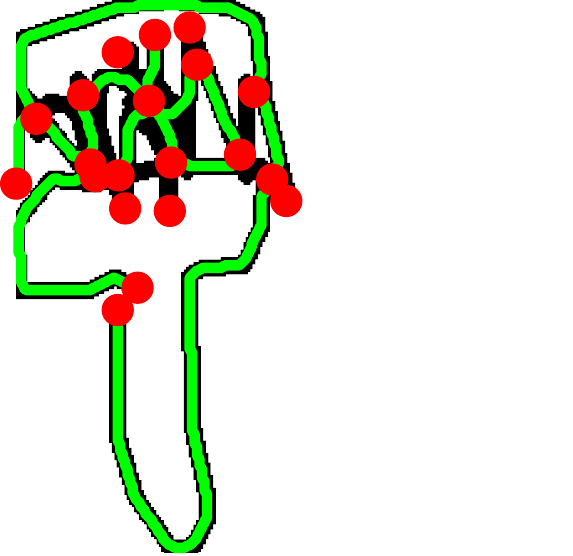} 
            \end{tabular}&  
             \begin{tabular}{cc}
                \includegraphics{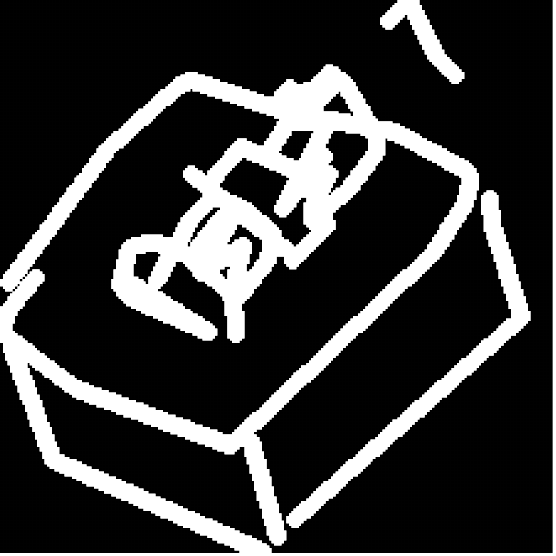} &
                \includegraphics{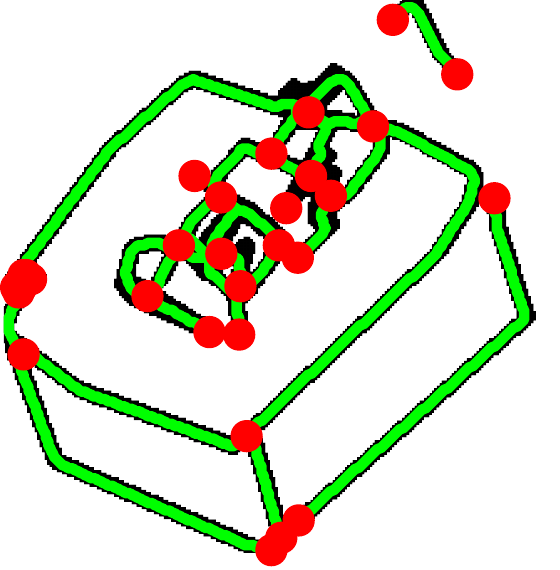} 
            \end{tabular}
        \end{tabular}
    \end{adjustbox}
    
    \caption{\textbf{The samples that our model struggles to handle.} We list 4 samples from 4 categories(blueberry, broom, paintbrush, toaster) that our method is not able to predict correctly. For each pair of the failure example,  the left image is the input sketch image with a dilation of 4 pixels, and the right image illustrates the strokes (green lines) and fork points (red dots).}
    \label{fig:limitation}
\end{figure}

\section{Limitations}
\noindent{\bf End-to-end stroke extraction} 
The major limitation in our method is in the preprocessing step. As shown in Fig.~\ref{fig:limitation}, many data points from the Quickdraw dataset contain strokes that form detailed parts of a whole or textures of parts, some of which can be quite abstract (see ``blueberry'' for example). Our stroke extraction procedure currently cannot correctly infer the stroke sequences of such sketches or produce abstraction of clustered strokes, e.g., those that represent a part with texture. 
To achieve this, we hypothesize that it is necessary to express the stroke extraction procedure as a differentiable program, so that it can be learned in an end-to-end fashion along with the GNN. Even so, it would still be questionable whether such strong extraction capabilities can be learned through static images. One idea that under our current investigate is to consider the ability of the extracted graph at predicting visual changes in dynamical environments during the learning of the stroke extraction program.  

\noindent{\bf Explainability of GNN} Apart from stroke extraction, we suspect that the observed limitation in accuracy (and thus robustness) is also due to the design of the mapping between the graph representation and its label. Specifically, there is a lack of connection between existing message passing architectures and the first-principal methods for classification through inference~\cite{lake2015human}.

\noindent{\bf Graph representations} The last source that accounts to the limited accuracy could be the graph representation. In this paper, we tested pairwise distances, which are invariant to rotation, translation, mirroring, and scaling (if all inputs are normalized). We are currently testing other potential representations with the same invariance properties, e.g., stroke curvature, and their combinations. However, a more systematic understanding of why some of these representations could work better is still missing.

\section{Conclusion and future work}

We present a stroke-based sketch representation with graph neural networks. We show that the proposed model is spatially robust (through robust classification and robust feature exploration experiments on MNIST and QuickDraw) with a strongly structured expression capability (through novel digits generation experiments). 

The promising properties of the model pave the way for a series of exciting future research, including but not limited to 1) stroke-based representation learning in an unsupervised manner; 2) augmenting the model's generalization capability by forming analogies between the graph representations. 3) forming representation of a complicated visual pattern with hierarchical graphs, further enhancing the structured expression capability of the model. 

\paragraph{Acknowledgement}
This work was supported by the  NSF DMR\#2020277 and the  DARPA GAILA ADAM project.


\newpage
{\small
\bibliographystyle{ieee_fullname}
\bibliography{egbib, YezhouBib}

\begin{thebibliography}{10}\itemsep=-1pt

\bibitem{popsci2017}
Kate Baggaley.
\newblock {There are two kinds of AI, and the difference is important: Most of
  today’s AI is designed to solve specific problems}.
\newblock {\em Popular Science}, 2017.

\bibitem{battaglia2018relational}
Peter~W Battaglia, Jessica~B Hamrick, Victor Bapst, Alvaro Sanchez-Gonzalez,
  Vinicius Zambaldi, Mateusz Malinowski, Andrea Tacchetti, David Raposo, Adam
  Santoro, Ryan Faulkner, et~al.
\newblock Relational inductive biases, deep learning, and graph networks.
\newblock {\em arXiv preprint arXiv:1806.01261}, 2018.

\bibitem{biederman1987recognition}
Irving Biederman.
\newblock Recognition-by-components: a theory of human image understanding.
\newblock {\em Psychological review}, 94(2):115, 1987.

\bibitem{bruna2013spectral}
Joan Bruna, Wojciech Zaremba, Arthur Szlam, and Yann LeCun.
\newblock Spectral networks and locally connected networks on graphs.
\newblock {\em arXiv preprint arXiv:1312.6203}, 2013.

\bibitem{cho2014properties}
Kyunghyun Cho, Bart Van~Merri{\"e}nboer, Dzmitry Bahdanau, and Yoshua Bengio.
\newblock On the properties of neural machine translation: Encoder-decoder
  approaches.
\newblock {\em arXiv preprint arXiv:1409.1259}, 2014.

\bibitem{cho2013learning}
Minsu Cho, Karteek Alahari, and Jean Ponce.
\newblock Learning graphs to match.
\newblock In {\em Proceedings of the IEEE International Conference on Computer
  Vision}, pages 25--32, 2013.

\bibitem{clark2015elementary}
Peter Clark.
\newblock {Elementary school science and math tests as a driver for AI: take
  the ARISTO challenge}.
\newblock In {\em AAAI}, pages 4019--4021, 2015.

\bibitem{collomosse2009storyboard}
John~P Collomosse, Graham McNeill, and Yu Qian.
\newblock Storyboard sketches for content based video retrieval.
\newblock In {\em 2009 IEEE 12th International Conference on Computer Vision},
  pages 245--252. IEEE, 2009.

\bibitem{desolneux2007gestalt}
Agnes Desolneux, Lionel Moisan, and Jean-Michel Morel.
\newblock {\em From gestalt theory to image analysis: a probabilistic
  approach}, volume~34.
\newblock Springer Science \& Business Media, 2007.

\bibitem{duvenaud2015convolutional}
David Duvenaud, Dougal Maclaurin, Jorge Aguilera-Iparraguirre, Rafael
  G{\'o}mez-Bombarelli, Timothy Hirzel, Al{\'a}n Aspuru-Guzik, and Ryan~P
  Adams.
\newblock Convolutional networks on graphs for learning molecular fingerprints.
\newblock {\em arXiv preprint arXiv:1509.09292}, 2015.

\bibitem{engstrom2019exploring}
Logan Engstrom, Brandon Tran, Dimitris Tsipras, Ludwig Schmidt, and Aleksander
  Madry.
\newblock Exploring the landscape of spatial robustness.
\newblock In {\em International Conference on Machine Learning}, pages
  1802--1811. PMLR, 2019.

\bibitem{geirhos2018imagenettrained}
Robert Geirhos, Patricia Rubisch, Claudio Michaelis, Matthias Bethge, Felix~A.
  Wichmann, and Wieland Brendel.
\newblock Imagenet-trained {CNN}s are biased towards texture; increasing shape
  bias improves accuracy and robustness.
\newblock In {\em International Conference on Learning Representations}, 2019.

\bibitem{gilmer2017neural}
Justin Gilmer, Samuel~S Schoenholz, Patrick~F Riley, Oriol Vinyals, and
  George~E Dahl.
\newblock Neural message passing for quantum chemistry.
\newblock In {\em International conference on machine learning}, pages
  1263--1272. PMLR, 2017.

\bibitem{ha2017neural}
David Ha and Douglas Eck.
\newblock A neural representation of sketch drawings.
\newblock {\em arXiv preprint arXiv:1704.03477}, 2017.

\bibitem{he2017sketch}
Jun-Yan He, Xiao Wu, Yu-Gang Jiang, Bo Zhao, and Qiang Peng.
\newblock Sketch recognition with deep visual-sequential fusion model.
\newblock In {\em Proceedings of the 25th ACM international conference on
  Multimedia}, pages 448--456, 2017.

\bibitem{hinton1979some}
Geoffrey Hinton.
\newblock Some demonstrations of the effects of structural descriptions in
  mental imagery.
\newblock {\em Cognitive Science}, 3(3):231--250, 1979.

\bibitem{hinton2021represent}
Geoffrey Hinton.
\newblock How to represent part-whole hierarchies in a neural network.
\newblock {\em arXiv preprint arXiv:2102.12627}, 2021.

\bibitem{hosseini2018assessing}
Hossein Hosseini, Baicen Xiao, Mayoore Jaiswal, and Radha Poovendran.
\newblock Assessing shape bias property of convolutional neural networks.
\newblock In {\em Proceedings of the IEEE Conference on Computer Vision and
  Pattern Recognition Workshops}, pages 1923--1931, 2018.

\bibitem{hummel1992dynamic}
John~E Hummel and Irving Biederman.
\newblock Dynamic binding in a neural network for shape recognition.
\newblock {\em Psychological review}, 99(3):480, 1992.

\bibitem{NEURIPS2019_e2c420d9}
Andrew Ilyas, Shibani Santurkar, Dimitris Tsipras, Logan Engstrom, Brandon
  Tran, and Aleksander Madry.
\newblock Adversarial examples are not bugs, they are features.
\newblock In {\em Advances in Neural Information Processing Systems},
  volume~32, 2019.

\bibitem{jia2017sequential}
Qi Jia, Meiyu Yu, Xin Fan, and Haojie Li.
\newblock Sequential dual deep learning with shape and texture features for
  sketch recognition.
\newblock {\em arXiv preprint arXiv:1708.02716}, 2017.

\bibitem{kodratoff1984learning}
Yves Kodratoff et~al.
\newblock Learning complex structural descriptions from examples.
\newblock {\em Computer vision, graphics, and image processing},
  27(3):266--290, 1984.

\bibitem{kriegeskorte2015deep}
Nikolaus Kriegeskorte.
\newblock Deep neural networks: a new framework for modeling biological vision
  and brain information processing.
\newblock {\em Annual review of vision science}, 1:417--446, 2015.

\bibitem{kubilius2016deep}
Jonas Kubilius, Stefania Bracci, and Hans~P Op~de Beeck.
\newblock Deep neural networks as a computational model for human shape
  sensitivity.
\newblock {\em PLoS computational biology}, 12(4):e1004896, 2016.

\bibitem{lake2012concept}
Brenden Lake, Ruslan Salakhutdinov, and Joshua Tenenbaum.
\newblock Concept learning as motor program induction: A large-scale empirical
  study.
\newblock In {\em Proceedings of the Annual Meeting of the Cognitive Science
  Society}, volume~34, 2012.

\bibitem{lake2015human}
Brenden~M Lake, Ruslan Salakhutdinov, and Joshua~B Tenenbaum.
\newblock Human-level concept learning through probabilistic program induction.
\newblock {\em Science}, 350(6266):1332--1338, 2015.

\bibitem{lake2017building}
Brenden~M Lake, Tomer~D Ullman, Joshua~B Tenenbaum, and Samuel~J Gershman.
\newblock Building machines that learn and think like people.
\newblock {\em Behavioral and brain sciences}, 40, 2017.

\bibitem{lam1992thinning}
Louisa Lam, Seong-Whan Lee, Ching~Y Suen, et~al.
\newblock Thinning methodologies-a comprehensive survey.
\newblock {\em IEEE Transactions on pattern analysis and machine intelligence},
  14(9):869--885, 1992.

\bibitem{landau1988importance}
Barbara Landau, Linda~B Smith, and Susan~S Jones.
\newblock The importance of shape in early lexical learning.
\newblock {\em Cognitive development}, 3(3):299--321, 1988.

\bibitem{levesque2011winograd}
Hector~J Levesque, Ernest Davis, and Leora Morgenstern.
\newblock The winograd schema challenge.
\newblock In {\em AAAI Spring Symposium: Logical Formalizations of Commonsense
  Reasoning}, 2011.

\bibitem{li2015gated}
Yujia Li, Daniel Tarlow, Marc Brockschmidt, and Richard Zemel.
\newblock Gated graph sequence neural networks.
\newblock {\em arXiv preprint arXiv:1511.05493}, 2015.

\bibitem{liao1990stroke}
Chia-Wei Liao and Jun~S Huang.
\newblock Stroke segmentation by bernstein-bezier curve fitting.
\newblock {\em Pattern Recognition}, 23(5):475--484, 1990.

\bibitem{liu1999identification}
Ke Liu, Yea~S. Huang, and Ching~Y. Suen.
\newblock Identification of fork points on the skeletons of handwritten chinese
  characters.
\newblock {\em IEEE transactions on pattern analysis and machine intelligence},
  21(10):1095--1100, 1999.

\bibitem{lowe2004distinctive}
David~G Lowe.
\newblock Distinctive image features from scale-invariant keypoints.
\newblock {\em International journal of computer vision}, 60(2):91--110, 2004.

\bibitem{mahdavi2020visual}
Mahshad Mahdavi and Richard Zanibbi.
\newblock Visual parsing with query-driven global graph attention (qd-gga):
  preliminary results for handwritten math formula recognition.
\newblock In {\em Proceedings of the IEEE/CVF Conference on Computer Vision and
  Pattern Recognition Workshops}, pages 570--571, 2020.

\bibitem{newyorker2013}
Gary Marcus.
\newblock {Why Can’t my computer understand me}.
\newblock {\em {The New Yorker}}, 2013.

\bibitem{nytimes2017gm}
Gary Marcus.
\newblock {Artificial Intelligence Is Stuck. Here’s How to Move It Forward.}
\newblock {\em {New York Times}}, 2017.

\bibitem{ramer1972iterative}
Urs Ramer.
\newblock An iterative procedure for the polygonal approximation of plane
  curves.
\newblock {\em Computer graphics and image processing}, 1(3):244--256, 1972.

\bibitem{riba2015handwritten}
Pau Riba, Josep Llad{\~a}s, and Alicia Forn{\'e}s.
\newblock Handwritten word spotting by inexact matching of grapheme graphs.
\newblock In {\em 2015 13th International Conference on Document Analysis and
  Recognition (ICDAR)}, pages 781--785. IEEE, 2015.

\bibitem{ribeiro2020sketchformer}
Leo Sampaio~Ferraz Ribeiro, Tu Bui, John Collomosse, and Moacir Ponti.
\newblock Sketchformer: Transformer-based representation for sketched
  structure.
\newblock In {\em Proceedings of the IEEE/CVF Conference on Computer Vision and
  Pattern Recognition}, pages 14153--14162, 2020.

\bibitem{ritter2017cognitive}
Samuel Ritter, David~GT Barrett, Adam Santoro, and Matt~M Botvinick.
\newblock Cognitive psychology for deep neural networks: A shape bias case
  study.
\newblock In {\em International conference on machine learning}, pages
  2940--2949. PMLR, 2017.

\bibitem{sabour2018matrix}
Sara Sabour, Nicholas Frosst, and Geoffrey Hinton.
\newblock Matrix capsules with em routing.
\newblock In {\em 6th international conference on learning representations,
  ICLR}, volume 115, 2018.

\bibitem{scarselli2008graph}
Franco Scarselli, Marco Gori, Ah~Chung Tsoi, Markus Hagenbuchner, and Gabriele
  Monfardini.
\newblock The graph neural network model.
\newblock {\em IEEE transactions on neural networks}, 20(1):61--80, 2008.

\bibitem{singh2001part}
Manish Singh and Donald~D Hoffman.
\newblock Part-based representations of visual shape and implications for
  visual cognition.
\newblock In {\em Advances in psychology}, volume 130, pages 401--459.
  Elsevier, 2001.

\bibitem{bi2017}
Sam Snead.
\newblock {Facebook's AI boss: `In terms of general intelligence, we’re not
  even close to a rat’}.
\newblock {\em Business Insider}, 2017.

\bibitem{song2017deep}
Jifei Song, Qian Yu, Yi-Zhe Song, Tao Xiang, and Timothy~M Hospedales.
\newblock Deep spatial-semantic attention for fine-grained sketch-based image
  retrieval.
\newblock In {\em Proceedings of the IEEE international conference on computer
  vision}, pages 5551--5560, 2017.

\bibitem{szegedy2015going}
Christian Szegedy, Wei Liu, Yangqing Jia, Pierre Sermanet, Scott Reed, Dragomir
  Anguelov, Dumitru Erhan, Vincent Vanhoucke, and Andrew Rabinovich.
\newblock Going deeper with convolutions.
\newblock In {\em Proceedings of the IEEE conference on computer vision and
  pattern recognition}, pages 1--9, 2015.

\bibitem{tsipras2018robustness}
Dimitris Tsipras, Shibani Santurkar, Logan Engstrom, Alexander Turner, and
  Aleksander Madry.
\newblock Robustness may be at odds with accuracy.
\newblock In {\em International Conference on Learning Representations}, 2019.

\bibitem{van2015part}
Anton van~den Hengel, Chris Russell, Anthony Dick, John Bastian, Daniel Pooley,
  Lachlan Fleming, and Lourdes Agapito.
\newblock Part-based modelling of compound scenes from images.
\newblock In {\em Proceedings of the IEEE Conference on Computer Vision and
  Pattern Recognition}, pages 878--886, 2015.

\bibitem{van2008visualizing}
Laurens Van~der Maaten and Geoffrey Hinton.
\newblock Visualizing data using t-sne.
\newblock {\em Journal of machine learning research}, 9(11), 2008.

\bibitem{wu2014learning}
Qi Wu, Hongping Cai, and Peter Hall.
\newblock Learning graphs to model visual objects across different depictive
  styles.
\newblock In {\em European Conference on Computer Vision}, pages 313--328.
  Springer, 2014.

\bibitem{xie2019deep}
Yao Xie, Peng Xu, and Zhanyu Ma.
\newblock Deep zero-shot learning for scene sketch.
\newblock In {\em 2019 IEEE International Conference on Image Processing
  (ICIP)}, pages 3661--3665. IEEE, 2019.

\bibitem{xu2020deep}
Peng Xu, Timothy~M Hospedales, Qiyue Yin, Yi-Zhe Song, Tao Xiang, and Liang
  Wang.
\newblock Deep learning for free-hand sketch: A survey and a toolbox.
\newblock {\em arXiv preprint arXiv:2001.02600}, 2020.

\bibitem{xu2018sketchmate}
Peng Xu, Yongye Huang, Tongtong Yuan, Kaiyue Pang, Yi-Zhe Song, Tao Xiang,
  Timothy~M Hospedales, Zhanyu Ma, and Jun Guo.
\newblock Sketchmate: Deep hashing for million-scale human sketch retrieval.
\newblock In {\em Proceedings of the IEEE conference on computer vision and
  pattern recognition}, pages 8090--8098, 2018.

\bibitem{xu2020learning}
Peng Xu, Yongye Huang, Tongtong Yuan, Tao Xiang, Timothy~M Hospedales, Yi-Zhe
  Song, and Liang Wang.
\newblock On learning semantic representations for large-scale abstract
  sketches.
\newblock {\em IEEE Transactions on Circuits and Systems for Video Technology},
  31(9):3366--3379, 2020.

\bibitem{xu2021multigraph}
Peng Xu, Chaitanya~K Joshi, and Xavier Bresson.
\newblock Multigraph transformer for free-hand sketch recognition.
\newblock {\em IEEE Transactions on Neural Networks and Learning Systems},
  2021.

\bibitem{xu2020fine}
Peng Xu, Kun Liu, Tao Xiang, Timothy~M Hospedales, Zhanyu Ma, Jun Guo, and
  Yi-Zhe Song.
\newblock Fine-grained instance-level sketch-based video retrieval.
\newblock {\em IEEE Transactions on Circuits and Systems for Video Technology},
  31(5):1995--2007, 2020.

\bibitem{yang2021sketchaa}
Lan Yang, Kaiyue Pang, Honggang Zhang, and Yi-Zhe Song.
\newblock Sketchaa: Abstract representation for abstract sketches.
\newblock In {\em Proceedings of the IEEE/CVF International Conference on
  Computer Vision}, pages 10097--10106, 2021.

\bibitem{yang2021sketchgnn}
Lumin Yang, Jiajie Zhuang, Hongbo Fu, Xiangzhi Wei, Kun Zhou, and Youyi Zheng.
\newblock Sketchgnn: Semantic sketch segmentation with graph neural networks.
\newblock {\em ACM Transactions on Graphics (TOG)}, 40(3):1--13, 2021.

\bibitem{yang2020sketchgcn}
Lumin Yang, Jiajie Zhuang, Hongbo Fu, Kun Zhou, and Youyi Zheng.
\newblock Sketchgcn: Semantic sketch segmentation with graph convolutional
  networks.
\newblock {\em arXiv preprint arXiv:2003.00678}, 3, 2020.

\bibitem{yu2017sketch}
Qian Yu, Yongxin Yang, Feng Liu, Yi-Zhe Song, Tao Xiang, and Timothy~M
  Hospedales.
\newblock Sketch-a-net: A deep neural network that beats humans.
\newblock {\em International journal of computer vision}, 122(3):411--425,
  2017.

\end{thebibliography}
}

\end{document}